\documentclass{article}

     \usepackage[final]{neurips_2020}
     
\usepackage[utf8]{inputenc} 
\usepackage[T1]{fontenc}    
\usepackage{hyperref}       
\usepackage{url}            
\usepackage{booktabs}       
\usepackage{amsfonts}       
\usepackage{nicefrac}       
\usepackage{microtype}      

\usepackage{graphicx}
\usepackage{wrapfig}
\usepackage{subcaption}
\usepackage[ruled]{algorithm2e}
\usepackage{amssymb,amsmath}
\usepackage{amsthm}

\newtheorem*{theorem*}{Theorem}

\usepackage{bm}
\def\given{\middle\vert}
\def\grad{\nabla}
\def\expectation{\mathbb{E}}
\def\prob{P}
\def\defeq{\dot=}
\newcommand{\deriv}[2][]{\frac{\partial#1}{\partial#2}}

\def\states{S}

\def\Action{A}

\def\actions{A}

\def\Var{\mathrm{Var}}

\title{Reward Propagation \\ Using Graph Convolutional Networks}

\author{
    Martin Klissarov \\
    Mila, McGill University\\
    \texttt{martin.klissarov@mail.mcgill.ca} \\
    \And
    Doina Precup \\
    Mila, McGill University and DeepMind\\
    \texttt{dprecup@cs.mcgill.ca} \\
}

\begin{document}

\maketitle

\begin{abstract}
Potential-based reward shaping provides an approach for designing good reward functions, with the purpose of speeding up learning. However, automatically finding potential functions for complex environments is a difficult problem (in fact, of the same difficulty as learning a value function from scratch). We propose a new framework for learning potential functions by leveraging ideas from graph representation learning. Our approach relies on Graph Convolutional Networks which we use as a key ingredient in combination with the probabilistic inference view of reinforcement learning. More precisely, we leverage Graph Convolutional Networks to perform message passing from rewarding states. The propagated messages can then be used as potential functions for reward shaping to accelerate learning. We verify empirically that our approach can achieve considerable improvements in both small and high-dimensional control problems.
\end{abstract}

\section{Introduction}

Reinforcement learning (RL) algorithms provide a solution to the problem of learning a policy that optimizes an expected, cumulative function of rewards. Consequently, a good reward function is critical to the practical success of these algorithms. However, designing such a function can be challenging \citep{DBLP:journals/corr/AmodeiOSCSM16}. Approaches to this problem include, amongst others, intrinsic motivation~\citep{Oudeyer2007WhatII,5508364}, optimal rewards~\citep{5471106} and potential-based reward shaping~\citep{Ng:1999:PIU:645528.657613}. The latter provides an appealing formulation as it does not change the optimal policy of an MDP while potentially speeding up the learning process. However, the design of potential functions used for reward shaping is still an open question.

In this paper, we present a solution to this problem by leveraging the probabilistic inference view of RL~\citep{Toussaint:2006:PIS:1143844.1143963,ziebart2008maximum}. In particular, we are interested in formulating the RL problem as a directed graph whose structure is analogous to hidden Markov models. In such graphs it is convenient to perform inference through message passing with algorithms such as the forward-backward algorithm ~\citep{Rabiner86anintroduction}. This inference procedure essentially propagates information from the rewarding states in the MDP and results in a function over states. This function could then naturally be leveraged as a potential function for potential-based reward shaping.
However, the main drawback of traditional message passing algorithms is that they can be computationally expensive and are therefore hard to scale to large or continuous state space. 

We present an implementation that is both scalable and flexible by drawing connections to spectral graph theory~\citep{Chung97a}. We use Graph Convolutional Networks (GCN) ~\citep{DBLP:journals/corr/KipfW16} to propagate information about the rewards in an environment through the message passing mechanism defined by the GCN's structural bias and loss function.  
Indeed, GCNs belong to the larger class of Message Passing Neural Networks \citep{DBLP:journals/corr/GilmerSRVD17} with the special characteristic that their message passing mechanism builds on the graph Laplacian. The framework of Proto-Value Functions \citep{Mahadevan2005} from the reinforcement learning literature has previously studied the properties of the graph Laplacian and we build on these findings in our work.


We first evaluate our approach in tabular domains where we achieve similar performance compared to potential based reward shaping built on the forward-backward algorithm. Unlike hand-engineered potential functions, our method scales naturally to more complex environments; we illustrate this on navigation-based vision tasks from the MiniWorld environment ~\citep{gym_miniworld}, on a variety of games from the Atari 2600 benchmark \citep{DBLP:journals/corr/abs-1207-4708} and on a set of continuous control environments based on MuJoCo \citep{conf/iros/TodorovET12}  , where our method fares significantly better than actor-critic algorithms \citep{Sutton1999,SchulmanWDRK17} and additional baselines.

\section{Background and Motivation}
A Markov Decision Process $\mathcal{M}$ is a tuple $\langle \mathcal{S}, \mathcal{A}, \gamma, r, P \rangle$ with a finite state space $\mathcal{S}$, a finite action space $\mathcal{A}$, discount factor $\gamma\in [0,1) $, a scalar reward function $r : \states \times \actions \to Dist(\mathbb{R})$ and a transition probability distribution $P: \states \times \actions \to Dist(\states)$. A policy $\pi:\states \to Dist(\actions)$ specifies a way of behaving, and its value function is the expected return obtained by following $\pi$: $V_\pi(s) \defeq \expectation_\pi\left[ \sum_{i=t}^\infty \gamma^{t-i} r(S_i, A_i) \given S_t = s\right]$.
The value function $V_\pi$ satisfies the following Bellman equation:
$V_\pi(s) = \sum_{a} \pi\left(a \given s\right)\left( r(s, a) + \gamma \sum_{s'} \prob\left(s' \given s, a\right) V_\pi(s')\right)$
 where $s'$ is the state following state $s$.
The policy gradient theorem \citep{Sutton1999} provides the gradient of the expected discounted return from an initial state distribution $d(s_0)$ with respect to a parameterized stochastic policy $\pi_\theta$: $\deriv[J(\theta)]{\theta} = \sum_{s} d(s;\theta) \sum_{a} \deriv[\pi\left(a \given s\right)]{\theta}Q_{\pi}(s, a)$
where we simply write $\pi$ for $\pi_{\theta}$ for ease of notation and $d(s;\theta) = \sum_{s_0} d(s_0) \sum_{t=0}^{\infty} \gamma^t P^{\pi}(S_t = s | S_0 = s_0)$ is the discounted state occupancy measure. 

\paragraph{Reward Shaping.}

The framework reward shaping augments the original reward function by adding a shaping function, resulting in the following equation: $R'(S_t,A_t,S_{t+1}) = r(S_t,A_t) + F(S_t,S_{t+1})$
where  $F(S_t,S_{t+1})$ is the shaping function which can encode expert knowledge or represent concepts such as curiosity~\citep{5508364,Oudeyer2007WhatII}. \citet{Ng:1999:PIU:645528.657613} showed that a necessary and sufficient condition for preserving the MDP's optimal policy when using $R'$ instead of $r$ is for the shaping function to take the following form,
\begin{align*}
    F(S_t,S_{t+1}) = \gamma\Phi(S_{t+1}) - \Phi(S_t)
\end{align*}
where $\Phi$ is the scalar potential function $\Phi : \states  \to \mathbb{R}$, which can be any arbitrary function defined on states. 
In their work, the potential function was defined as a distance to a goal position. 
Different alternatives have been explored since, such as learning from human feedback~\citep{Harutyunyan:2015:SMH:2772879.2773501} or using similarity-based shaping for learning from demonstrations~\citep{Brys:2015:RLD:2832581.2832716}. \citet{10.1145/1273496.1273572,grzes} have also considered automatically learning the potential function. These approaches either require a human in the loop or are not easily scalable to large problems. 
A key difference and contribution of this work is that we instead aim to learn end-to-end the potential function that scale naturally to complex environments.

\paragraph{RL as Probabilistic Inference.}
 \begin{wrapfigure}{r}{.3\textwidth}
    \centering
     \includegraphics[width=0.3\textwidth]{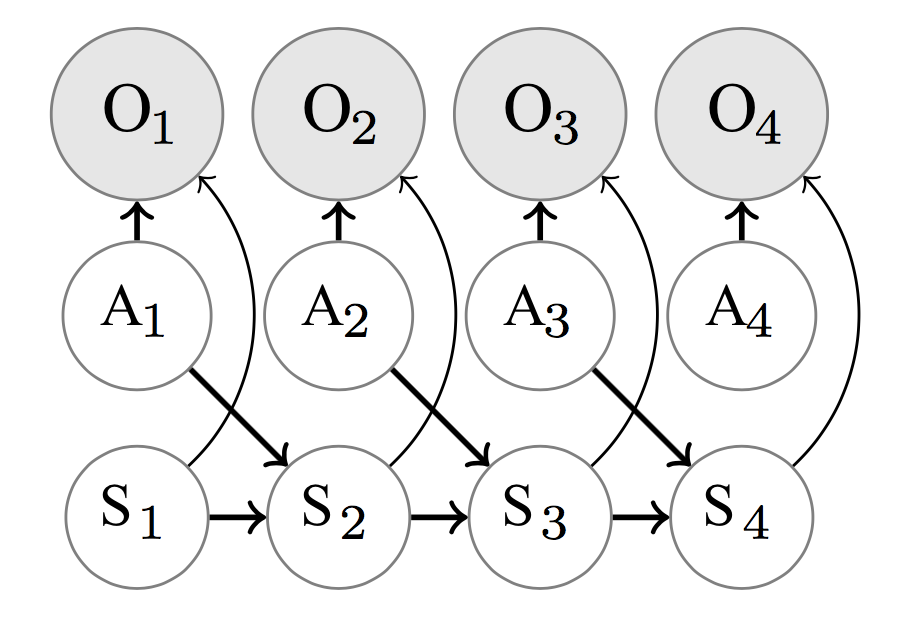}
  \caption{\small \textbf{Graphical model of the control task in RL.} $O_t$ is an observed variable, while $\states_t$ and $\Action_t$ are the state and action latent variables. } 
   \label{fig:graphmodel}
\end{wrapfigure}
 Consider the graphical model in Fig.\ref{fig:graphmodel}, where $O_t$ is a binary variable dependent on the action $A_t$ and the state $S_t$. The distribution over this optimality variable is defined with respect to the reward:
\begin{align*}
    p( O_t = 1 | S_t, A_t) = f(r(S_t,A_t))
\end{align*}
where $f(r(S_t,A_t))$ is a function used to map rewards into probability space. In previous work \citep{Toussaint:2009:RTO:1553374.1553508,DBLP:journals/corr/abs-1805-00909}, this is taken to be the exponential function (together with the assumption that rewards are negative) in order to facilitate derivation. In our case, we simply use the sigmoid function to allow for any types of rewards. In this model, the $O_t$ variables are considered to be observed while the actions $A_t$ and states $S_t$ are latent. 

This particular structure is analogous to hidden Markov models (HMM), where we can derive the forward-backward messages to perform inference \citep{ziebart2008maximum,Kappen2012OptimalCA,Toussaint:2006:PIS:1143844.1143963}. As we show next, in the case of reinforcement learning as a graphical model, message passing can be understood as inferring a function over states that is closely related to the value function.

As in the case of HMMs, the forward-backward messages in the reinforcement learning graph take following form, 
 $\beta(S_t) = p(O_{t:T}|S_t)$ and  $\alpha(S_t) = p(O_{0:t-1}|S_t)p(S_t)$
 where the notation $O_{t:T}$ defines the set of variables $(O_t,O_{t+1},...,O_T)$. 
 The backward messages $\beta(S_t)$ represent the probability of a future optimal trajectory given the current state, where future optimal trajectory can be translated as the probability of a high return given a state. This measure can be shown to be the projection into probability space of an optimistic variant of the value function from maximum-entropy RL  \citep{DBLP:journals/corr/abs-1805-00909}. In the case of the forward messages $\alpha(S_t)$, they represent the probability of a past optimal trajectory, scaled by the probability of the current state. There is currently no similar quantity in reinforcement learning, but a natural analogue would be to consider value functions that look backwards in time $V^{\pi}_{\alpha}(s) = \expectation{[\sum_{i=0}^{t-1} \gamma^{t-i} r(S_i,A_i)|S_t=s]}$.

By using the conditional independence properties of the RL graph, the terms for the forward and backward messages  can be expanded to obtain their recursive form,
 \begin{align}
     \alpha(S_t,A_t) 
     &= \sum_{S_{t-1}}\sum_{A_{t-1}} p(S_t|S_{t-1},A_{t-1})p(A_t)
      p(O_{t-1}|S_{t-1},A_{t-1})\alpha(S_{t-1},A_{t-1}) \label{alpha}
      \\
     \beta(S_t,A_t)
     &= \sum_{S_{t+1}}\sum_{A_{t+1}} p(S_{t+1}|S_{t},A_{t})p(A_{t+1})p(O_{t}|S_{t},A_{t})\beta(S_{t+1},A_{t+1})
 \label{beta}
 \end{align}
With the associated base case form:
\begin{align}
    \beta(S_T,A_T) 
    &= f(r(S_T,A_T)), \quad
    \alpha(S_1,A_1) 
    \propto \expectation_{S_0,A_0} [f(r(S_0,A_0))]
    \label{basecase_alpha}
\end{align}
 Since potential based functions for reward shaping are defined only on the state space, we will marginalize the actions and use the marginalized messages defined as $\alpha(S_t)$ and $\beta(S_t)$. Given that the forward-backward messages carry meaningful quantities with respect to the rewards in an MDP, they are well-suited to be used as potential functions:
\begin{align}
    F(S_t,S_{t+1}) = \gamma \Phi_{\alpha\beta}(S_{t+1}) - \Phi_{\alpha\beta}(S_t)
    \label{messpass}
\end{align}
where $\Phi_{\alpha\beta}(S_t) = \alpha(S_t)\beta(S_t) \propto p(O_{0:T}|S_t)$. The potential function  then represents the probability of optimality for a \textit{whole} trajectory, given a state. That is, they represent the probability that a state lies in the path of a high-return trajectory. This again is in contrast to the usual value function which considers only future rewarding states, given a certain state.

Obtaining the messages themselves, however, is a challenging task as performing exact inference is only possible for a restricted class of MDPs due to its computation complexity of $O(N^2T)$ where $N$ is the number of states and $T$ is the length of a trajectory. As we are interested in a generalizable approach, we will leverage the recently introduced Graph Convolutional Networks ~\citep{DBLP:journals/corr/KipfW16,DBLP:journals/corr/DefferrardBV16}.

\paragraph{Graph Convolutional Networks.}
Graph Convolutional Networks (GCNs) have mainly been used in semi-supervised learning for labeling nodes on a graph. The datasets' labeled nodes contain information that is propagated by the GCN, leading to a probability distribution defined over all nodes. A 2-layer GCN can be expressed as:
\begin{align}
    \Phi_{GCN}(X) = \text{softmax} \big( \hat T \text{ ReLU} \big( \hat T X W^{(0)} \big) W^{(1)} \big)
    \label{2layergcn}
\end{align}
where $W^{(i)}$ is the weight matrix of layer $i$ learned by gradient descent and $X$ is the input matrix with shape $N^{nodes} \times M^{features}$. The matrix $ \hat T$ is a transition matrix defined as $\hat T = D^{-1/2} \tilde{A}D^{-1/2}$, where $A$ is the adjacency matrix with added self-connections and $D$ is the degree matrix, that is $D_{ii} = \sum_j A_{ij}$. At the core of the mechanism implemented by GCNs is the idea of spectral graph convolutions which are defined through the graph Laplacian. In the next section we highlight some of its properties through the Proto-Value Function framework \citep{Mahadevan2005,MahadevanM07}
 from the reinforcement learning literature.

\section{Proposed Method}

We propose to apply GCNs on a graph in which each state is a node and edges represent a possible transition between two states. In this graph we will propagate information about rewarding states through the message passing mechanism implemented by GCNs. The probability distribution at the output of the GCN, defined as $\Phi_{GCN}(s)$, can then be used as a potential function for potential-based reward shaping. 

To clearly define a message passing mechanism it is necessary to establish the base case and the recursive case. This is made explicit through the GCN's loss function,
\begin{align}
    \mathcal{L} = \mathcal{L}_0 + \eta \mathcal{L}_{prop}
    \label{gcnloss}
\end{align}
where $\mathcal{L}_0$ is the supervised loss used for the base case and $L_{prop}$ the propagation loss implementing the recursive case. We define the base case similarly to the forward-backward algorithm, that is, by considering the first and last states of a trajectory. Additionally, as our goal is to propagate information from rewarding states in the environment, we emphasize this information by including in the set of base cases the states where the reward is non-zero. As shown in Eq.\ref{basecase_alpha}, the value of the base case depends directly (or through expectation) on the environmental reward. When using GCNs, it is straightforward to fix the values of the base cases by considering its supervised loss, defined as the cross entropy between labels $Y$ and predictions $\hat Y$ which is written as $H(Y, \hat Y)$. To implement the base case, the supervised loss then simply takes the following form,
\begin{align*}
    \mathcal{L}_0 &= H(p(O|S),\Phi_{GCN}(S)) = \sum_{S  \in\ \mathbb{S}} p(O|S) \log\big( \Phi_{GCN}(S) \big) 
\end{align*} 
where $\mathbb{S}$ is the set of base case states.  The recursive case of the message passing mechanism is attended by the propagation loss in Eq.$\ref{gcnloss}$ defined as,
$$\mathcal{L}_{prop} = \sum_{v,w} A_{vw} || \Phi_{GCN}(X_w) - \Phi_{GCN}(X_v) || ^2$$
 where $A_{vw}$ is the adjacency matrix taken at node $v$ and $w$.
 While the recursive form of the forward-backward messages in Eq.\ref{alpha}-\ref{beta} averages the neighboring messages through the true transition matrix, the GCN's propagation loss combines them through the graph Laplacian. Moreover, the mechanism of recursion is also at the core of the GCN's structural bias. To see why, we have to consider them as a specific instance of Message Passing Neural Networks (MPNN) \citep{DBLP:journals/corr/GilmerSRVD17}.  More precisely, the messages that GCNs propagate take the form:
    $m_v = \sigma\big( W^T \sum_w \hat T_{vw} m_w)$
where $W$ is the matrix of parameters, $v$ is a specific node, $w$ are its neighbours and $m_w$ is the message from node $w$. We now motivate the choice of using the graph Laplacian as a surrogate. 
 
\paragraph{Motivations for using the graph Laplacian.}
The main approximation introduced thus far was to replace the true transition matrix by the graph Laplacian. This approximation is also at the core of the Proto-Value Function framework \citep{MahadevanM07} which addresses the problem of representation learning using spectral analysis of diffusion operators such as the graph Laplacian. Proto-value functions are defined as the eigenvectors following the eigendecomposition of the true transition matrix $P^{\pi}$ of an MDP. These vectors can then be linearly combined to obtain any arbitrary value function. However, as the transition matrix is rarely available and hard to approximate, \citet{MahadevanM07} use the graph Laplacian matrix as an efficient surrogate to perform the eigendecomposition. This choice is motivated by the fact that projections of functions on the eigenspace of the graph Laplacian produce the smoothest approximation with respect to the underlying state-space topology of the MDP \citep{Chung97a}, where smoothness is defined through the Sobolov norm \citep{NIPS2005_2871}.
As value functions are generally smooth and respect the MDP's underlying topology, the graph Laplacian is considered to be a viable surrogate. In our work, we do not aim at approximating the value function, but we argue that preserving these qualities is important in order to obtain a useful signal to accelerate learning. 



\subsection{Practical Considerations}

To implement our approach on a wide range of MDPs, the question of how to estimate the underlying graph arises. In the Proto-Value Function framework, the authors draw a large number of samples and incrementally construct the underlying graph. However, representing the whole graph is not a scalable solution as performing inference (even on GPU) would be too costly. A possible solution would be to consider discretizing the state-space. Indeed, it has been shown that animals explore their environment with the help of grid cells that activate on particular intervals \citep{PMID:5124915,moser1998functional,gridcells}. However, such an implementation would introduce a whole range of hyperparameters that would require tuning and might severely affect the final performance. 
\begin{figure}           
    \begin{subfigure}[b]{.25\columnwidth}
    \centering
        \includegraphics[width=1.\columnwidth]{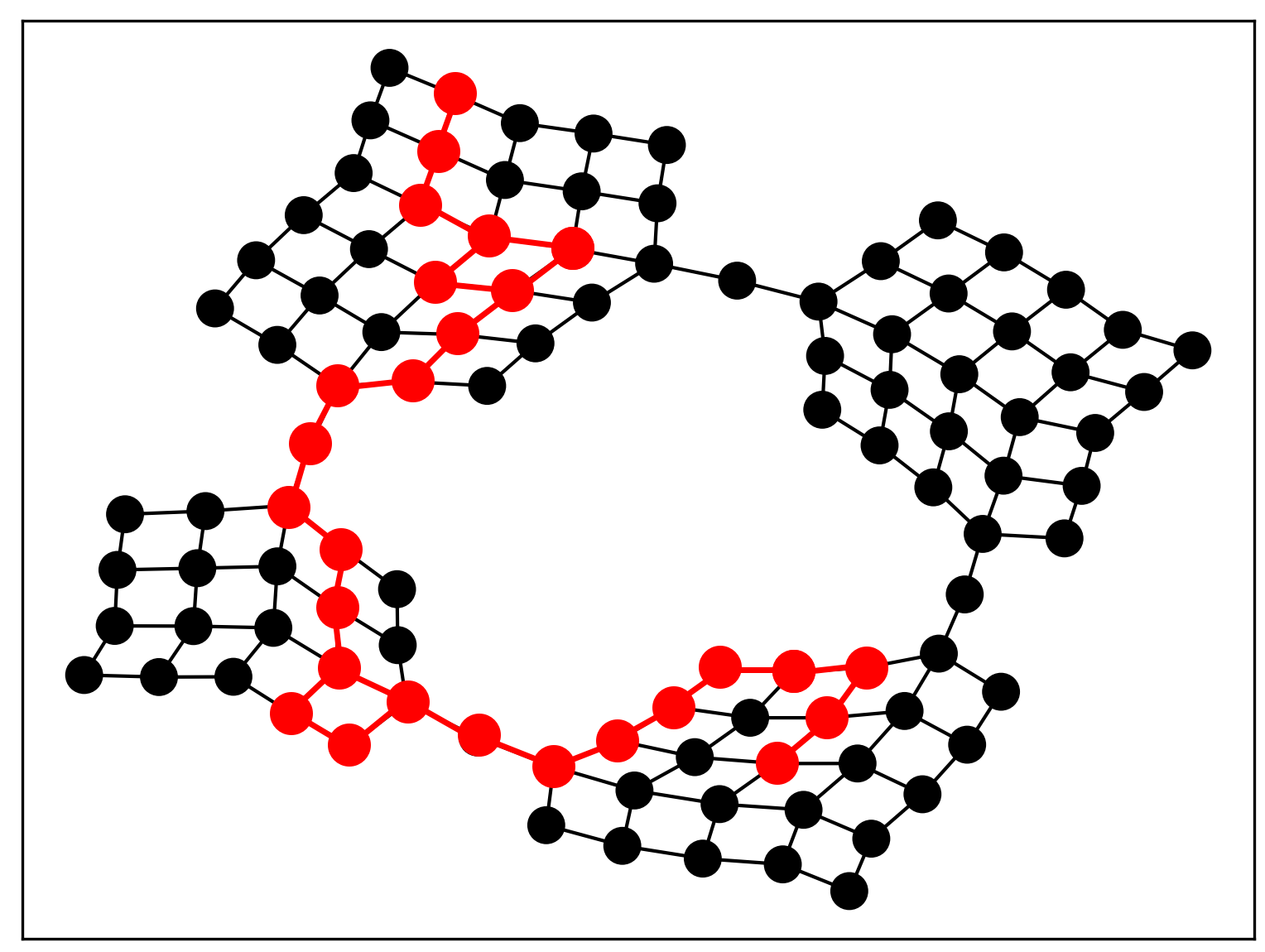}
        \caption{}
        \label{fig:samplegraph}
    \end{subfigure}   
    \quad
    \begin{subfigure}[b]{.35\columnwidth}
    \centering
    \includegraphics[width=1.\columnwidth]{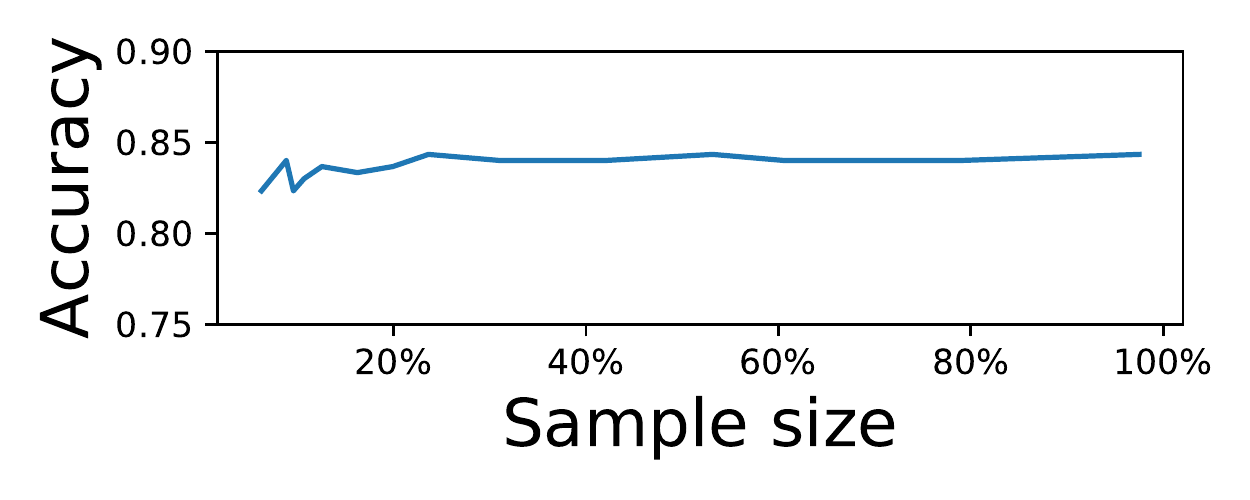}
    \caption{}
    \label{fig:valaccs}
    \end{subfigure}
    \quad
    \begin{subfigure}[b]{.35\columnwidth}
        \centering
        \begin{subfigure}[b]{1.\columnwidth}
        \centering
            \includegraphics[width=1.\columnwidth]    {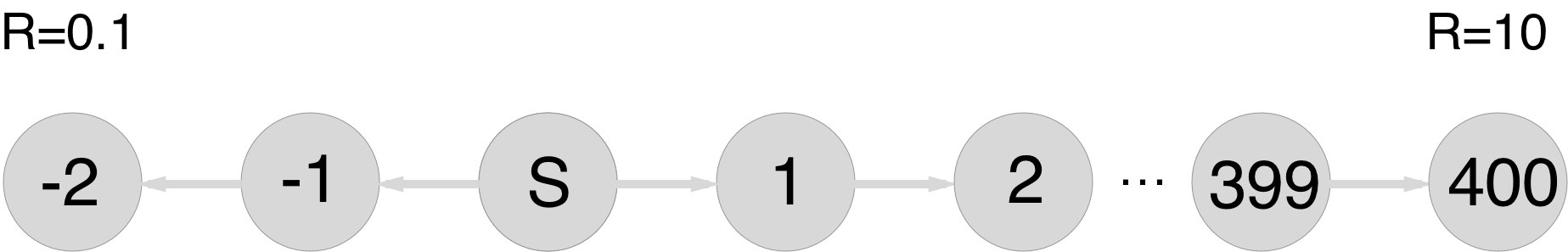}
        \end{subfigure} 
        \begin{subfigure}[b]{1.\columnwidth}
        \centering
            \includegraphics[width=1.\columnwidth]    {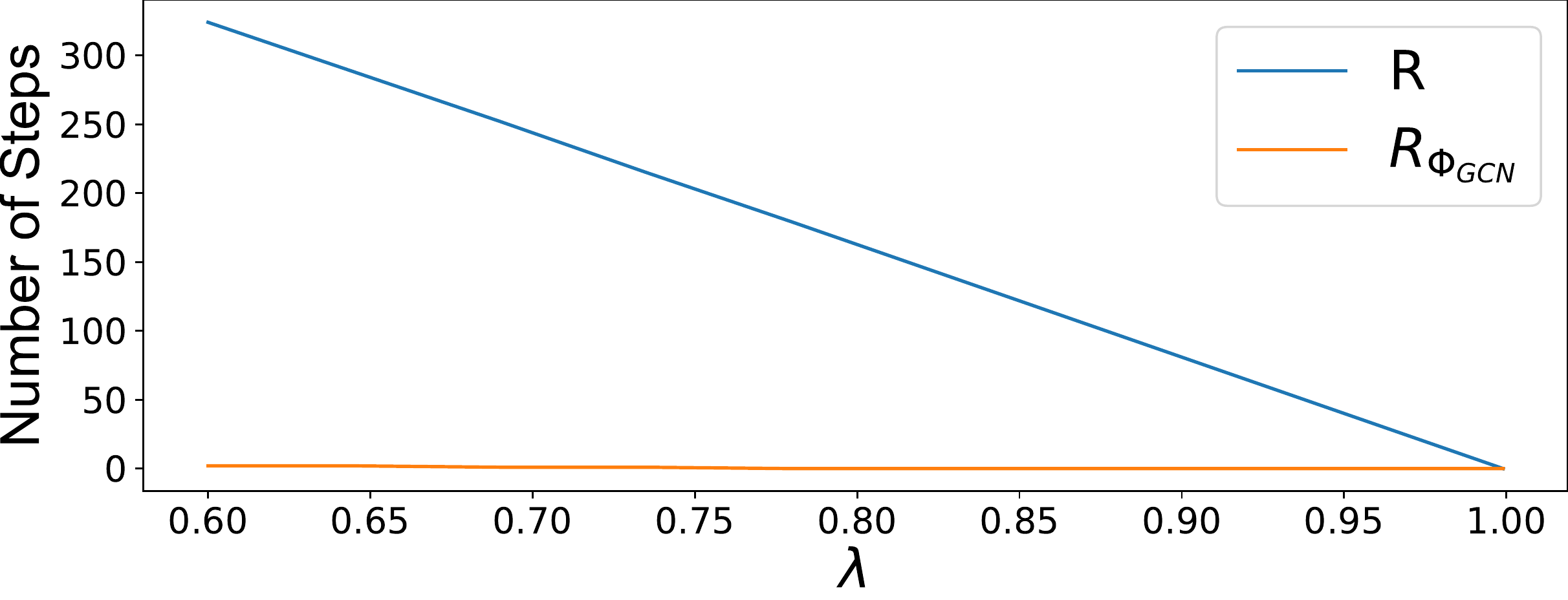}
        \end{subfigure} 
        \caption{}
        \label{fig:decays}
    \end{subfigure}
\caption{a) \textbf{Underlying graph for the FourRooms domain.} Red nodes and edges represent the sampled trajectory in the environment. b) \textbf{Validation accuracy on the Cora dataset.} We train a GCN on this dataset by only providing a certain percentage of the total training data at each iteration. Although we only provide sample graphs, the validation accuracy remains mostly constant. c) \textbf{Numbers of iterations to     convergence} between the GCN-shaped reward function  $R_{\Phi_{GCN}}$ and the original reward function $R$. }
\end{figure}

To address this, we propose a straightforward solution by choosing to approximate the underlying graph through sampled trajectories on which we train the GCN. The idea is illustrated in Fig \ref{fig:samplegraph}, where the whole graph is shown, but we have highlighted in red only a subset of states and edges corresponding to a sampled trajectory used to train the GCN at this particular iteration.

To investigate whether this sampling approach will still encompass the complexities of the whole graph, we apply the GCN on samples from the Cora dataset \citep{sen:aimag08} and evaluate its validation accuracy on the whole graph, as presented in Fig.\ref{fig:valaccs}. We notice that although we only use samples of the whole graph, the GCN is still able to implicitly encode the underlying structure and maintain a similar validation accuracy across sample size. This is made possible by learning the weights of the GCNs, making this a paramater-based approach. An important benefit of this solution is the fact that the computational overhead is greatly reduced, making it a practical solution for large or continuous MDPs. We investigate further the consequences of this approximation in Appendix \ref{app_smooth} where we show that the resulting function on states, $\Phi(s)$, will tend to be more diffuse in nature. This function over states can then potentially be leveraged by the agent to learn more efficiently. Our sampling strategy is closely related to the one employed by previous work based on the graph Laplacian  ~\citep{MachadoBB17,eigenoptiondisco}, although with the important difference that we do not proceed to the eigen-decomposition of the transition matrix.

\subsection{Illustration}

To illustrate the benefits offered by forming potential based functions through the GCN's propagation mechanism, we consider a toy example depicted in Fig.\ref{fig:decays} where the agent starts in the state S. It can then choose to go left and obtain a small reward of 0.1 after 2 steps or go right where after 400 steps it gets a reward of 10. The optimal policy in case is to go right. In this toy example we update both actions at each iteration avoiding any difficulties related to exploration. The targets used to update the action-value function are the $\lambda$-returns. In Fig.\ref{fig:decays} we plot the number of iterations required to converge to the optimal policy as a function of the $\lambda$ parameter.

We notice that, in the case of the original reward function (denoted as $R$), the number of steps required to converge depends linearly on the value of $\lambda$, whereas the potential shaped reward function (denoted as $R_{\Phi}$) is mostly constant. The only value for which both methods are equal is when $\lambda=1$. However, in practical settings, such high vales of $\lambda$ lead to prohibitively high variance in the updates. The observed difference between the two approaches has previously been investigated more rigorously by \cite{dejong}. The authors show that the number of decisions it takes to experience accurate feedback, called the \textit{reward horizon}, directly affects how difficult it is to learn from this feedback.
When using potential based reward shaping, the reward horizon can then be scaled down, reducing the learning difficulty. However, as our approach learns the potential function by experiencing external rewards, its purpose is not to improve exploration.
Instead, our approach can be understood as an attempt to accelerate learning by emphasizing information about rewarding states through potential functions with the guarantee of preserving the optimal policy.

\subsection{Algorithm}

We now present our implementation of the ideas outlined above in the policy gradient framework.\footnote{The ideas are general and could also be used with Q-learning or other policy-based methods.} We define two kinds of action value functions that will be used: the original function, $Q^{\pi}(s,a) = \expectation {\big[\sum_{t} \gamma^t r(S_t,A_t)\big]}$ and the reward-shaped function, $ Q^{\pi}_{\Phi}(s,a) = \expectation \big[ \sum_{t} \gamma^t ( r(S_t,A_t) + \gamma \Phi_{GCN}(S_{t+1}) - \Phi_{GCN}(S_t) ) \big]$. We combine them though a scalar $\alpha$ as in $Q^{\pi}_{comb}= \alpha  Q^{\pi}(s,a) + (1-\alpha) Q^{\pi}_{\Phi}(s,a)$.  Algorithm \ref{algo_box} then describes the end-to-end training approach. Potential-based reward shaping in the context of policy-based approaches has interesting connections to the use of a baseline \citep{Schulmanetal_ICLR2016}. In our approach, we can use the identity $Q^{\pi}(s,a) - \Phi(s) = Q^{\pi}_{\Phi}(s,a)$ to notice that the resulting baseline is simply the potential function at a given state. In general, it is hard to evaluate the benefits of using a particular baseline, although it is possible to obtain bounds on the resulting variance as shown in \cite{greensmith}. Adopting their analysis, we show in Appendix \ref{sec:app_baseline} that the highest upper bound is less or equal than the one obtained by the value function $V^{\pi}(s)$.

\begin{algorithm}[h]
\DontPrintSemicolon
\SetAlgoLined
Create empty graph $G$\\
\For{Episode=0,1,2,....}{
    \For{t=1,2...T}{
    Add transition $(S_{t-1},S_{t})$ to graph $G$
    }
    \uIf{mod(Episode,N)}{
    Train the GCN on the approximate graph.
    }
    $Q^{\pi}_{comb}= \alpha  Q^{\pi} + (1-\alpha) Q^{\pi}_{\Phi}$\\
    \text{Maximize } $E_{\pi} \big[ \grad{\log \pi(A_t|S_t)} Q^{\pi}_{comb}(S_t,A_t)  \big] $\\
Reset $G$ to empty graph (optional)\;
}
\caption{Reward shaping using GCNs}
\label{algo_box}
\end{algorithm}

\section{Experiments and Results}
\subsection{Tabular}
\begin{figure}[h]
    \centering
    \begin{subfigure}[c]{0.17\columnwidth}
    \centering
        \includegraphics[width=1.\columnwidth]{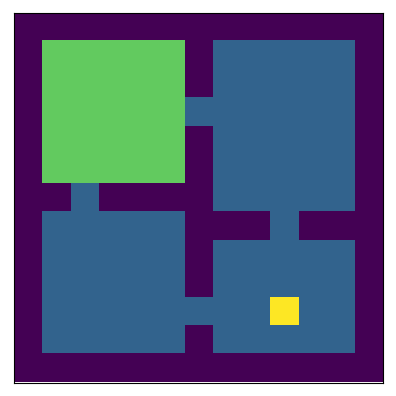}
    \end{subfigure}
    \begin{subfigure}[c]{0.27\columnwidth}
    \centering
        \includegraphics[width=1.\columnwidth]{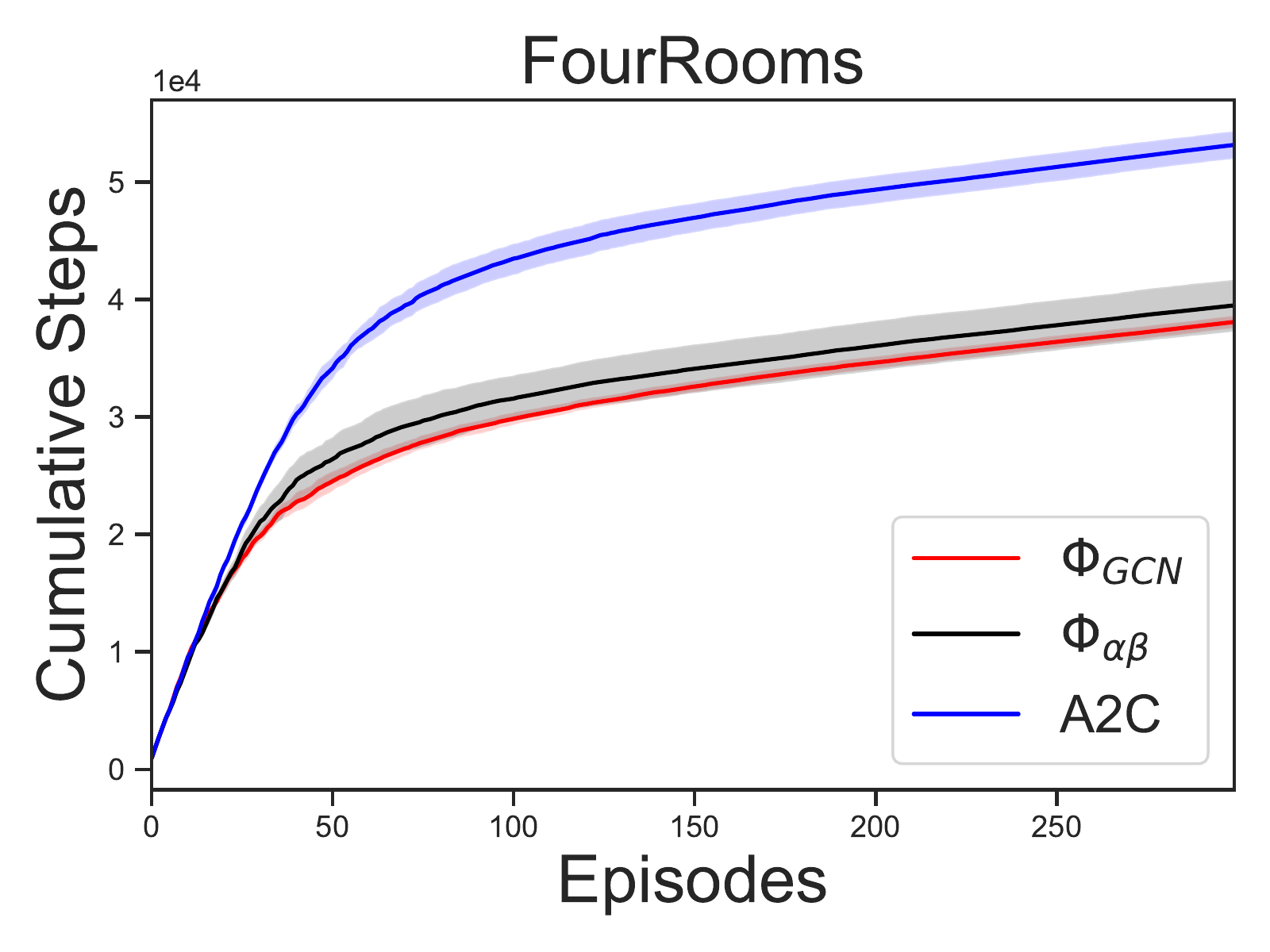}
    \end{subfigure} 
    \quad
    \begin{subfigure}[c]{0.17\columnwidth}
    \centering
        \includegraphics[width=1.\columnwidth]{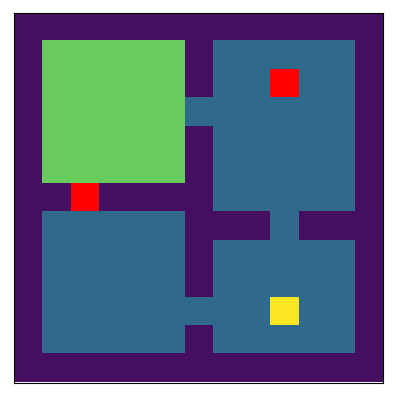}
    \end{subfigure}
    \begin{subfigure}[c]{0.27\columnwidth}
    \centering
        \includegraphics[width=1.\columnwidth]{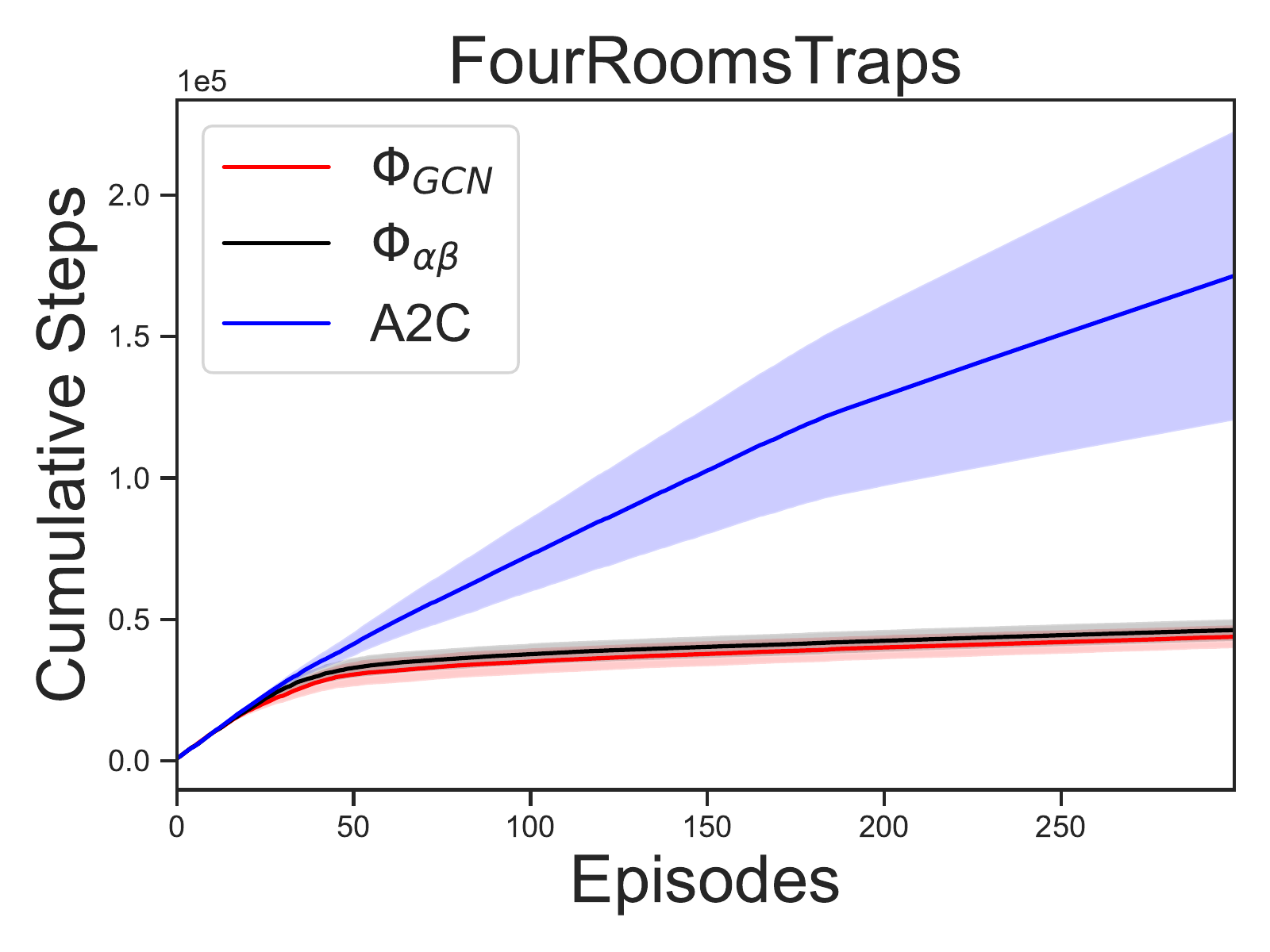}
    \end{subfigure} 
    \caption{\textbf{Tabular environments.} Results are presented for the classic FourRooms domain and a variant called FourRoomsTraps. In both cases the agent starts in a random position inside the green square and has to get to the yellow goal, while avoiding red regions.}
    \label{fig:tab}
\end{figure}

\textit{Experimental Setup:} We perform experiments in the tabular domains depicted in Fig.~\ref{fig:tab} where we explore the classic FourRooms domain and the FourRoomsTraps variant where negative rewards are scattered through the rooms. The results are presented in the form of  cumulative steps (i.e. regret). In these experiments we add noise in the action space (there is a 0.1 probability of random action) in order to avoid learning a deterministic series of actions. We compare our approach, denoted $\Phi_{GCN}$, to a actor-critic algorithm using $\lambda$-returns for the critic, denoted A2C. 

Our approach shows significant improvements over the classic actor-critic algorithm, which suggests that our method is able to provide the agent with valuable information to accelerate learning. As the adjacency matrix can be fully constructed, we also compare our approach to a formal implementation of the forward-backward algorithm in order to propagate reward information through the RL graph (as in Eq. \ref{messpass}) denoted as $\Phi_{\alpha \beta}$. In Fig.\ref{fig:tab}, we notice that both message passing mechanisms produce very similar results in terms of performance. We also provide  illustrations of $\Phi_{GCN}$ and $\Phi_{\alpha \beta}$ in Appendix \ref{sec:tab} where we show very similar distributions over states and where we include all the values of the hyperparameters.

\subsection{MiniWorld}
 
\begin{figure}[ht]
    \centering
    \begin{subfigure}[c]{0.3\columnwidth}
    \centering
        \includegraphics[width=1.\columnwidth]{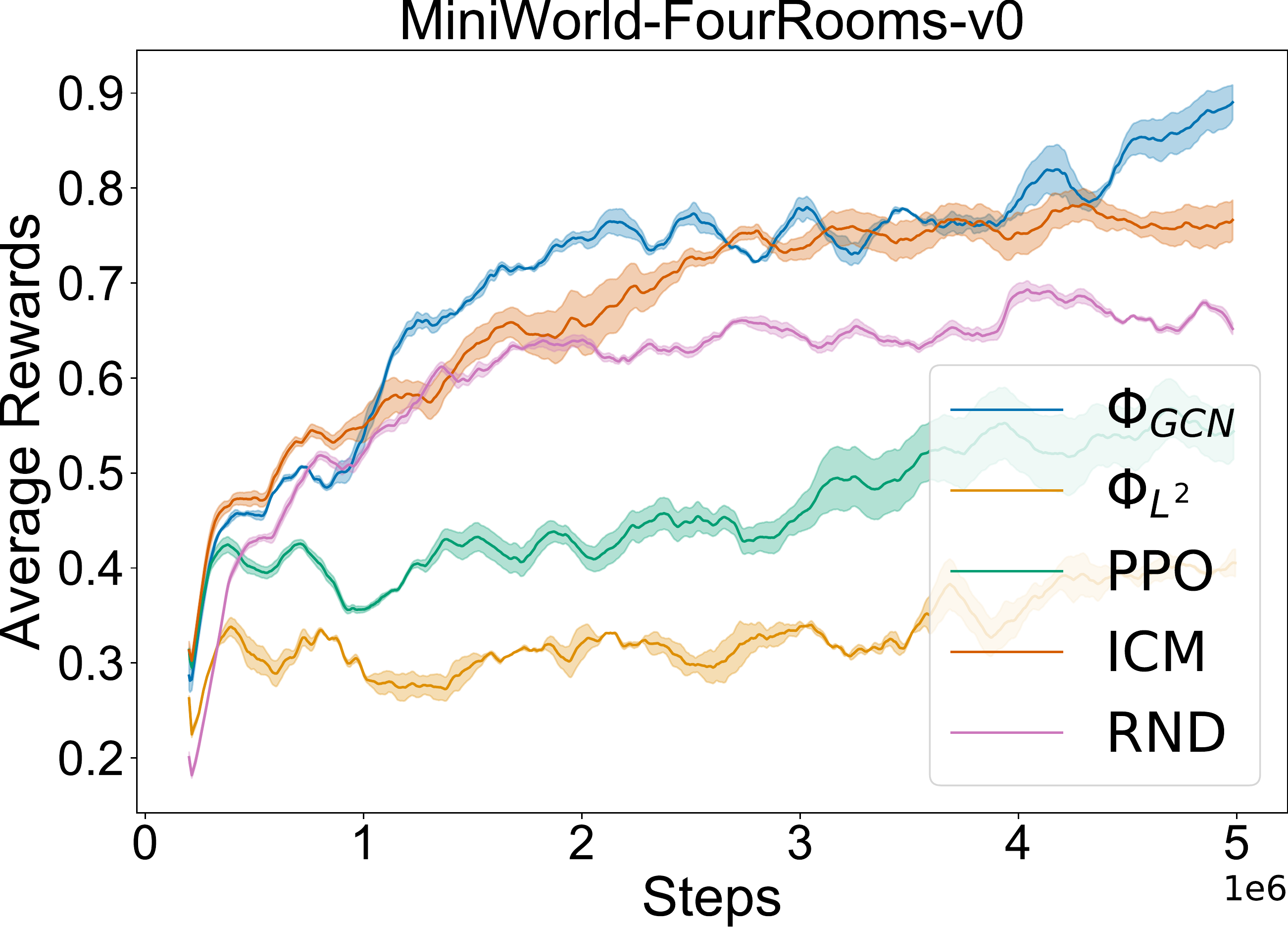}
    \end{subfigure} 
    \begin{subfigure}[c]{0.3\columnwidth}
    \centering
        \includegraphics[width=1.\columnwidth]{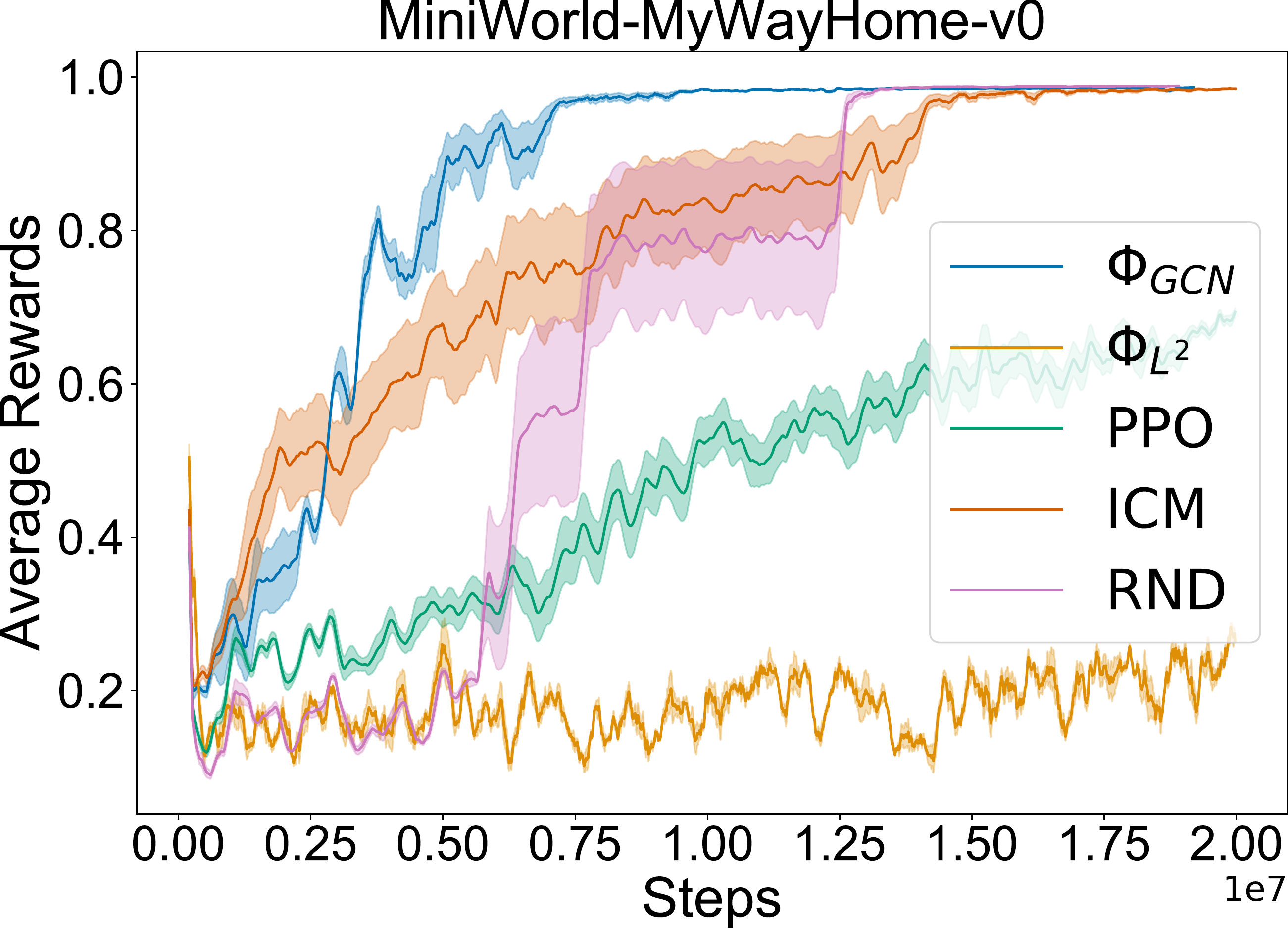}
    \end{subfigure} 
    \begin{subfigure}[c]{0.3\columnwidth}
    \centering
        \includegraphics[width=1.\columnwidth]{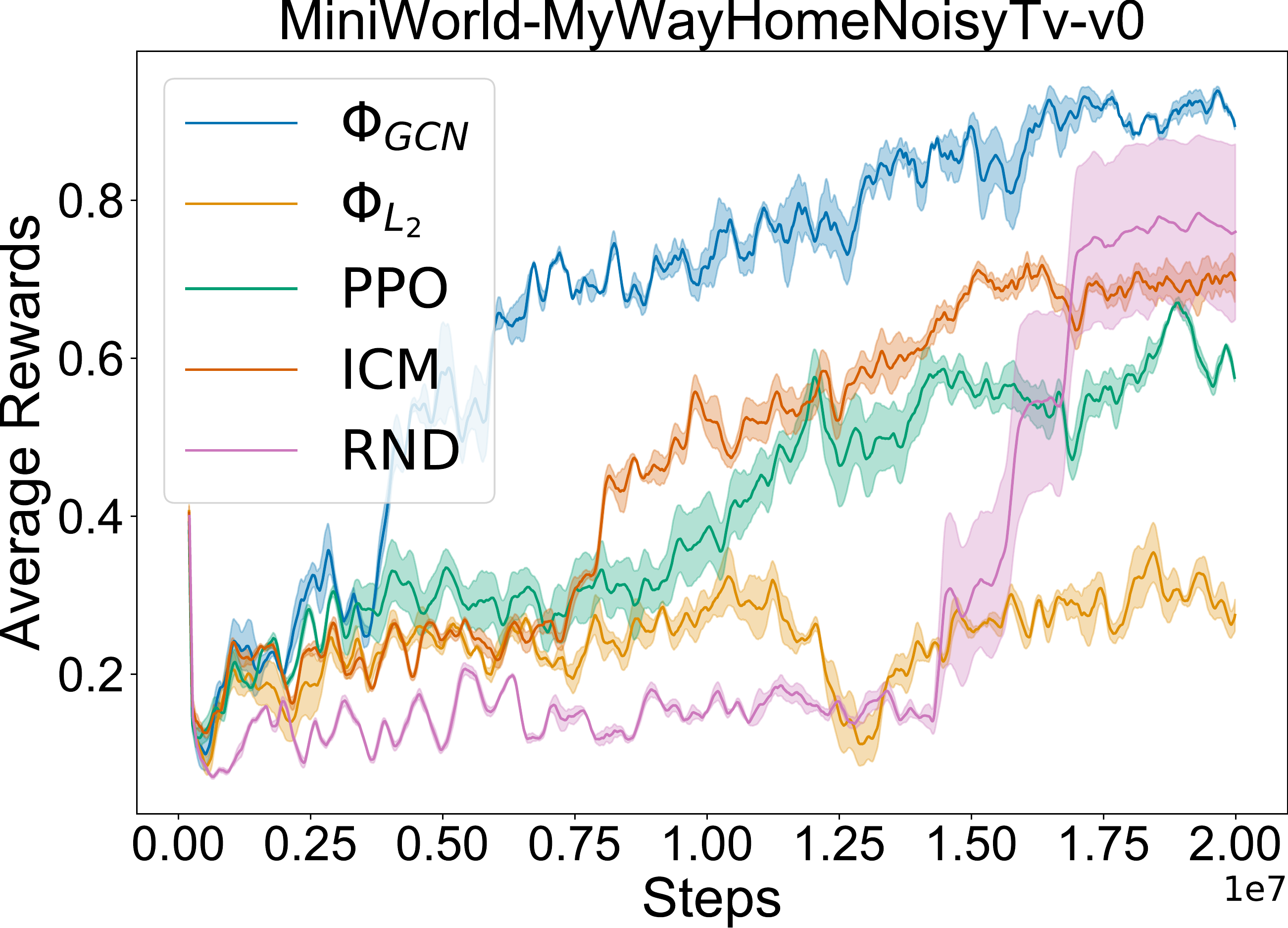}
    \end{subfigure} 
    \begin{subfigure}[c]{0.22\columnwidth}
    \centering
        \includegraphics[width=1.\columnwidth]{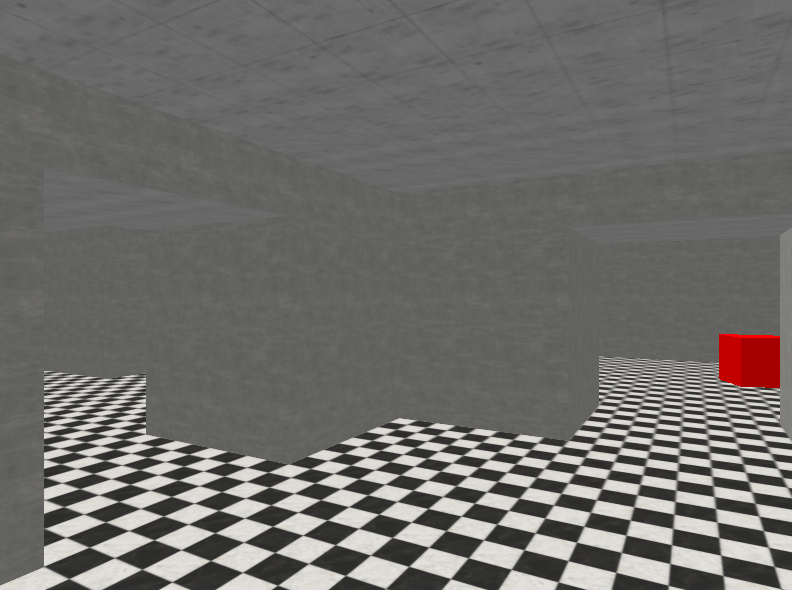}
    \end{subfigure}
    \hspace{30pt}
    \begin{subfigure}[c]{0.22\columnwidth}
    \centering
        \includegraphics[width=1.\columnwidth]{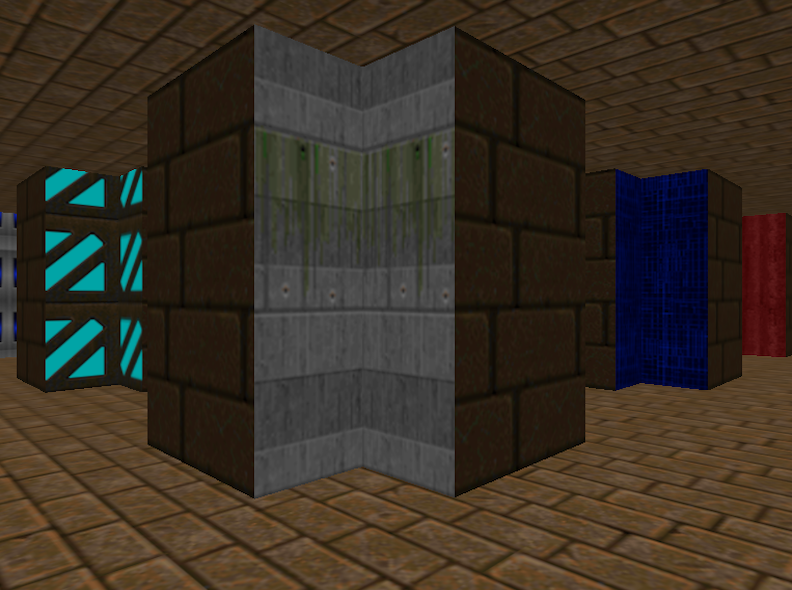}
    \end{subfigure}
    \hspace{30pt}
    \begin{subfigure}[c]{0.22\columnwidth}
    \centering
        \includegraphics[width=1.\columnwidth]{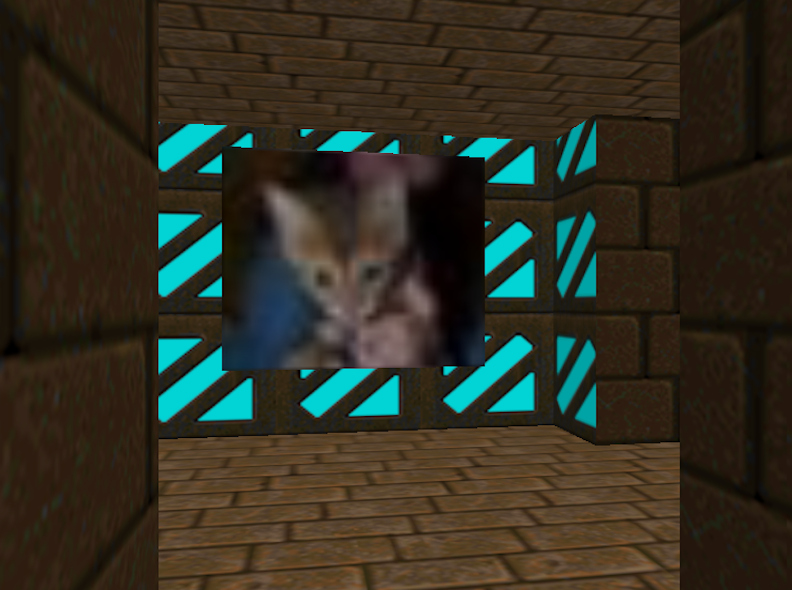}
    \end{subfigure}
    \caption{\textbf{High dimensional control.}
    In all environments, the results show the difficulty of constructing a helpful potential function through the $\Phi_{L^2}$ baseline. Our approach on the other hand almost doubles the PPO baseline. We also present other approaches using reward shaping where the aim is to improve exploration.
    }
    \label{fig:navi}
\end{figure}

\textit{Experimental Setup:}
We further study how  we can improve performance in more complex environments where hand-designed potential functions are hard to scale. We work with the MiniWorld \citep{gym_miniworld} simulator and explore a vision-based variant of the classic FourRooms domains called MiniWorld-FourRooms-v0 in which the agent has to navigate to get to the red box placed at a random position throughout the rooms. We also experiment with MiniWorld-MyWayHome-v0 which is the analogue of the challenging Doom-MyWayHome-v0 \citep{DBLP:journals/corr/KempkaWRTJ16}. The goal is to navigate through nine rooms with different textures and sizes to the obtain a reward. Finally, the last environment is MiniWorld-MyWayHomeNoisyTv-v0, a stochastic variant on MiniWorld-MyWayHome-v0 inspired by \cite{burda} that introduces a television activated by the agent that displays random CIFAR-10 images \citep{cifar10}. 
In all environments we use Proximal Policy Optimization \citep{SchulmanWDRK17} for the policy update. All details about hyperparameters and network architectures are provided in the Appendix \ref{sec:control}. However, it is important to note that through all these experiments we use the same values for the hyperparameters, making it a general approach. For all the experiments presented in Fig. \ref{fig:navi} we add noise in the action space and randomize the agent's starting position to avoid deterministic solutions.

To highlight the difficulty of learning potential functions in the environments we investigate, we provide an additional baseline, $\Phi_{L^2}$, that naively implements the $L^2$ distance between a state and the goal and leverage it as the potential function. For completeness, we also provide two exploration baselines based on reward shaping: the random network distillation approach (RND) \citep{rnd} and the intrinsic curiosity module (ICM) \citep{pathakICMl17curiosity}. However, we emphasize that our approach is not meant to be a solution for exploration, but instead aims at better exploiting a reward signal in the environment. 

The results in Fig.~\ref{fig:navi} show that in all environments, the naive reward shaping function does not help performance while our approach almost doubles the baseline score. We also notice that compared to the exploration baseline we learn faster or achieve better final performance. The greatest difference with exploration approaches is shown in the MiniWorld-MyWayHomeNoisyTv-v0 task, where exploration approaches are less robust to stochasticity in the environment. This again highlights a major difference of our approach: it does not necessarily seek novelty. Moreover, we emphasize that potential-based reward shaping is the only approach that guarantees invariance with respect to the optimal policy.

An important hyperparameter in our approach is effectively $\alpha$ which trades-off between the reward shaped return and the default return. In Fig \ref{fig:alpha} of the Appendix \ref{sec:alpha} we show the results obtained on MiniWorld-FourRooms-v0 across all values of $\alpha$. We also investigate the role of $\eta$, the hyperparameter trading-off between the two losses of the GCN, in Appendix \ref{eta_hyp} and  provide evidence that it controls the bias-variance trade-off of the propagation process.

\subsection{Atari 2600}
\textit{Experimental Setup:}  The Atari benchmark is relevant to investigate as it includes a wide variety of games that range from reactive games to hard exploration games. We showcase the applicability of our method by running experiments over 40M frames on 20 Atari games that exhibit such variety. We include sticky actions to avoid deterministic solutions \citep{DBLP:journals/corr/abs-1709-06009} and use the same values of hyperparameters across all games (details are in Appendix \ref{app:atar}).
 
\begin{figure}[ht]
    \centering
    \begin{subfigure}[c]{0.22\columnwidth}
    \centering
        \includegraphics[width=1.\columnwidth]{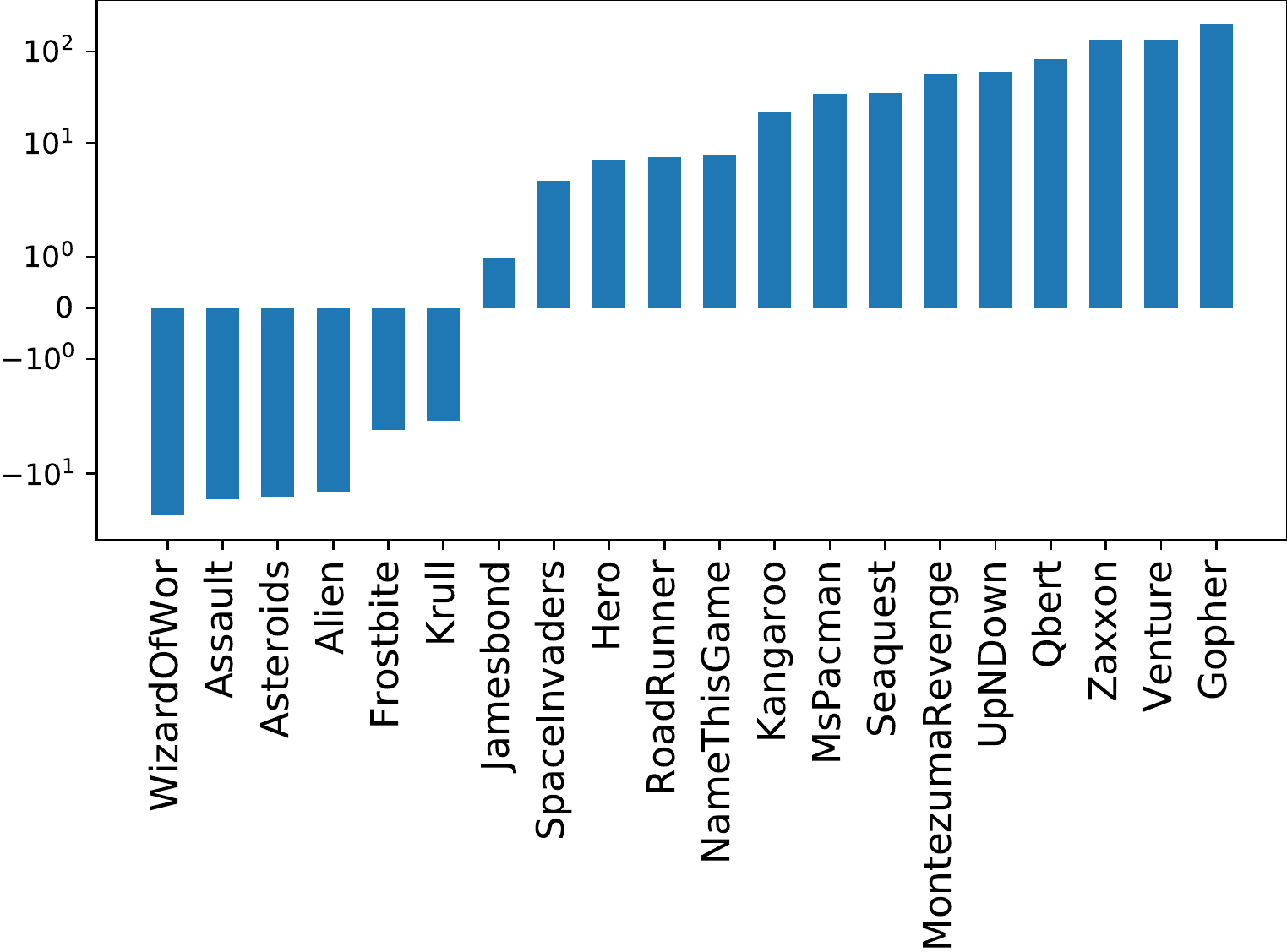}
\caption{$\Phi_{GCN}$}
\label{phi}
    \end{subfigure} 
    \begin{subfigure}[c]{0.22\columnwidth}
    \centering
        \includegraphics[width=1.\columnwidth]{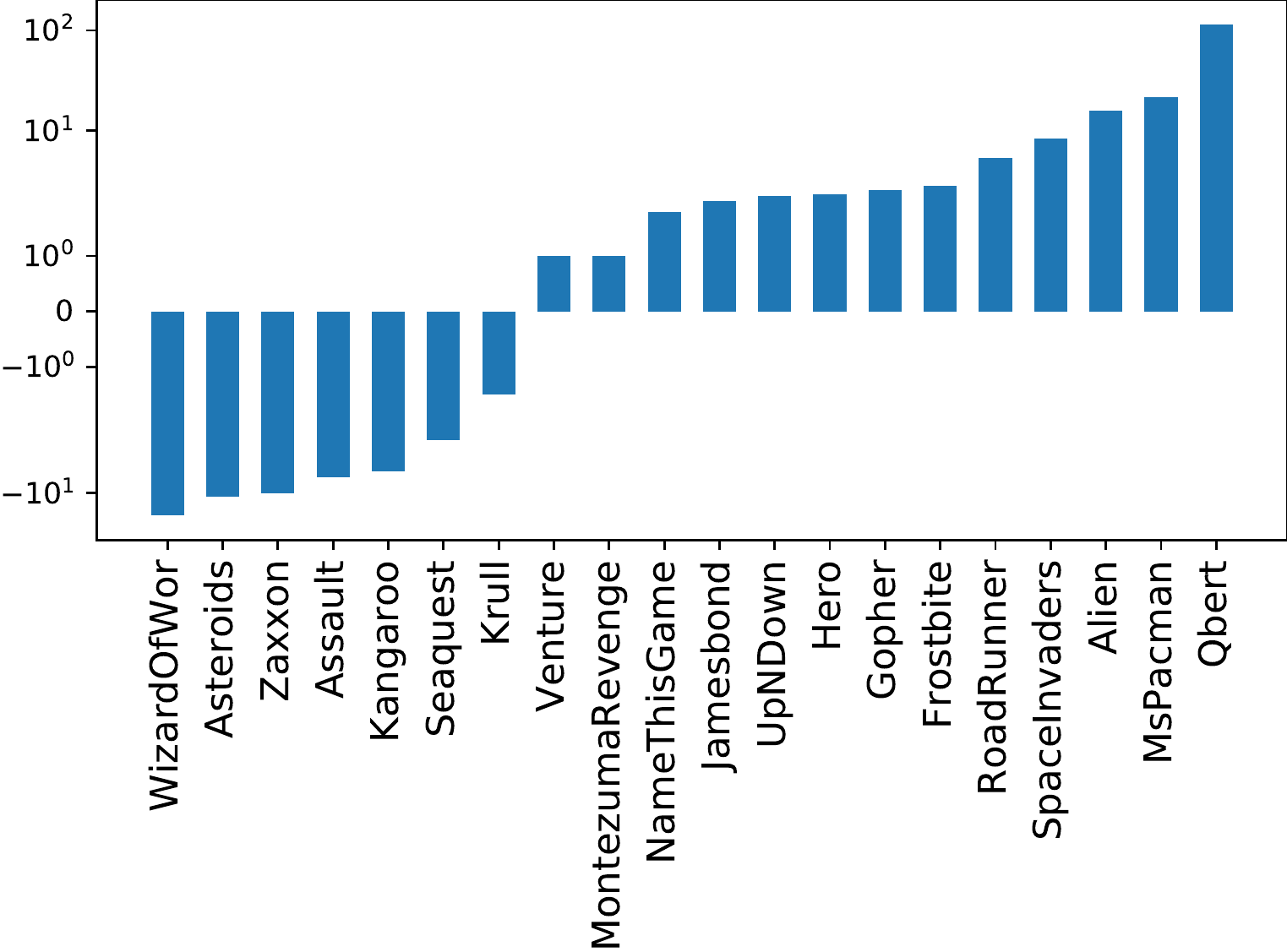}
\caption{ICM}
\label{icm}
    \end{subfigure} 
    \begin{subfigure}[c]{0.22\columnwidth}
    \centering
        \includegraphics[width=1.\columnwidth]{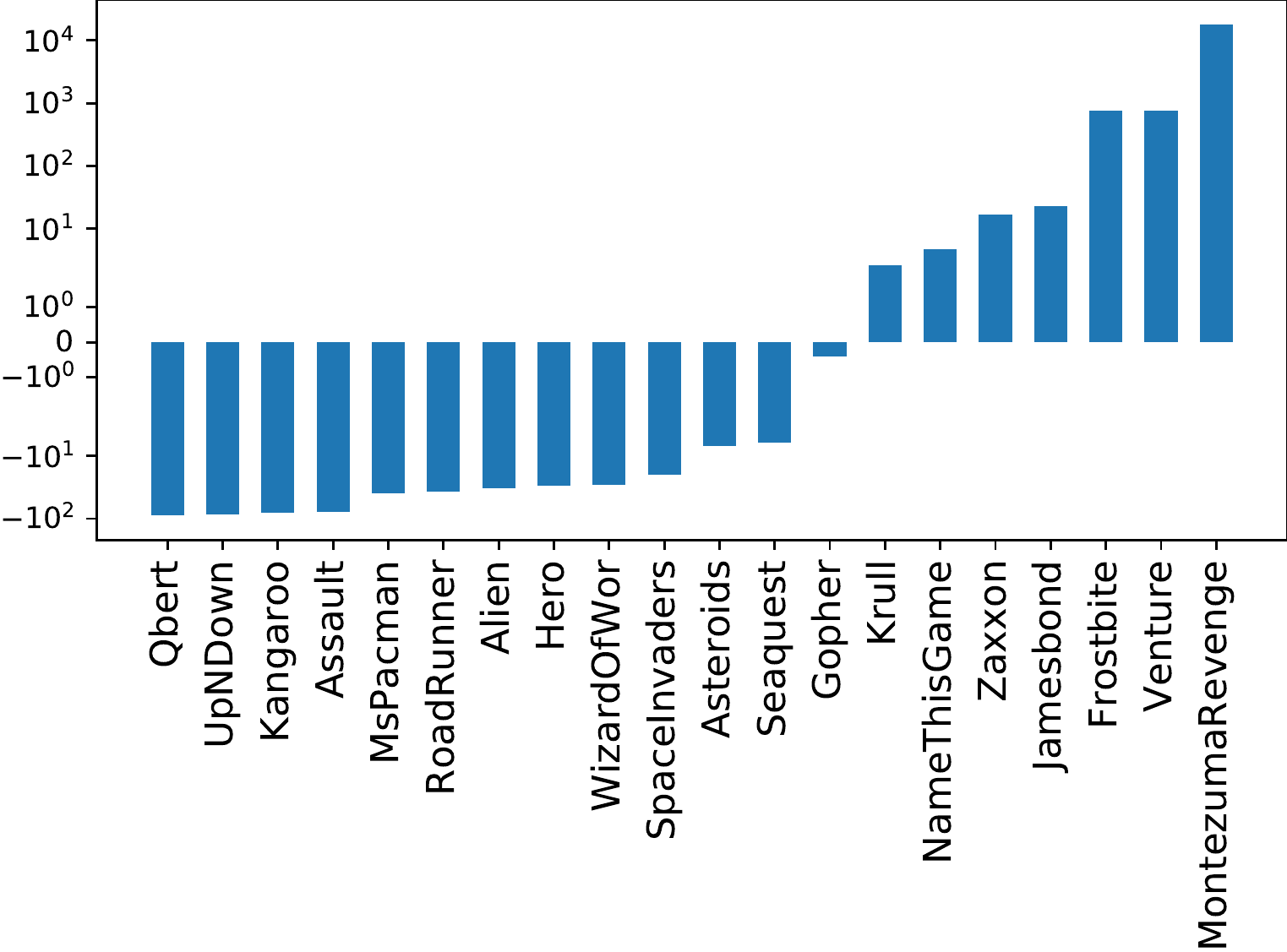}
    \caption{RND}
    \end{subfigure} 
    \begin{subfigure}[c]{0.22\columnwidth}
    \centering
        \includegraphics[width=1.\columnwidth]{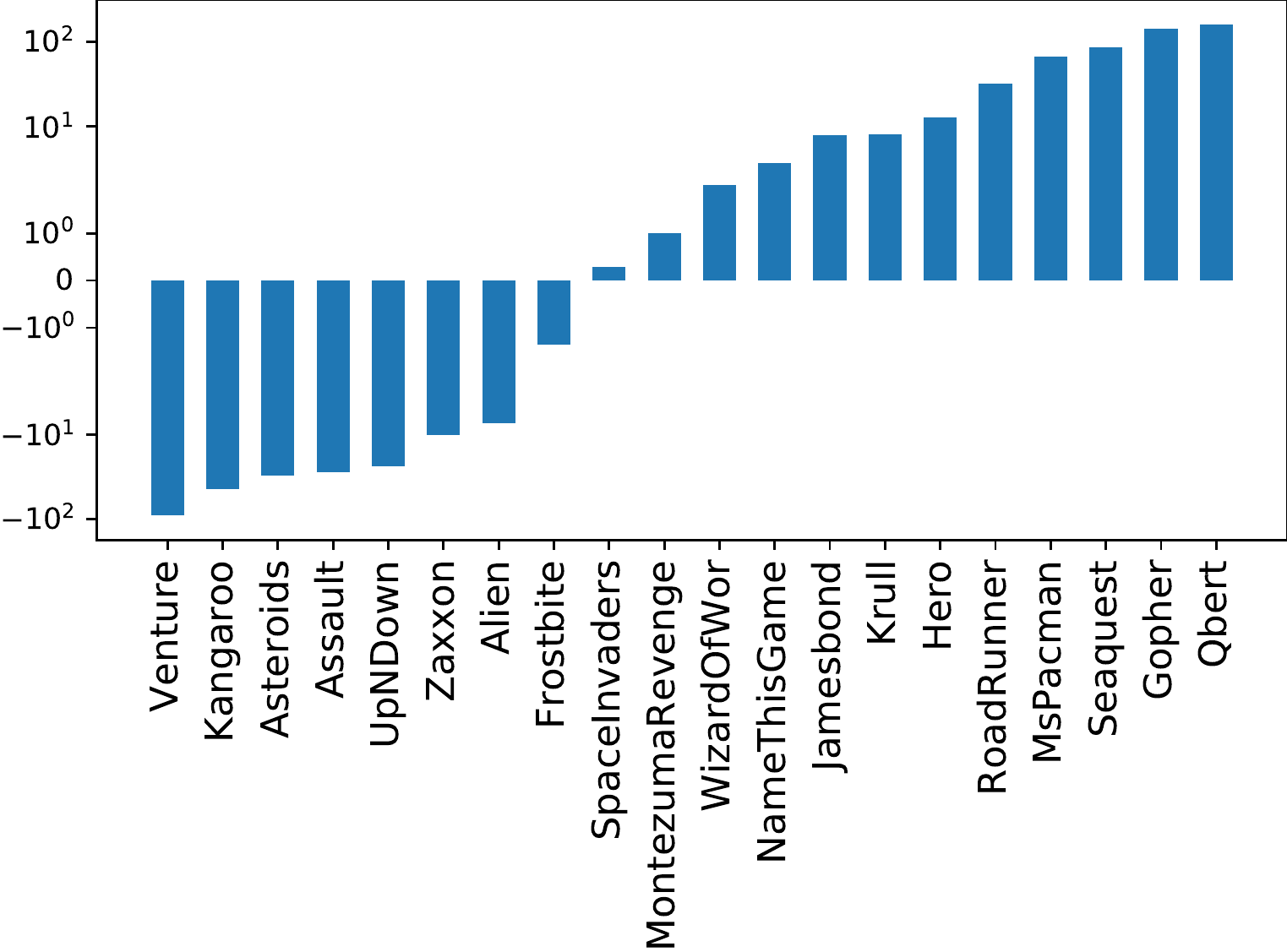}
    \caption{LIRPG}
    \end{subfigure} 
    \caption{\textbf{Performance improvement} in log scale over the PPO baseline where we compare $\Phi_{GCN}$ to the intrinsic curiosity module (ICM), Random Network Distillation (RND) and Learning Intrinsic Rewards for Policy Gradient (LIRPG) .}
    \label{fig:atari}
\end{figure}
\begin{wraptable}{r}{0pt}
      \small
          \centering
        \begin{tabular}[b]{cc}\hline
            Method & FPS \\ \hline
            PPO & 1126 \\
            $\Phi_{GCN}$ & 1054 \\
            RND & 987 \\
            ICM & 912 \\
            LIRPG & 280 \\ \hline
          \end{tabular}
         \caption{Frames-Per-Second (FPS) on Atari games}
          \label{tab:fps}
\end{wraptable}
As in the MiniWorld experiments, we compare our approach to two exploration baselines, that is ICM and RND, and present in Fig. \ref{fig:atari} the percentage of improvement over the PPO algorithm. Additionally, we also compare with  Learning Intrinsic Rewards for Policy Gradient (LIRPG) \citep{lirpg}. This baseline is more closely related to our approach in the sense that it does not specifically aim to improve exploration but still seeks to improve the agent's performance. Note however that only our approach is guaranteed invariance with respect to the optimal policy as we are building on the potential-based reward shaping framework while LIRPG builds on the optimal rewards framework \citep{singh}. 

We notice that in most games $\Phi_{GCN}$ shows good improvements especially in games such as Gopher or Zaxxon. Moreover, we notice that our approach is more consistent in improving the policy's performance as compared to the other approaches. Interestingly, the RND approach provides great gains on hard exploration games such as Venture, but otherwise reduces dramatically the score in a range of games. Finally, in Table \ref{tab:fps} we also compare the run-time of all the presented approaches. We did these evaluations on a single V100 GPU, 8 CPUs and 40GB of RAM. The time taken, in frames-per-second (FPS), for our approach $\Phi_{GCN}$ is very similar  to the PPO baseline, only slightly slower. We also compare favourably with respect to the RND, ICM, and LIRPG baselines. We believe this good performance stems directly from our sampling strategy that is minimal yet effective.

\subsection{MuJoCo}
\textit{Experimental Setup:}  To further investigate the applicability of our method, we perform experiments on environments with continuous states and actions by working with the MuJoCo benchmark \citep{conf/iros/TodorovET12}. To make the learning more challenging, we experiment with the delayed version of the basic environments where the extrinsic reward is rendered sparse by accumulating it over 20 steps before it is being provided to the agent. All values of hyperparameters are provided in Appendix \ref{app:muj}.

\begin{figure}[ht]
    \centering
    \begin{subfigure}[c]{0.23\columnwidth}
    \centering
        \includegraphics[width=1.\columnwidth]{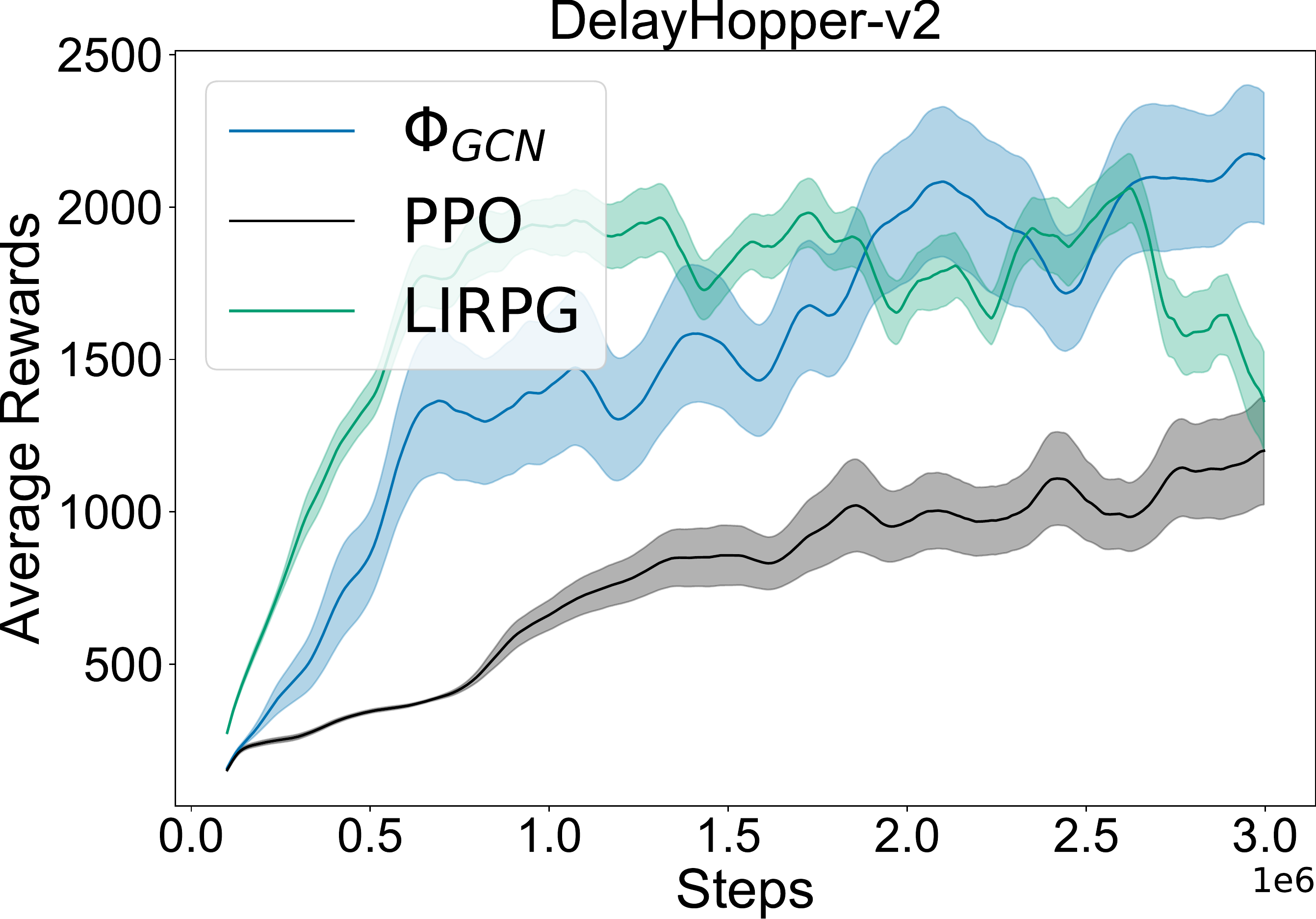}
    \caption{DelayHopper-v2}
    \end{subfigure} 
    \begin{subfigure}[c]{0.23\columnwidth}
    \centering
        \includegraphics[width=1.\columnwidth]{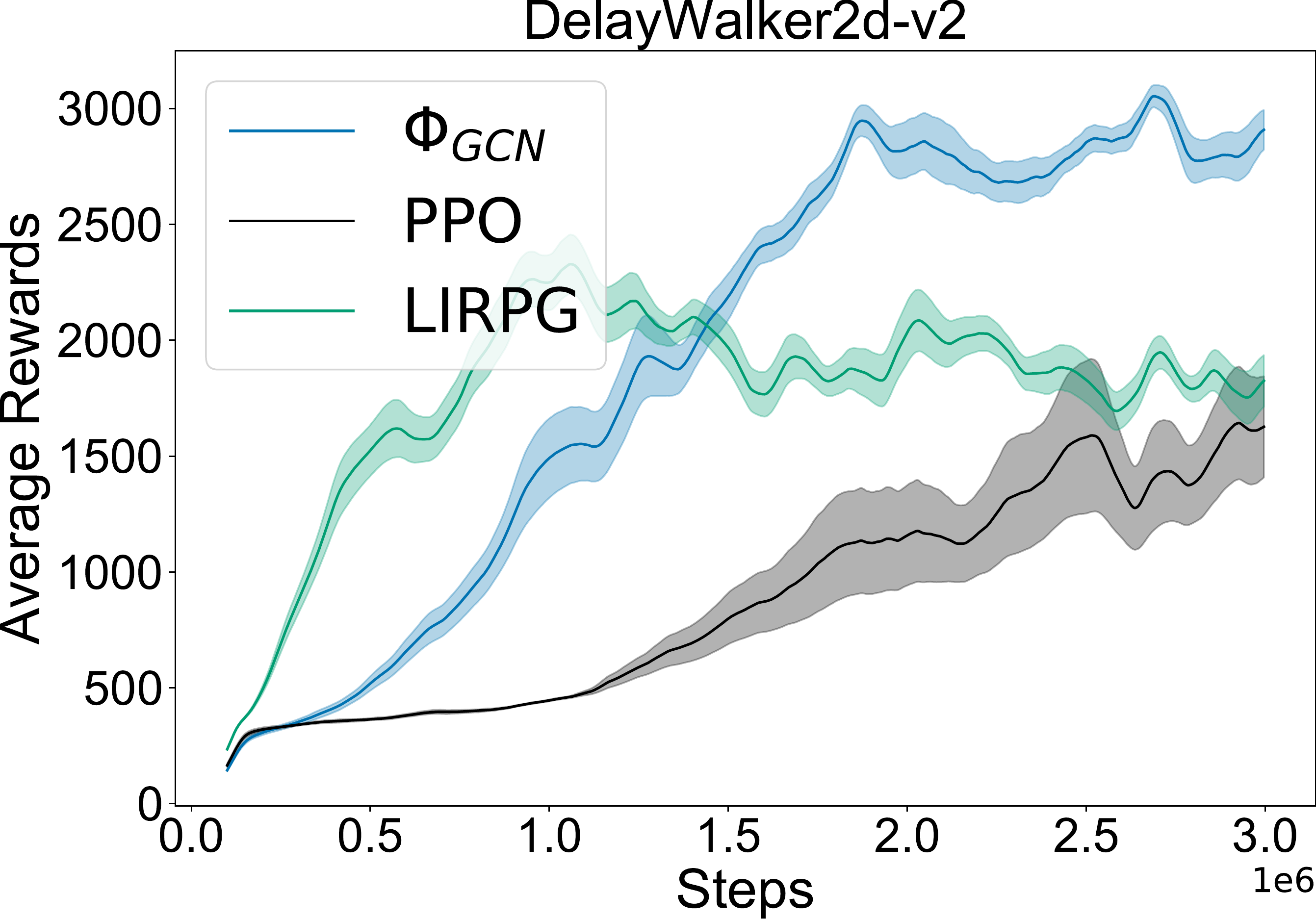}
    \caption{DelayWalker2d-v2}
    \end{subfigure} 
    \begin{subfigure}[c]{0.23\columnwidth}
    \centering
        \includegraphics[width=1.\columnwidth]{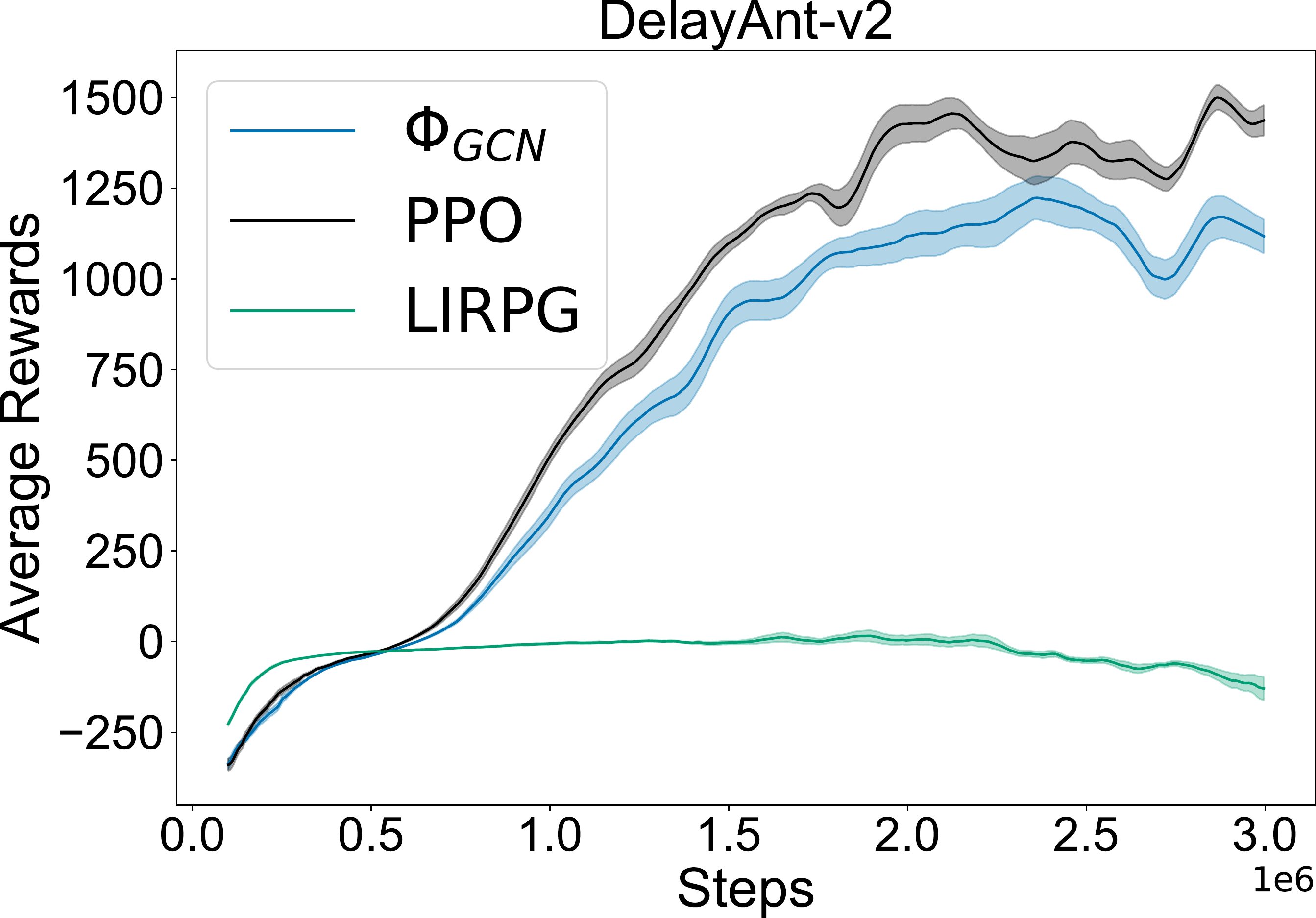}
    \caption{DelayAnt-v2}
    \end{subfigure} 
    \begin{subfigure}[c]{0.23\columnwidth}
    \centering
        \includegraphics[width=1.\columnwidth]{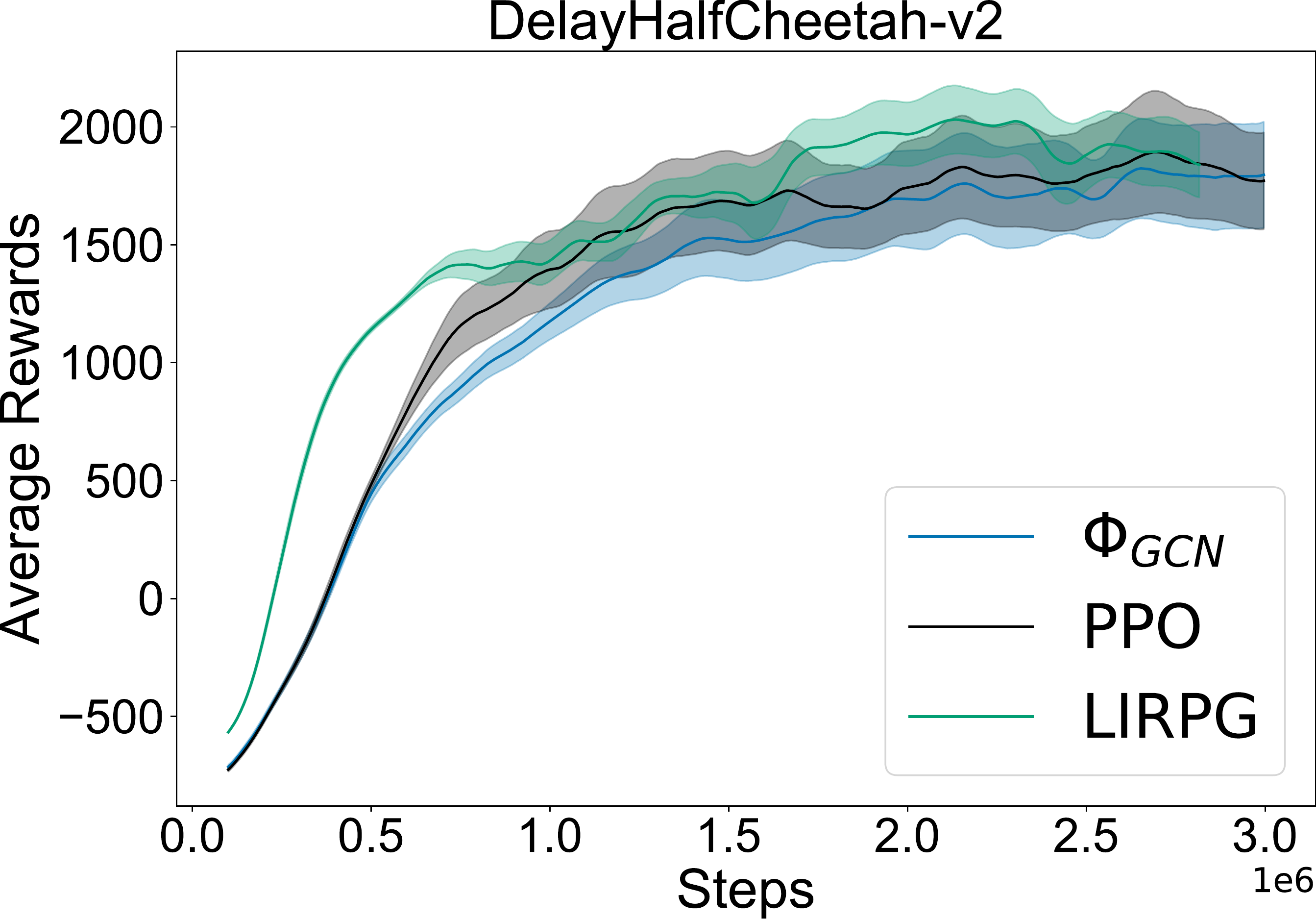}
    \caption{DelayHalfCheetah-v2}
    \end{subfigure} 
    \caption{\textbf{Continuous control.} In most environments we see significant improvements over the PPO baseline. We also compare favorably to the Learning Intrinsic Rewards for Policy Gradient (LIRPG) baseline.}
    \label{fig:mujo}
\end{figure}

In Fig. \ref{fig:mujo} we witness that our approach still provides significant improvements over the PPO baseline. We also compare to the LIRPG approach and notice that although $\Phi_{GCN}$ seems to learn a bit slower in the first iterations, the final performance presents a clear picture. We do not provide a comparison to ICM and RND as the former  has been formulated for discrete actions while the latter was designed to receive pixels as inputs.


\vspace{-7pt}
\section{Related Work}
\cite{MachadoBB17,eigenoptiondisco} extend the Proto-Value Functions framework to the hierarchical reinforcement learning setting. The eigenvectors of the graph Laplacian are used to define an intrinsic reward signal in order to learn options \citep{10.1016/S0004-3702(99)00052-1}. However, their reward shaping function is not guaranteed invariance with respect to the original MDP's optimal policy. More recently, \cite{DBLP:journals/corr/abs-1810-04586} propose a way to estimate the eigendecomposition of the graph Laplacian by minimize a particular objective inspired by the graph drawing objective \citep{10.5555/1756869.1756936}. The authors then propose to use the learned representation for reward shaping but restrict their experiments to  navigation-based tasks.
Our approach is also reminiscent of the Value Iteration Networks \citep{DBLP:journals/corr/TamarLA16} where the authors propose to use convolutional neural networks to perform value iteration. In their approach, value iteration is performed on an approximate MDP which may not share similarities to the original MDP, whereas we propose to approximate the underlying graph of the original MDP. Return Decomposition for Delayed Rewards \citep{rudder} is another reward shaping method that tackles the problem of credit assignment through the concept of reward redistribution. However, it is not straightforward to apply in practice whereas potential-based reward shaping takes a simple form.
Finally, our method is most closely related to the DAVF approach \citep{Klissarov2018DiffusionBasedAV}, however in their work the authors leverage GCNs to obtain value functions whereas we look to define potential functions for reward shaping.

\vspace{-7pt}
\section{Discussion}
We presented a scalable method for learning potential functions by using a Graph Convolutional Network to propagate messages from rewarding states. 
As in the Proto-Value Function framework \citep{MahadevanM07,Mahadevan2005} this propagation mechanism is based on the graph Laplacian which is know to induce a smoothing prior on the functions over states. 
 As shown in the experiments and illustrations, the resulting distribution over states can then be leveraged by the agent to accelerate the performance.
Moreover, unlike other reward shaping techniques, our method is guaranteed to produce the same optimal policy as in the original MDP. 



We believe that our approach shows potential in leveraging advances from graph representation learning  \citep{DBLP:journals/corr/BronsteinBLSV16,DBLP:journals/corr/abs-1709-05584} in order to distribute information from rewarding states. In particular it would be interesting to explore with transition models other than the graph Laplacian that would be more closely related to the true transition matrix $P^{\pi}$, as in done in \cite{petrik} by the introduction of Krylov bases. Another possible improvement would be to represent actions and potentially the policy itself when approximating the graph. This could be leveraged  through look-ahead advice \citep{wiewiora}. 

Finally, in this paper we moslty took a model-free approach to exploit GCNs in the context of reinforcement learning. Future work could consider a more sophisticated approach by taking inspiration form grid cell-like constructs \citep{PMID:5124915} or by combining our sampling strategy with model roll-outs \citep{dyna} to construct the MDP's underlying graph.

\newpage
\section*{Broader Impact}
We believe that our research provides scalable ways to learn potential functions for reward shaping yet maintaining guarantees of invariance with respect to the optimal policy of the original setting. We believe that one of the most promising attribute of our line of research is the fact that the optimal policy remains unchanged. This is a fundamental feature for sensitive applications such as healthcare, recommendation systems and financial trading. 

Moreover, by using potential based reward shaping, our method is meant to accelerate learning. This is another very important characteristic with regards to applications where sample complexity, memory and compute are of importance. Accelerating learning can also have downsides in the situations where there is competition between technologies. This could lead to one competitor obtaining an advantage due to faster learning, which can then exacerbate this advantage (rich get richer situation). We suggest that any publications that proceed in this line of research to be open about implementation details and hyperparameters.

Another point to consider when proceeding to applications is related to the complexity of the MDP. If the application is small enough, it would be a good approach to try and store the whole underlying graph. We expect that in such settings our approach will provide significant improvements with reduced complexity when compared to approaches that require the eigen-decomposition of a transition model. If the MDP is too large to be reconstructed, we suggest applying our sampling strategy that will avoid increases in computational complexity. In these settings, we believe our approach can still provide robust improvements, however it might be more sensitive to the choice of hyperparameters. However, we have shown that across a wide range of games, high values of $\alpha$ provide good improvements and we expect this to be the case in many applications.

\begin{ack}
The authors would like to thank the National Science and Engineering Research Council of Canada (NSERC) and the Fonds de recherche du Quebec - Nature et Technologies (FRQNT)  for funding this research; Khimya Khetarpal for her invaluable and  timely help; Zafareli Ahmed and Sitao Luan for useful discussions on earlier versions of the draft and the anonymous  reviewers for providing critical and constructive feedback.
\end{ack}

\bibliography{ref}
\bibliographystyle{plainnat}

\clearpage
\onecolumn
\appendix

\section{Appendix}

\subsection{Tabular experiments}
\label{sec:tab}

\subsubsection{Implementation Details}
For our experiments of the FourRooms and FourRoomsTraps domains we based our implementation on \citep{bacon2017option} and ran the experiments for 300 episodes that last a maximum of 1000 steps. As these are tabular domains, each state is defined by a single feature for both the actor and the critic. For the $\Phi_{GCN}$ and $\Phi_{\alpha \beta}$ results we used the full graph of the MDP to compute the message passing mechanism. Full hyperparameters are listed here:

\begin{table}[hbt!]
    \centering
    \begin{tabular}{c|c}
    Hyperparameter & Value \\
    \hline
     Actor lr & 1e-1\\
     Critic lr & 1e-1 \\
     Discount & 0.99  \\
     Max Steps & 1000  \\
     Temperature & 1e-1 \\
     GCN: hidden size & 64 \\
     GCN: $\alpha$ & 0.6 \\
     GCN: $\eta$ & 1e1 \\
     \end{tabular}
    \caption{Hyperparameters for the FourRooms and FourRoomsTraps domain }
    \label{tab:my_label}
\end{table}

\subsubsection{Visualization}
We visually inspect the output of $\Phi_{GCN}$ and $\Phi_{\alpha \beta}$ on the FourRooms and FourRoomsTraps domain. In Fig we notice a close ressemblance in the final output. In this particular environment, performing the messages obtained the forward-backward algorithm are therefore well approximated by the propagation mechanism of the GCN. This also results in very similar empirical performance.

\begin{figure}[h]
\begin{center}
    \begin{subfigure}[c]{0.26\textwidth}
    \centering
        \includegraphics[width=1.\columnwidth]{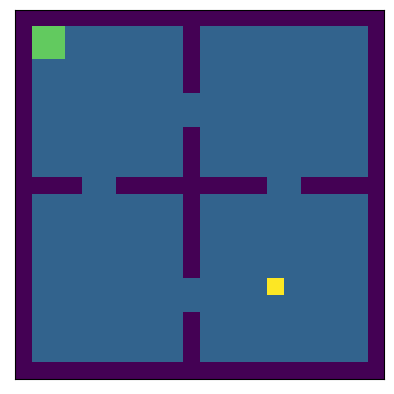}
        \caption[]
        {{\small FourRooms}}
    \end{subfigure} 
    \begin{subfigure}[c]{0.3\textwidth}
    \centering
        \includegraphics[width=1.\columnwidth]{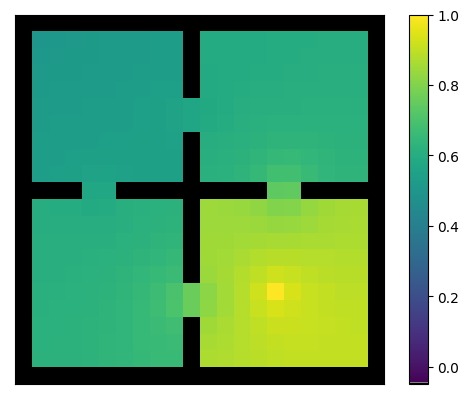}
        \caption[]
        {{\small $\Phi_{\alpha \beta}$  }}
    \end{subfigure} 
    \begin{subfigure}[c]{0.3\textwidth}
    \centering
        \includegraphics[width=1.\columnwidth]{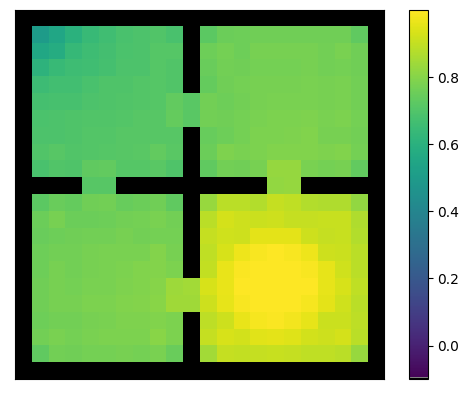}
        \caption[]
        {{\small $\Phi_{GCN}$}}
    \end{subfigure} 
    \caption{Visualization of the function over states obtained through message passing $\Phi_{\alpha \beta}$ and by the GCN $\Phi_{GCN}$ on the FourRooms domain where the agent starts in the upper left states (green) and has to get to the goal (yellow) in the bottom right room. }
    \label{fig:4r_vis}
    \end{center}
\end{figure}

\begin{figure}[h]
\begin{center}

    \begin{subfigure}[c]{0.26\textwidth}
    \centering
        \includegraphics[width=1.\columnwidth]{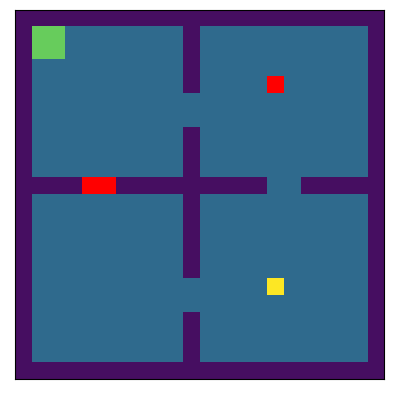}
        \caption[]
        {{\small  FourRoomsTraps}}
    \end{subfigure} 
    \begin{subfigure}[c]{0.3\textwidth}
    \centering
        \includegraphics[width=1.\columnwidth]{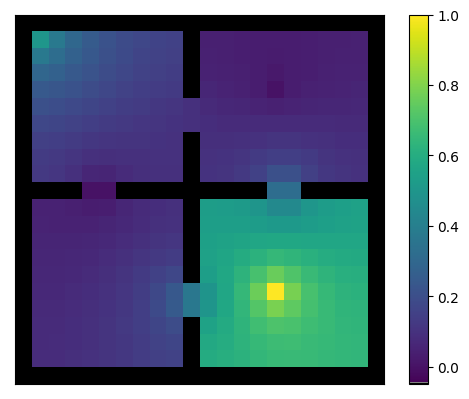}
        \caption[]
        {{\small $\Phi_{\alpha \beta}$ }}
    \end{subfigure} 
    \begin{subfigure}[c]{0.3\textwidth}
    \centering
        \includegraphics[width=1.\columnwidth]{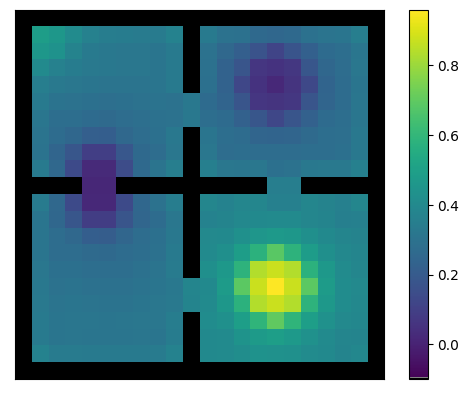}
        \caption[]
        {{\small  $\Phi_{GCN}$}}
    \end{subfigure} 
    \caption{Visualization of the function over states obtained through message passing $\Phi_{\alpha \beta}$ and by the GCN $\Phi_{GCN}$ on the FourRooms domain where the agent starts in the upper left states (green) and has to get to the goal (yellow) in the bottom right room while aboid the negative rewards (red).}
    \label{fig:4rt_vis}
    \end{center}
\end{figure}
\vspace{2cm}

\newpage
\subsection{High-Dimensional control}
\label{sec:control}

For experiments on MiniWorld-FourRooms-v0, MiniWorld-MyWayHome-v0 and  MiniWorld-MyWayHomeNoisyTv-v0 we based our implementation on \citep{pytorchrl} and ran the experiments for 5M steps and 20M steps respectively with 10 random seeds. The input provided to the GCN is the last hidden layer of the shared CNN network used by the actor and the critic.  The architecture for the actor-critic was  kept to be the same with respect to the original codebase. We provide a full list of the values for the hyperparameters:

\begin{figure}
    \centering
    \includegraphics[width=.7\textwidth]{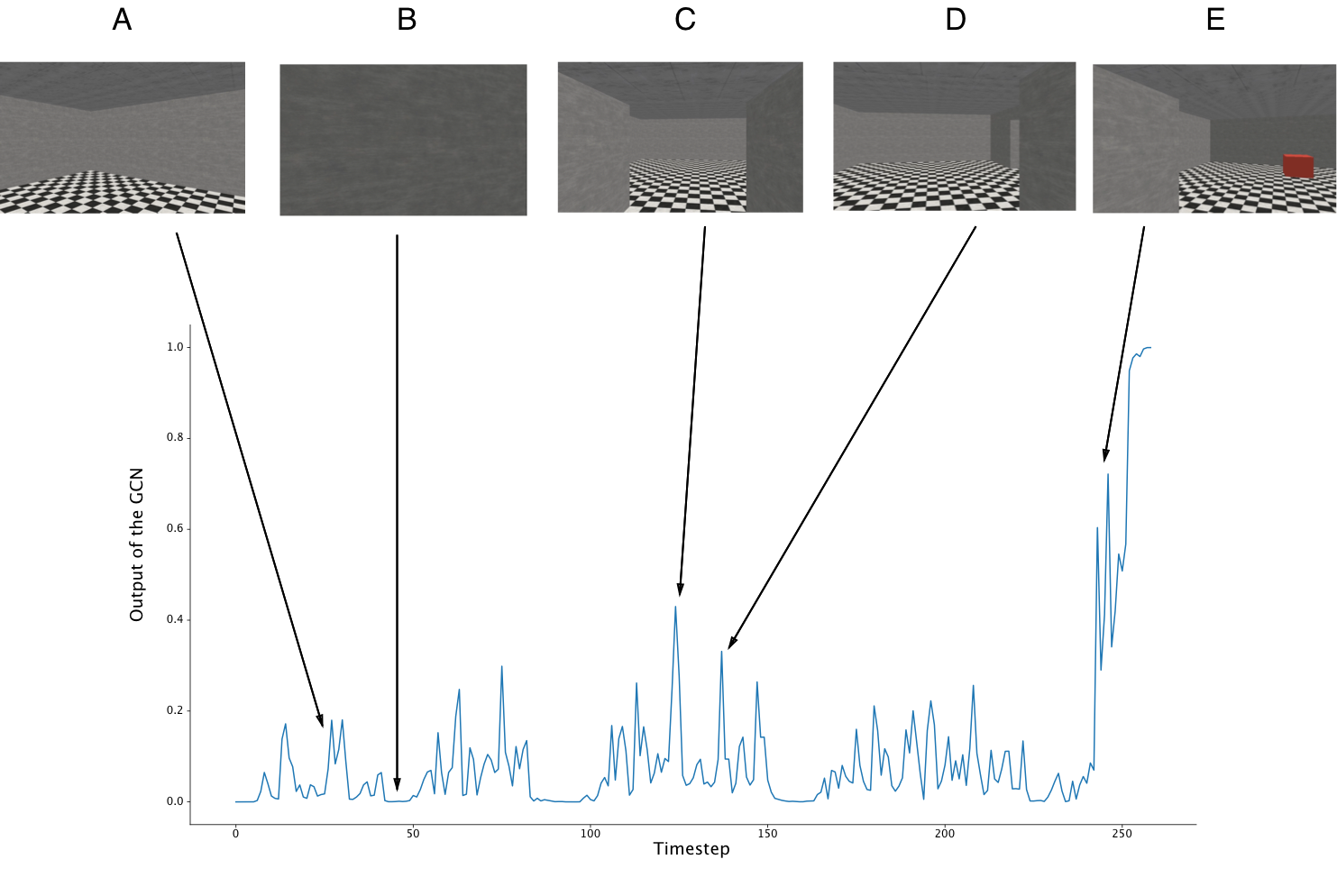}
    \caption{Visualization of $\Phi_{GCN}$ across timesteps in an episode on the MiniWorld-FourRooms-v0 domain. \textbf{A} The agent is scanning the room in search of the red box (the goal). \textbf{B} While scanning the room, the agent is faced with a close wall, therefore the output of the GCN is low as there is nothing promising at this timestep. \textbf{C} The agent is about to enter the hallway between two rooms. We notice that consistently the output of the GCN spikes at such key moments. \textbf{D} The agent is about to cross to the next room and the output of the GCN is high to encourage crossing over. \textbf{E} The agent has seen the red box and the GCN's output spikes in order to push the agent towards the goal.}
    \label{fig:visuplotminiw}
\end{figure}

We also provide visualization of the output of the GCN on the MiniWorld-FourRooms-v0 domain, as shown in Figure \ref{fig:visuplotminiw}. In general, we notice that the output of the GCN is low in uninteresting moments (such as the agent facing the wall), but increases at key moments such as when it crosses hallways between rooms and in the sight of the goal.

Finally, we provide the hyperparameters used for experiments in Table 2.

\begin{table}[ht!]
    \centering
    \begin{tabular}{c|c}
    Hyperparameter & Value \\
    \hline
     Learing rate & 2.5e-4\\
     $\gamma$ & 0.99  \\
     $\lambda$ & 0.95  \\
     Entropy coefficient & 0.01  \\
     LR schedule & constant \\
     PPO steps & 128 \\
     PPO cliping value & 0.1 \\
     \# of minibatches & 4 \\
     \# of processes & 32 \\
     GCN: $\alpha$ & 0.8 \\
     GCN: $\eta$ & 1e1 \\
     ICM coeff. & 1e-2\\
     RND coeff. & 1.0
     \end{tabular}
    \caption{Hyperparameters for the MiniWorld experiments }
    \label{tab:miniw}
\end{table}

\subsection{Sweep across values of $\alpha$}
\label{sec:alpha}

\begin{figure}[h]
    \centering
    \includegraphics[width=.4\textwidth]{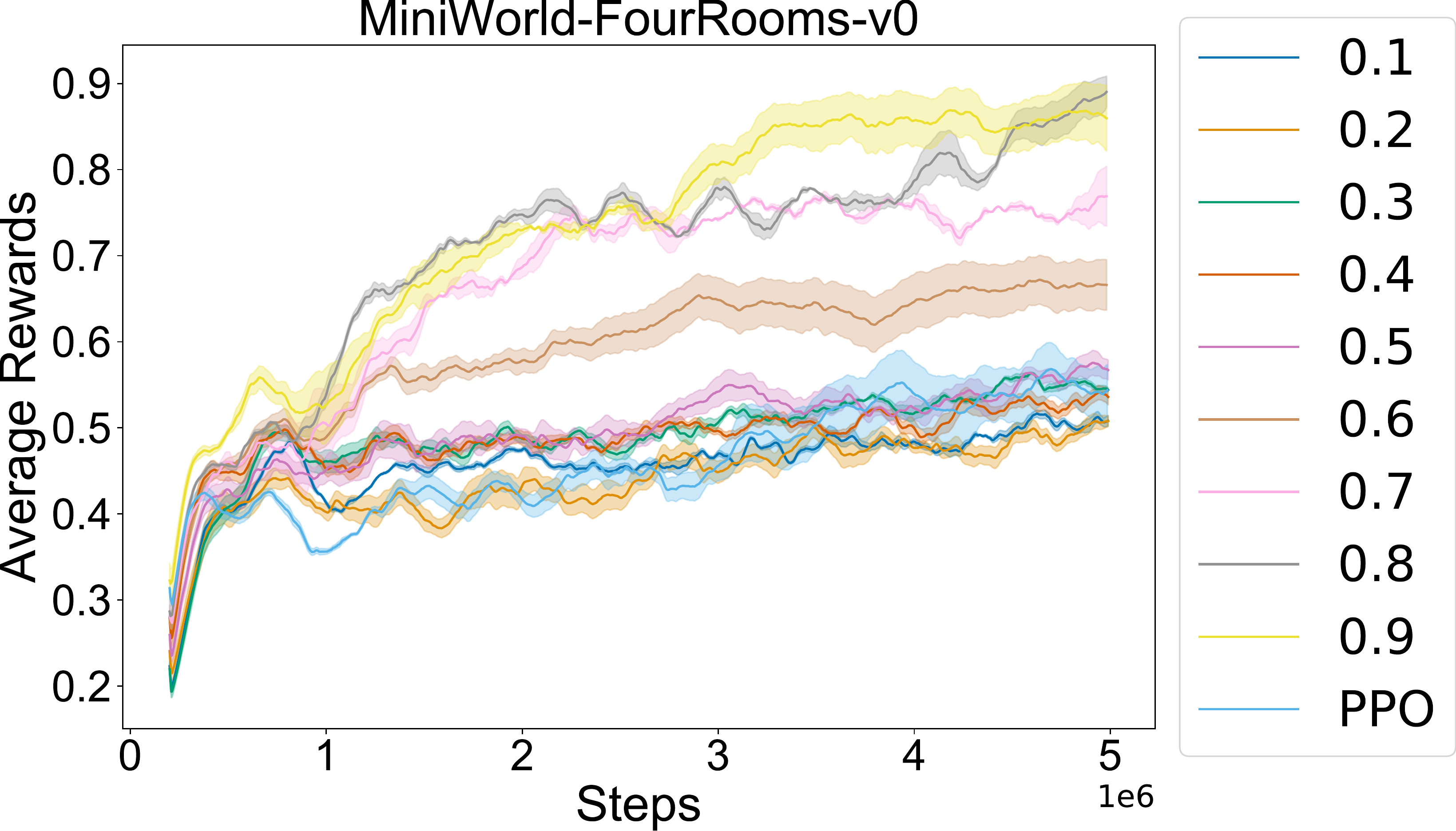}
    \caption{Performance on MiniWorld-FourRooms-v0 across values of $\alpha$. Higher values lead to better results.}
    \label{fig:alpha}
\end{figure}

In Fig.\ref{fig:alpha} we present a sweep over the values of $\alpha$ which controls the trade-off between the reward-shaped value function and the regular value function. We notice that improvements are consistent for values of $\alpha$ higher than 0.5.

\subsection{Investigation of the hyperparameter $\eta$}
\label{eta_hyp}
\begin{figure}[h]
\centering
    \begin{subfigure}[c]{.26\columnwidth}
    \centering
        \includegraphics[width=1.\columnwidth]{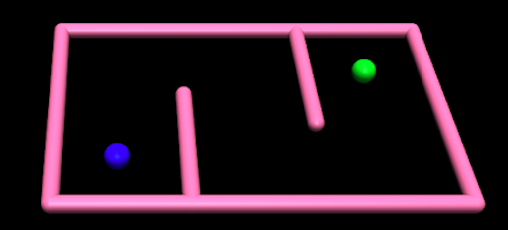}
        \caption{{\small  SMaze-v0 task.}}
        \label{fig:smaze}
    \end{subfigure}
    \begin{subfigure}[c]{.21\columnwidth}
    \centering
        \includegraphics[width=1.\columnwidth]{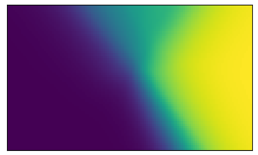}
        \caption{{\small  $\eta = 0.1$}}
    \end{subfigure}
    \begin{subfigure}[c]{.21\columnwidth}
    \centering
        \includegraphics[width=1.\columnwidth]{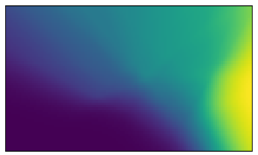}
        \caption{{\small  $\eta = 1.0$}}
    \end{subfigure}
    \begin{subfigure}[c]{.21\columnwidth}
    \centering
        \includegraphics[width=1.\columnwidth]{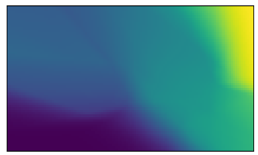}
        \caption{{\small  $\eta = 10.0$}}
    \end{subfigure}    
  \caption{\textbf{Comparison of the output} of $\Phi_{GCN}$ for different values of $\eta$  from Eq.~\ref{gcnloss}, which  controls the complexity of the output. The visualization is performed on  the SMaze-v0 environment shown in Fig.\ref{fig:smaze} where the agent is the blue dot starting in lower left corner and the goal in the green dot in the upper right corner. Lower values of $\eta$ lead to less propagation and therefore a more simple, yet biased, output. As $\eta$ increases, the  output shows a more complex structure.}
  \label{fig:mazelambda}
\end{figure}

We investigate the effect of the hyperparameter $\eta$ in the GCN's loss function (Eq.~\ref{gcnloss}) which mediates between $\mathcal{L}_{0}$ and  $\mathcal{L}_{prop}$, where the former is the supervised loss and the latter the propagation loss. The latter depends on the adjacency matrix of the graph and can therefore accentuate the recursive mechanism of the GCN as opposed to the supervised loss which controls the base case values. The Fig.~\ref{fig:mazelambda} illustrates the result of varying  $\eta$ in the SMaze-v0 environment shown in Fig.~\ref{fig:smaze} where the agent (blue dot) has to navigate to the goal (green dot). In Fig.~\ref{fig:mazelambda}b, we see that for a low value of $\eta$, the GCN outputs a simple solution where the states on the left have a low value (blue) while states on right have a higher value (yellow), which matches the position of the goal and the starting states. This solution shows more bias than variance as it favors a simple output. As we increase the value of $\eta$, the propagation loss has greater importance and the output $\Phi_{GCN}$ shows more complexity, incorporating more information about the environment's dynamics, such as the walls and the specific position of the goal.  There exists an interesting parallel with the $\lambda$ parameter in the TD($\lambda$) algorithm: $\lambda$ controls the trade-off between bias and variance of the return estimation, while $\eta$ controls the bias-variance trade-off of the propagation process.

\subsection{Atari 2600}
\label{app:atar}
For experiments on the Atari games we based our implementation on \citep{baselines} and we ran the experiments for 40M frames over 10 random seeds. The input provided to the GCN is the last hidden layer of the shared CNN network used by the actor and the critic.  The architecture for the actor-critic was  kept to be the same with respect to the original codebase. We provide a full list of the values for the hyperparameters:

\begin{table}[hbt!]
    \centering
    \begin{tabular}{c|c}
    Hyperparameter & Value \\
    \hline
     Learing rate & 2.5e-4\\
     $\gamma$ & 0.99  \\
     $\lambda$ & 0.95  \\
     Entropy coefficient & 0.01  \\
     LR schedule & constant \\
     PPO steps & 128 \\
     PPO cliping value & 0.1 \\
     \# of minibatches & 4 \\
     \# of processes & 8 \\
     GCN: $\alpha$ & 0.9 \\
     GCN: $\eta$ & 1e1 \\
     ICM coeff. & 1e-2\\
     LIRPG coeff. & 0.01
     \end{tabular}
    \caption{Hyperparameters for the Atari 2600 experiment used for the PPO, ICM and $\Phi_{GCN}$ algorithms. }
    \label{tab:atar}
\end{table}

\begin{table}[hbt!]
    \centering
    \begin{tabular}{c|c}
    Hyperparameter & Value \\
    \hline
     Learing rate & 1e-4\\
     $\gamma$ & 0.99  \\
     $\lambda$ & 0.95  \\
     Entropy coefficient & 0.001  \\
     LR schedule & constant \\
     PPO steps & 128 \\
     PPO cliping value & 0.1 \\
     \# of minibatches & 4 \\
     \# of processes & 128 \\
     RND int. coeff. & 1.0\\
     RND ext. coeff. & 2.0\\
     \end{tabular}
    \caption{Hyperparameters for the Atari 2600 experiment used for the RND algorithm. Notice that these hyperparameters are taken from \cite{rnd} as experiments were performed on Atari.}
    \label{tab:atar_rnd}
\end{table}

\begin{figure}
    \centering
    \includegraphics[width=1.\textwidth]{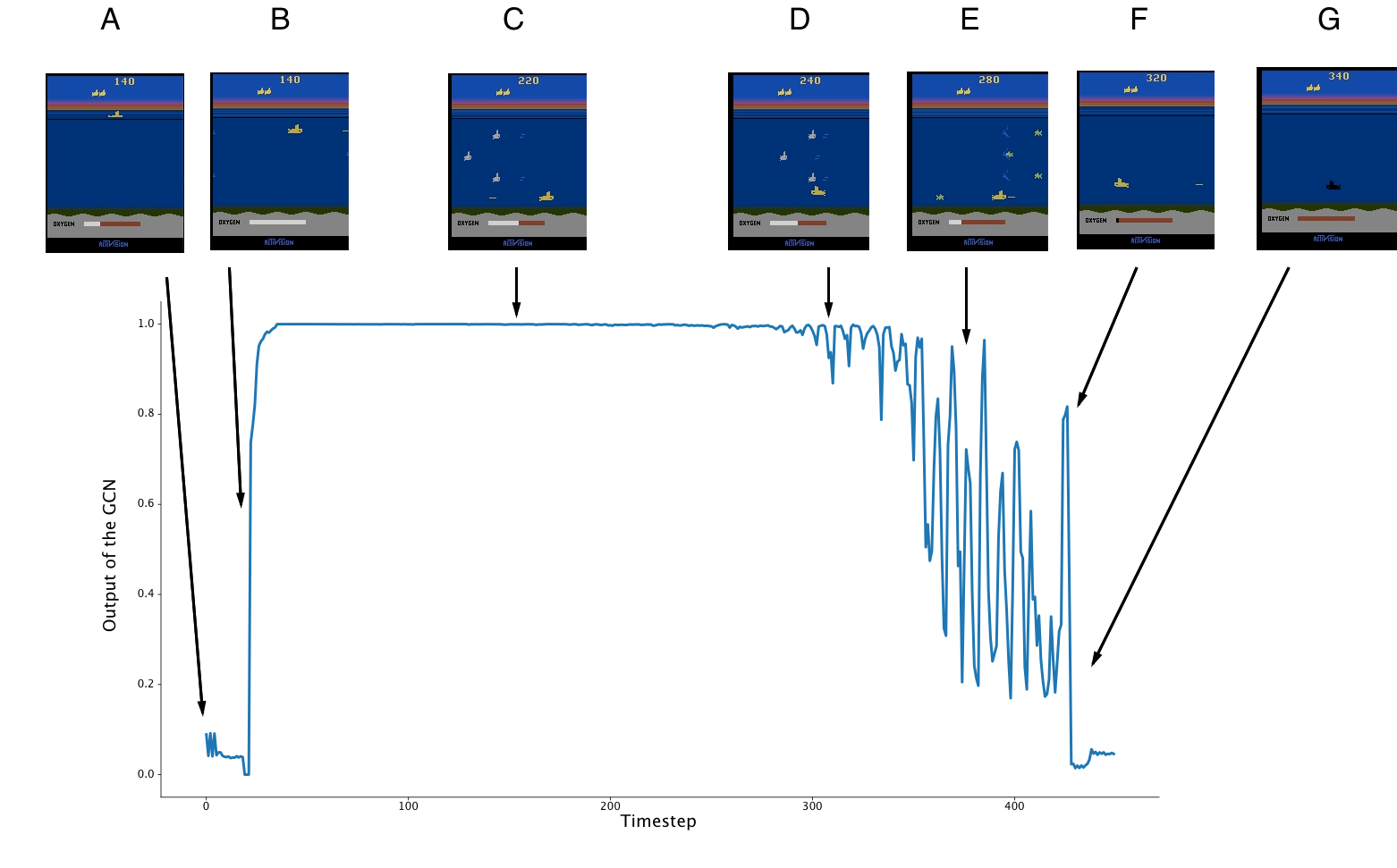}
    \caption{Visualization of the output of the GCN during an episode of the game SeaquestNoFrameSkip-v4. \textbf{A} The agent is waiting to fill up its oxygen tank, as plunging underwater before having a full tank is a direct loss of life. \textbf{B} The oxygen tank is full and the agent is moving towards the bottom of the sea \textbf{C} The agent stays at the most bottom part of the sea as this place is the safest. Indeed, if it moves up it has to deal with other submarines that shoot projectiles, as shown in the screenshot of the game. \textbf{D} The agent comes to close contact with another submarine, which would result in a loss of life, causing the GCN to output a decreased value. \textbf{E} The oxygen tank is getting low, which eventually leads to a loss of life. The GCN is gradually decreases its output as it can predict the incoming danger. \textbf{F} The agent momentarily moves up, which is the right thing to do in this situation: if it gets to the surface it can recharge its oxygen tank. \textbf{G} The agent has failed to reach the surface and loses a life. }
    \label{fig:seaq_visplot}
\end{figure}

We also provide a visualization of the output of the GCN in Figure \ref{fig:seaq_visplot}. In general, we notice that the GCN is sensitive to the volume of oxygen in the tank: it will encourage an agent with a full tank to move downwards and an agent with an almost empty tank to move upwards. The output of the GCN is also sensitive to situations when the agent gets dangerously close to other submarines.


\begin{figure}
    \centering
    \begin{subfigure}[c]{0.24\textwidth}
    \centering
      \includegraphics[width=1.\columnwidth]{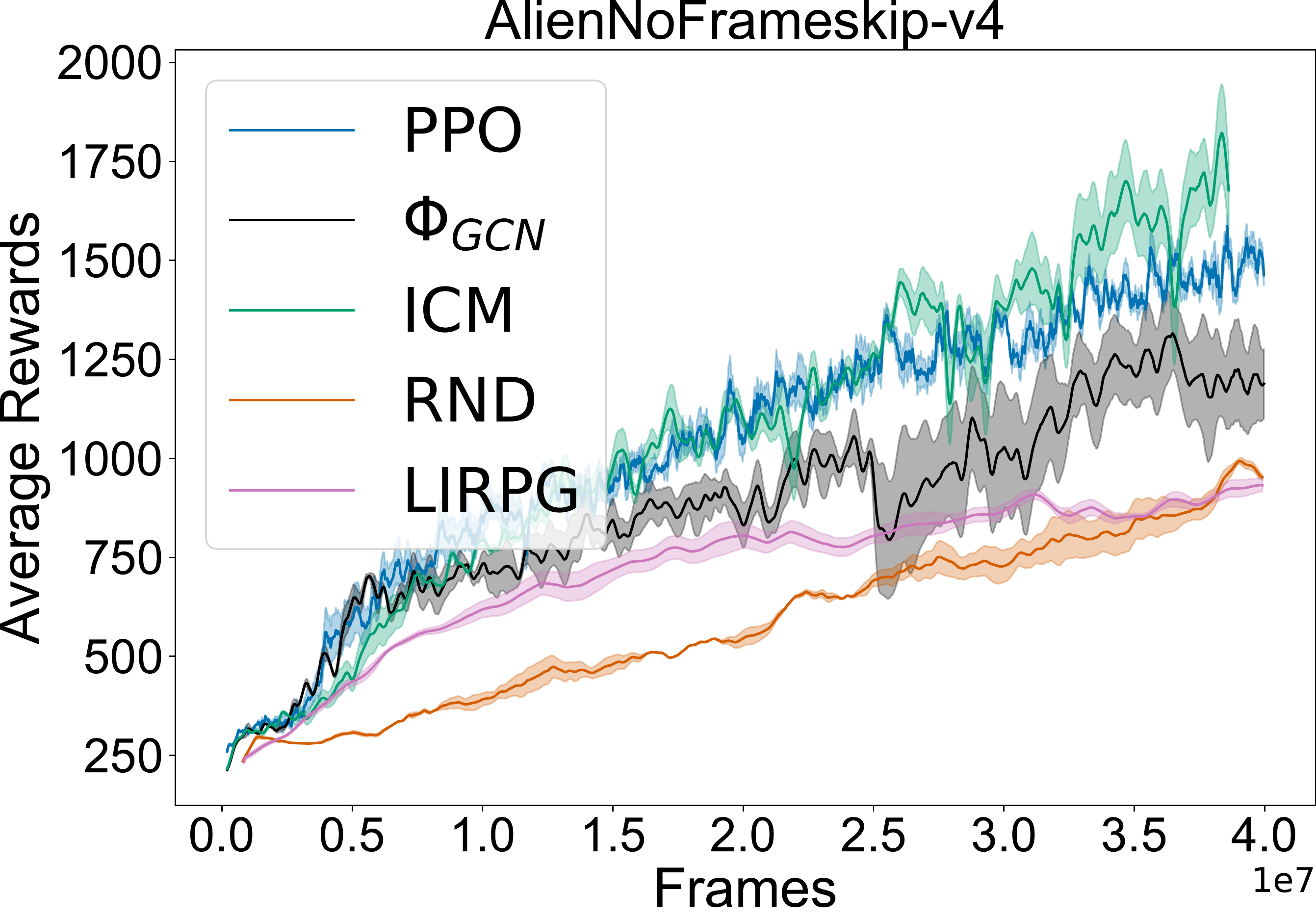}
    \end{subfigure} 
    \begin{subfigure}[c]{0.24\textwidth}
    \centering
      \includegraphics[width=1.\columnwidth]{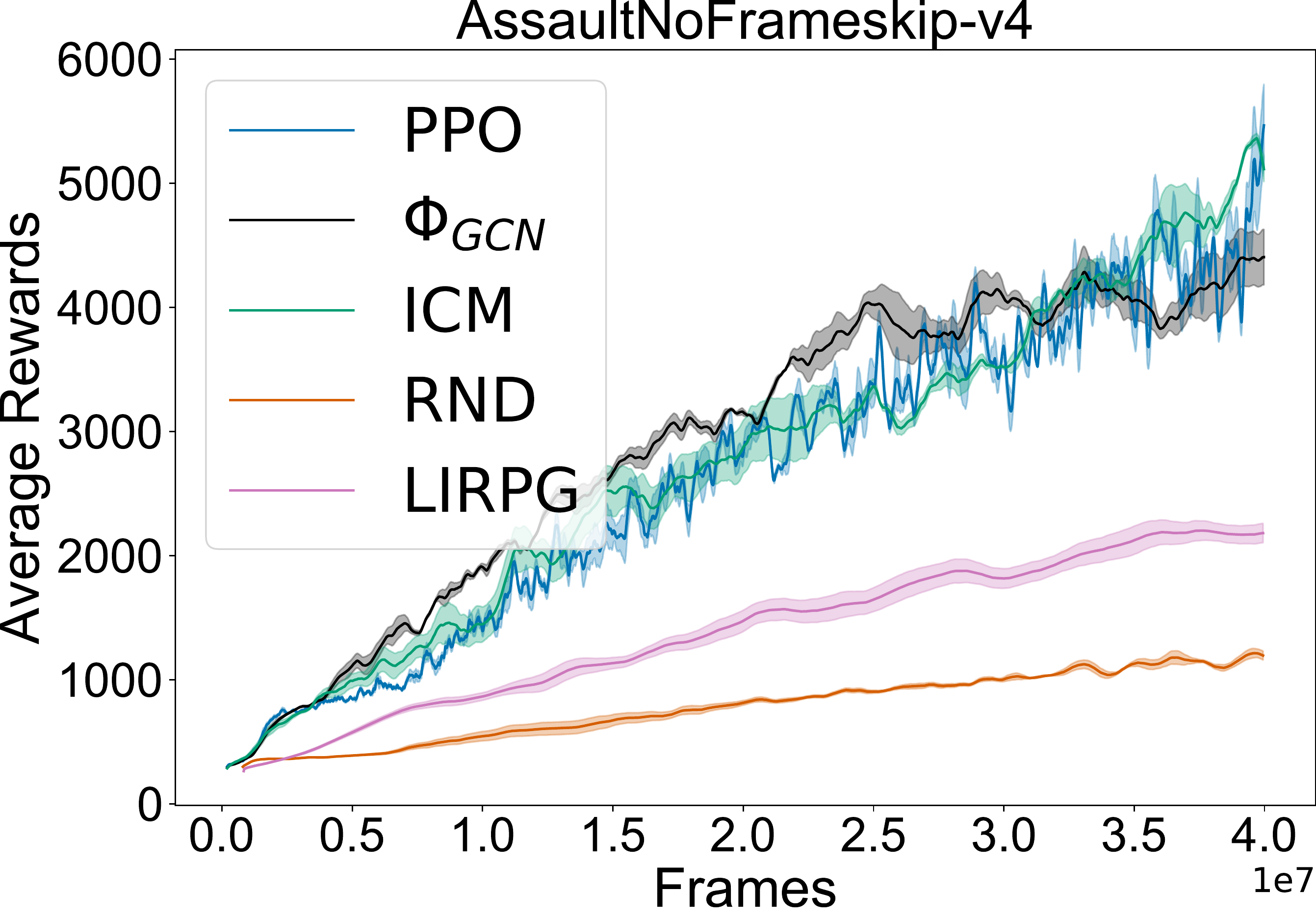}
    \end{subfigure} 
    \begin{subfigure}[c]{0.24\textwidth}
    \centering
      \includegraphics[width=1.\columnwidth]{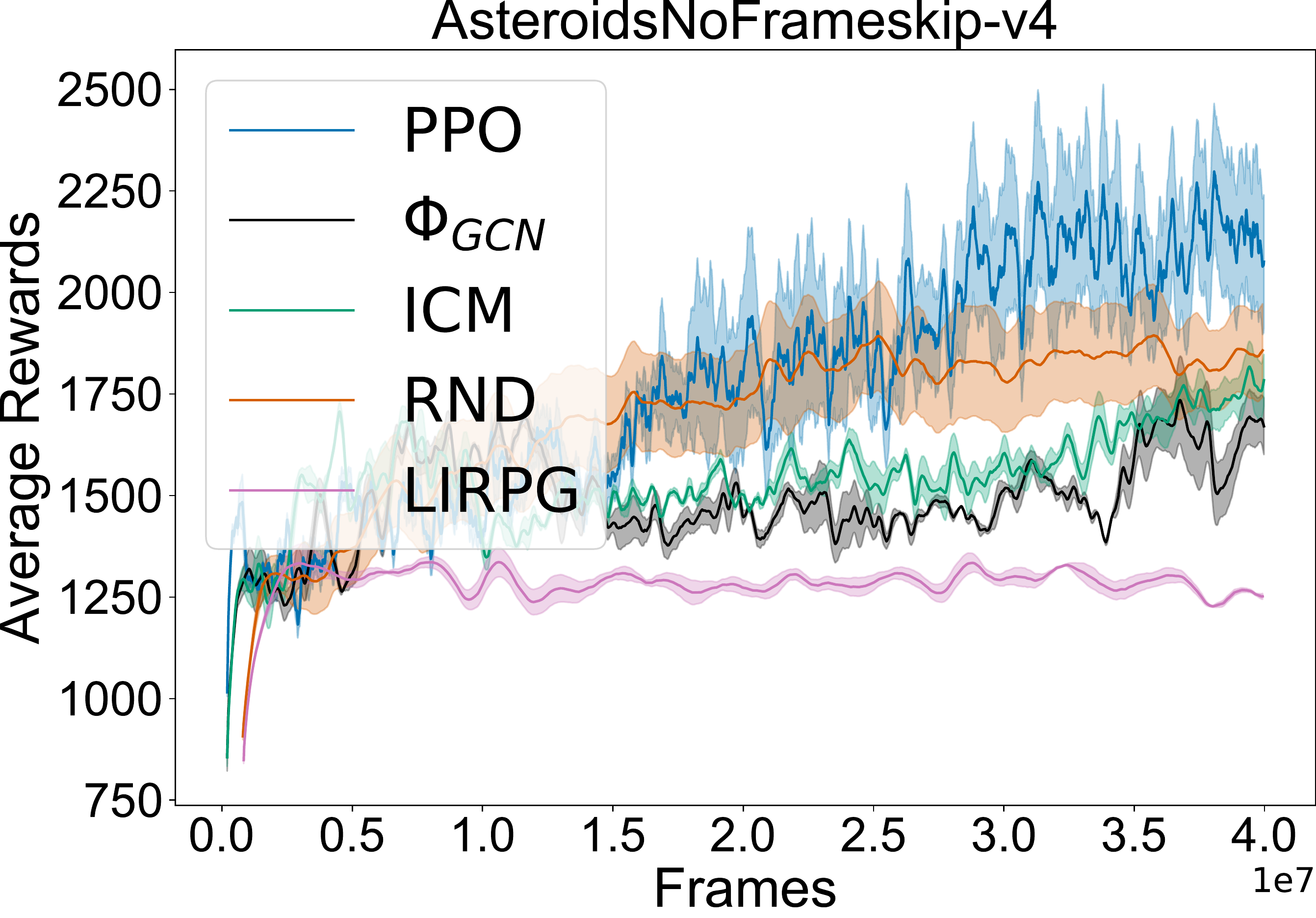}
    \end{subfigure} 
    \begin{subfigure}[c]{0.24\textwidth}
    \centering
      \includegraphics[width=1.\columnwidth]{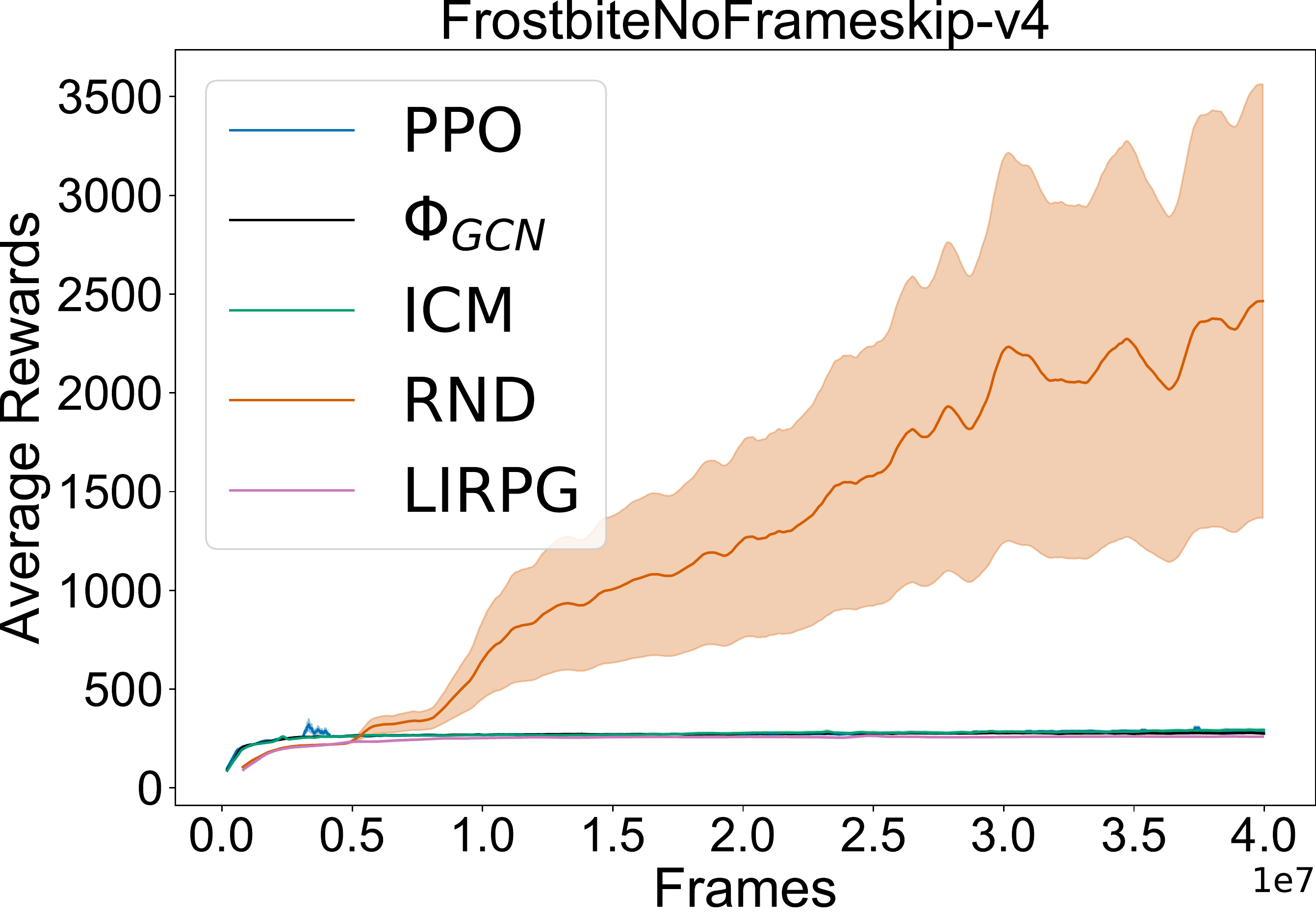}
    \end{subfigure} 
    
    \quad
    
    \begin{subfigure}[c]{0.24\textwidth}
    \centering
      \includegraphics[width=1.\columnwidth]{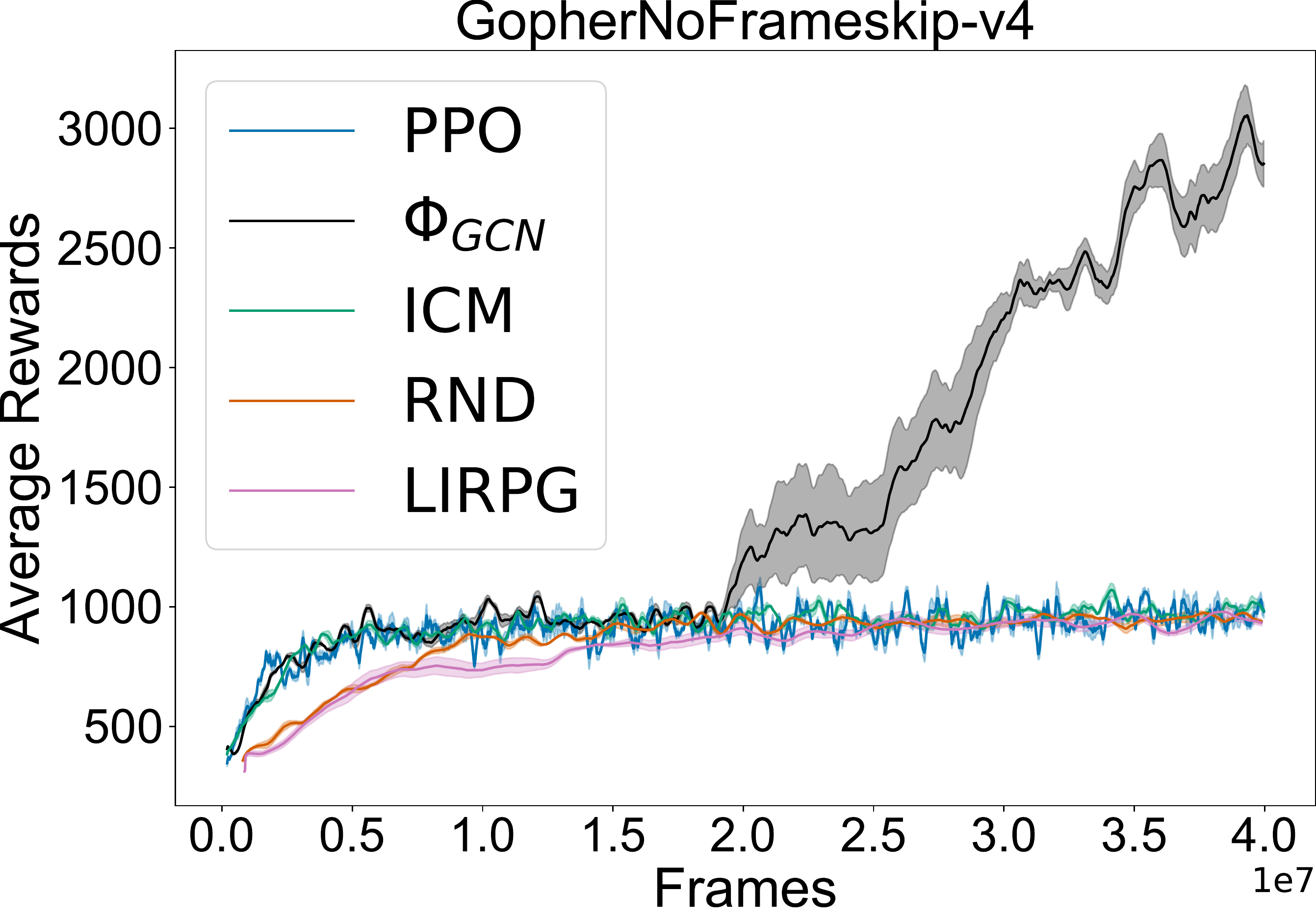}
    \end{subfigure} 
    \begin{subfigure}[c]{0.24\textwidth}
    \centering
      \includegraphics[width=1.\columnwidth]{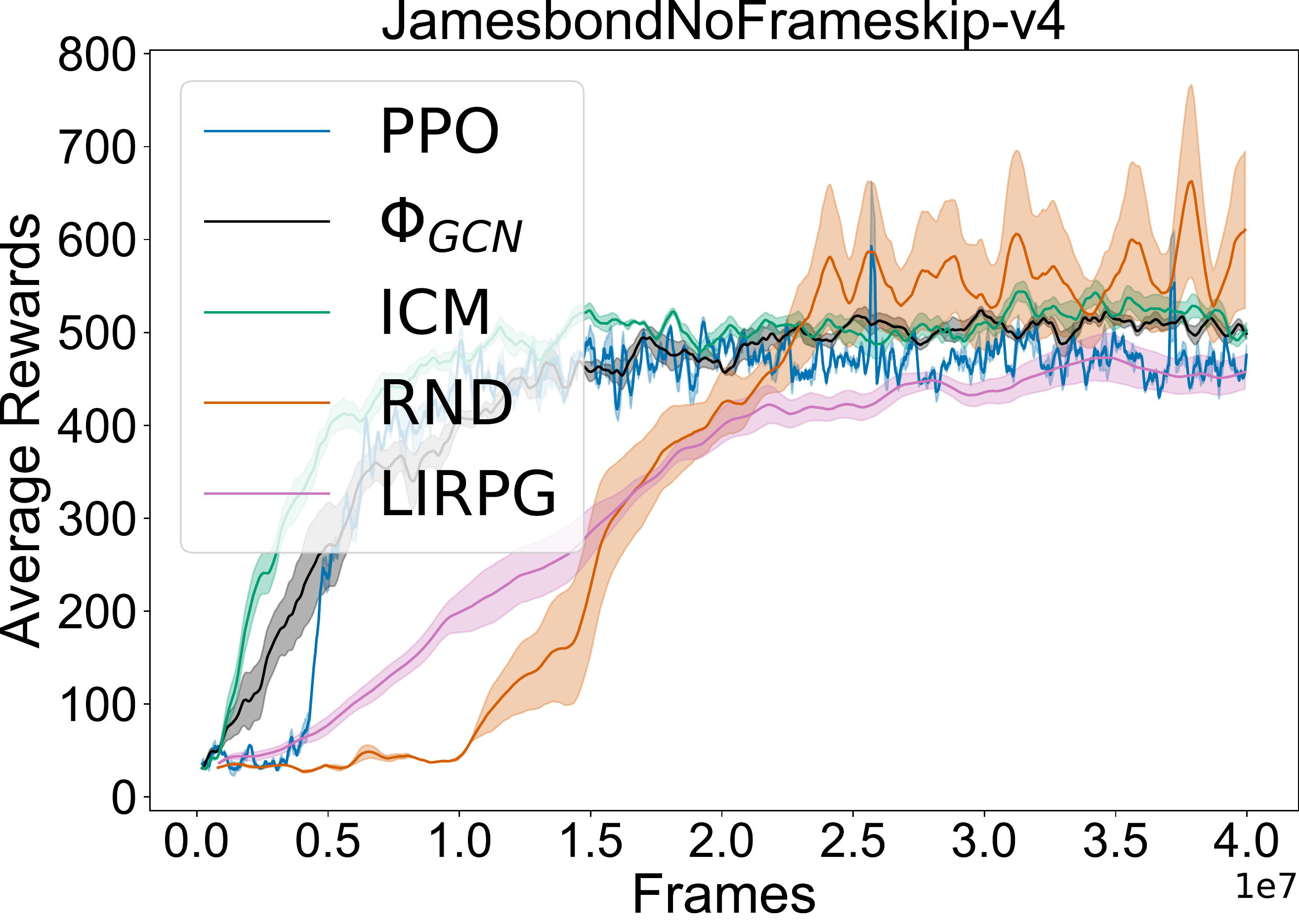}
    \end{subfigure} 
    \begin{subfigure}[c]{0.24\textwidth}
    \centering
      \includegraphics[width=1.\columnwidth]{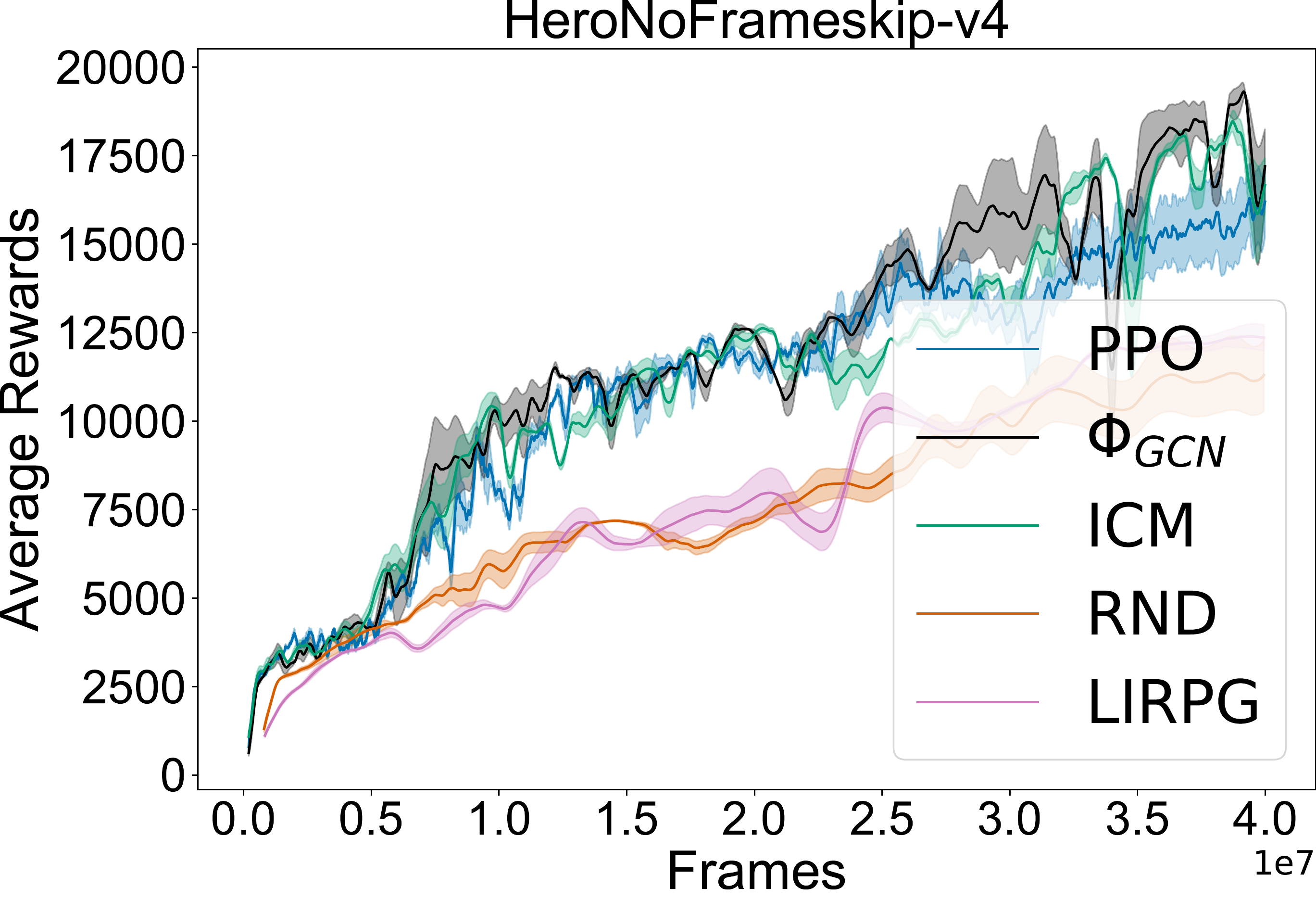}
    \end{subfigure} 
    \begin{subfigure}[c]{0.24\textwidth}
    \centering
      \includegraphics[width=1.\columnwidth]{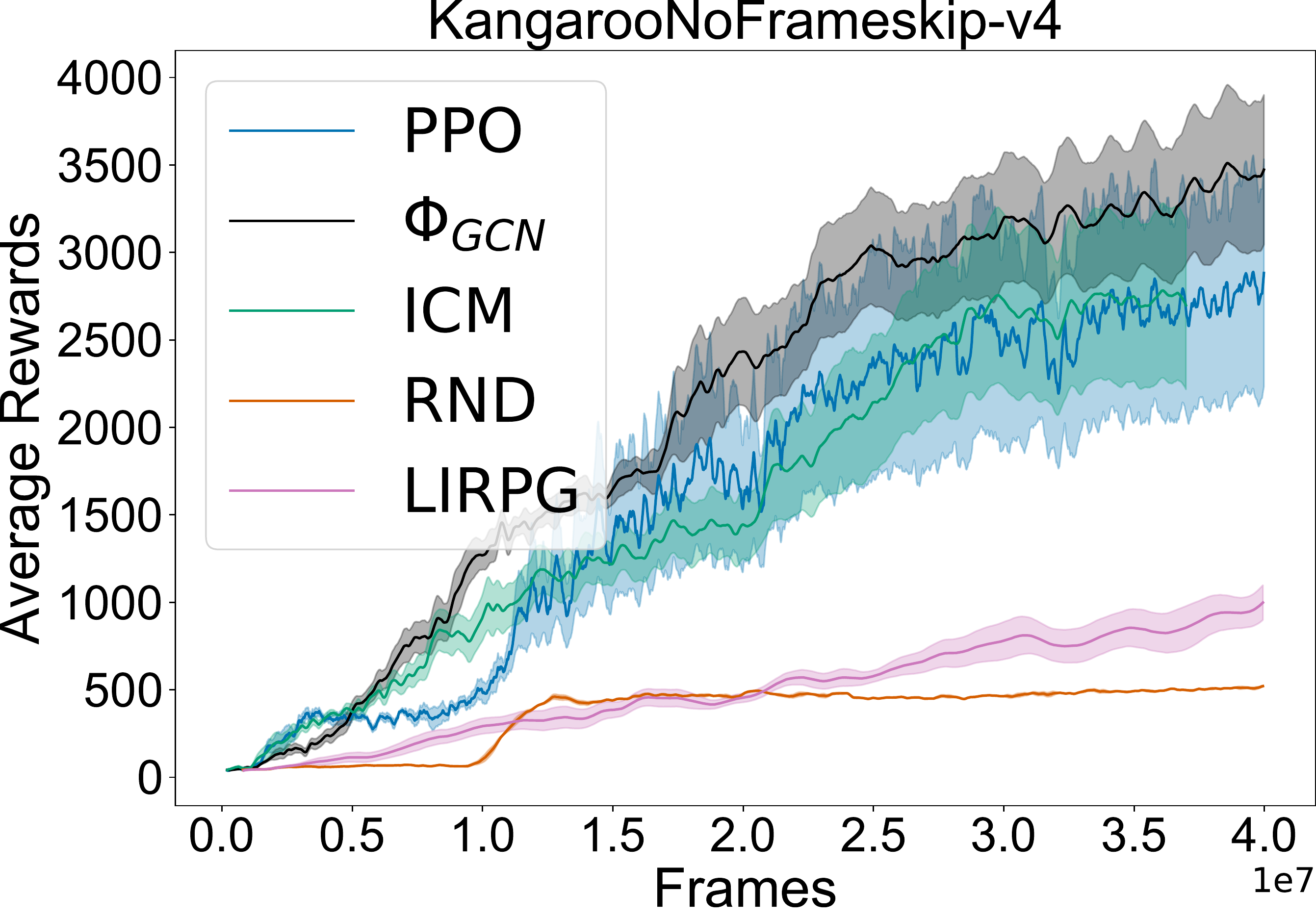}
    \end{subfigure}

    \quad
    
    \begin{subfigure}[c]{0.24\textwidth}
    \centering
      \includegraphics[width=1.\columnwidth]{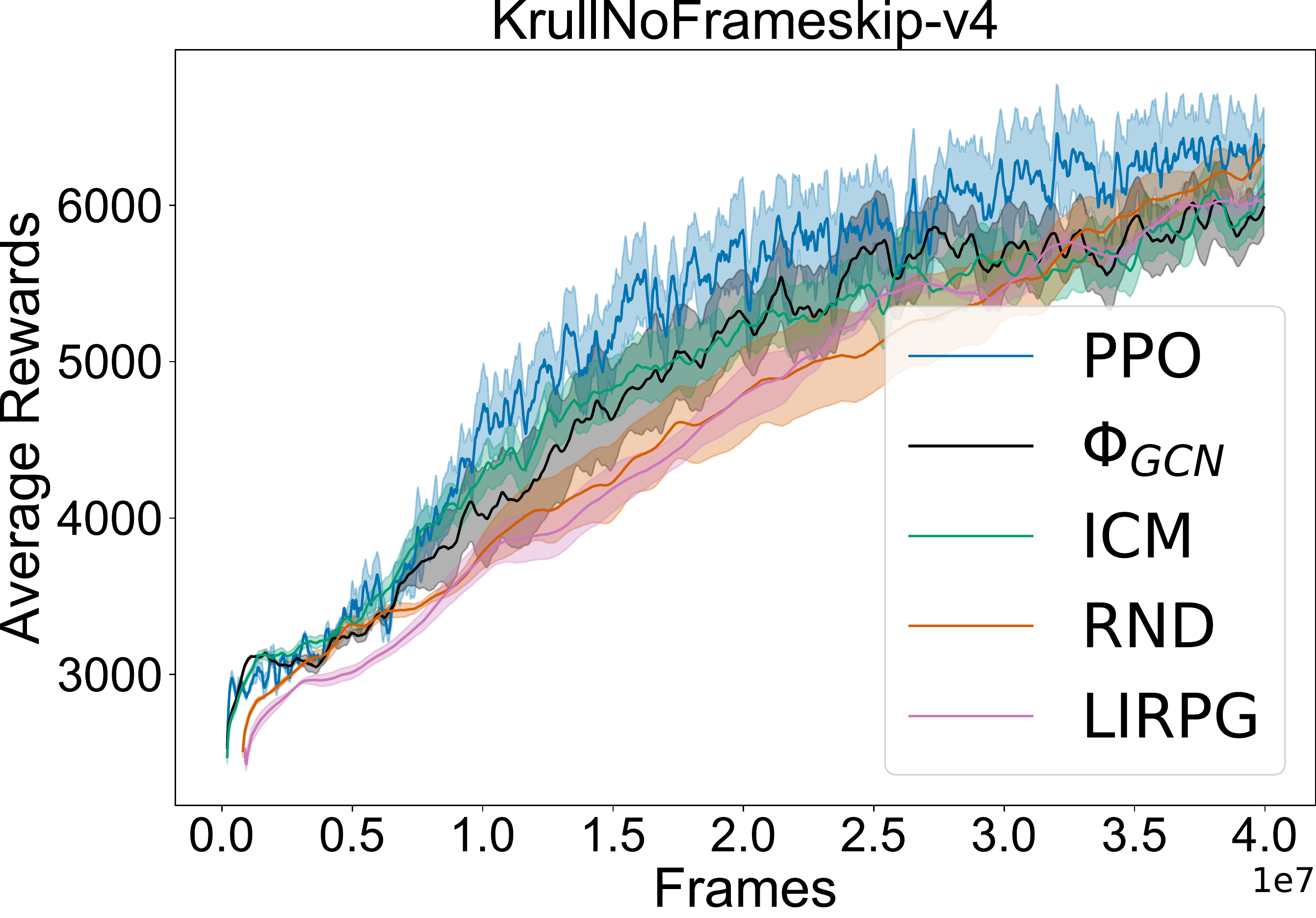}
    \end{subfigure} 
    \begin{subfigure}[c]{0.24\textwidth}
    \centering
      \includegraphics[width=1.\columnwidth]{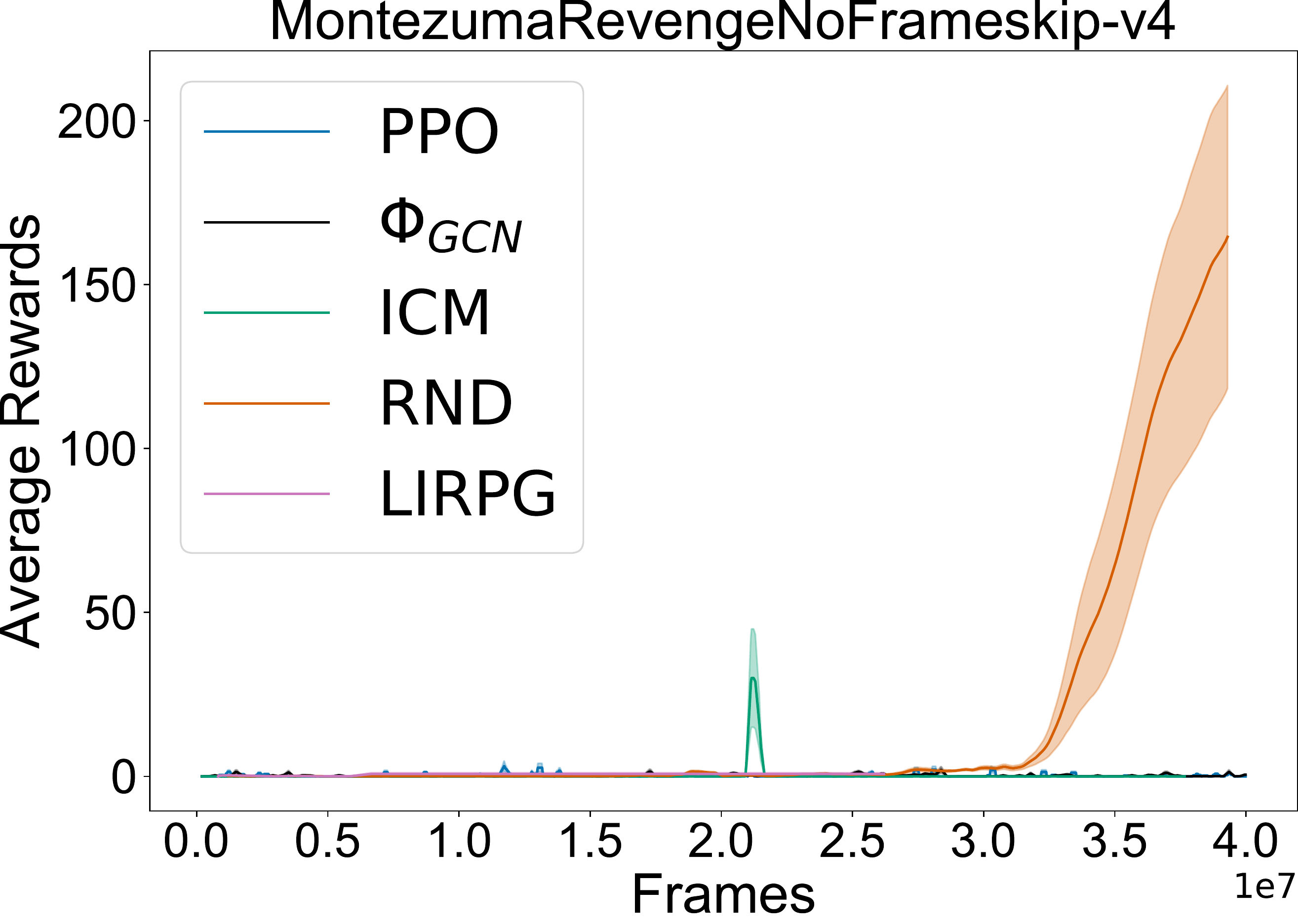}
    \end{subfigure} 
    \begin{subfigure}[c]{0.24\textwidth}
    \centering
      \includegraphics[width=1.\columnwidth]{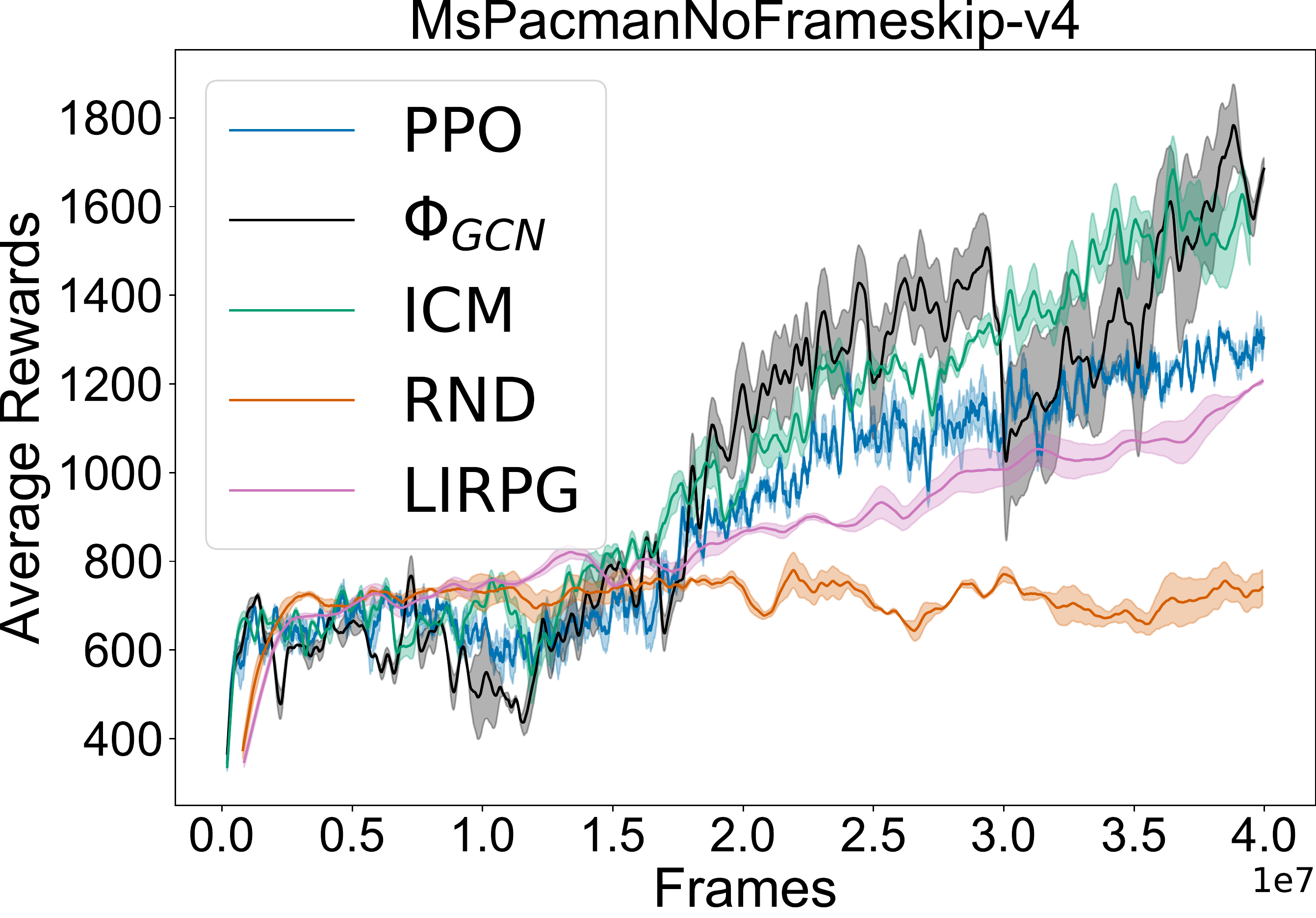}
    \end{subfigure} 
    \begin{subfigure}[c]{0.24\textwidth}
    \centering
      \includegraphics[width=1.\columnwidth]{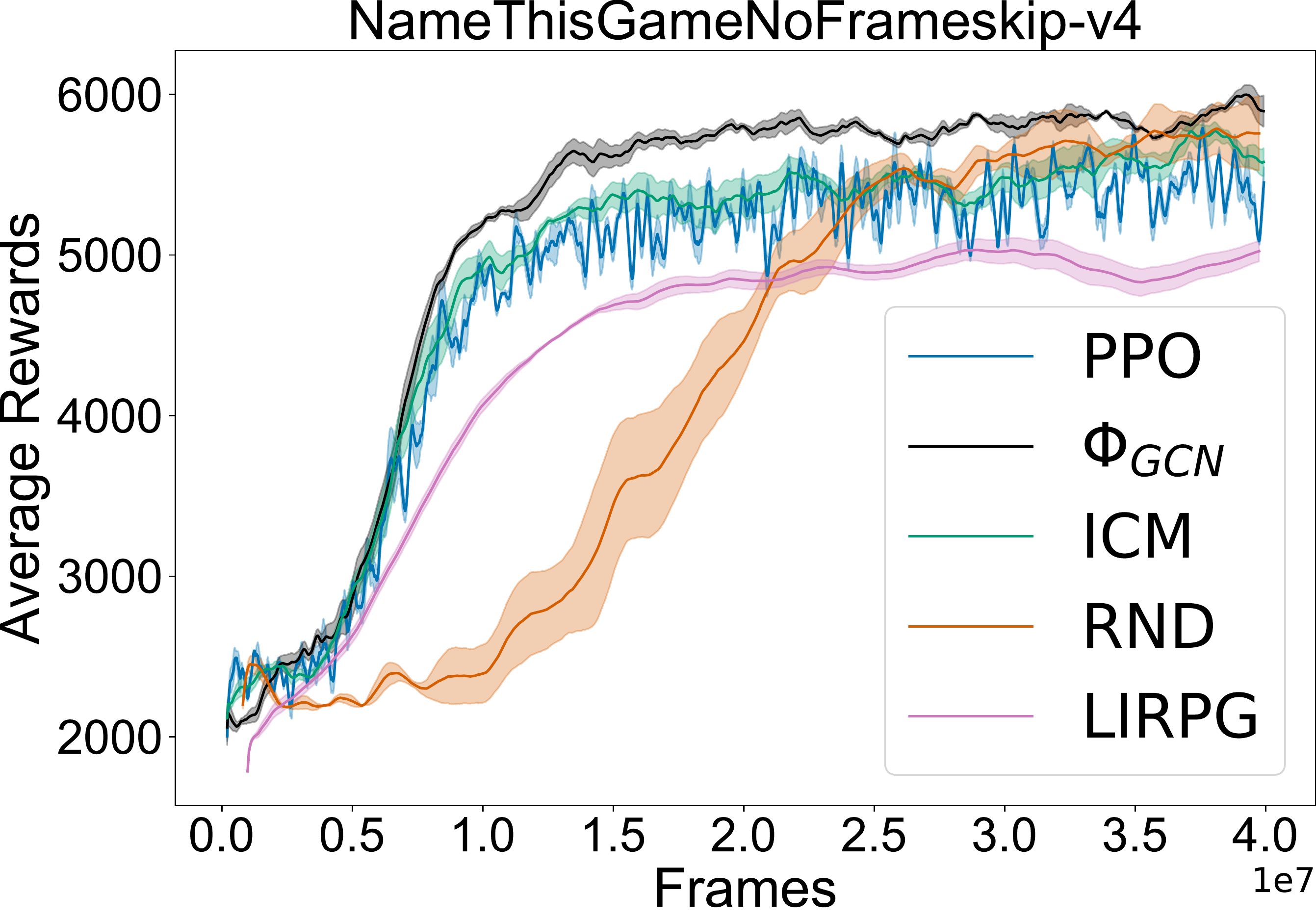}
    \end{subfigure}

    \quad
    
    \begin{subfigure}[c]{0.24\textwidth}
    \centering
      \includegraphics[width=1.\columnwidth]{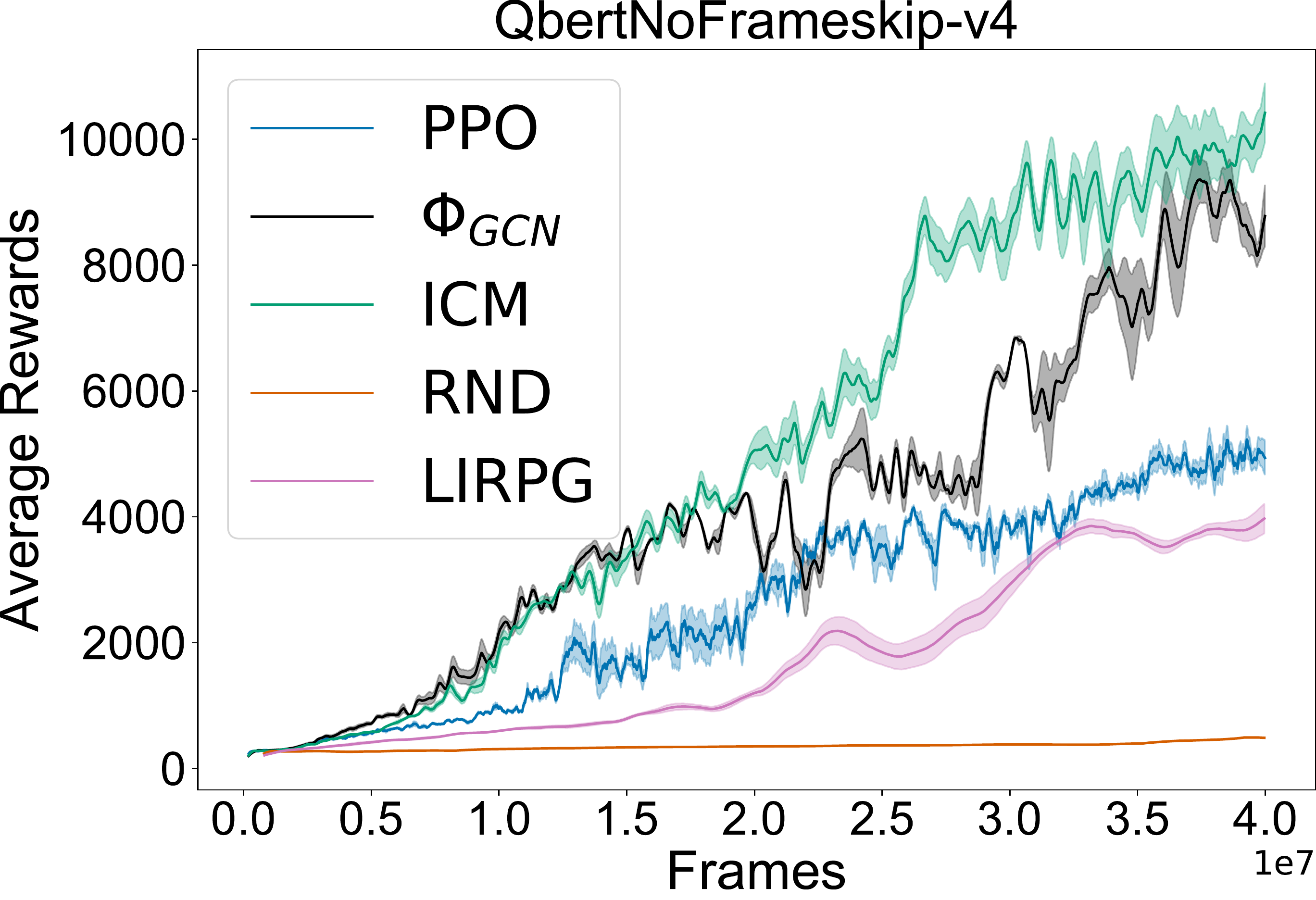}
    \end{subfigure} 
    \begin{subfigure}[c]{0.24\textwidth}
    \centering
      \includegraphics[width=1.\columnwidth]{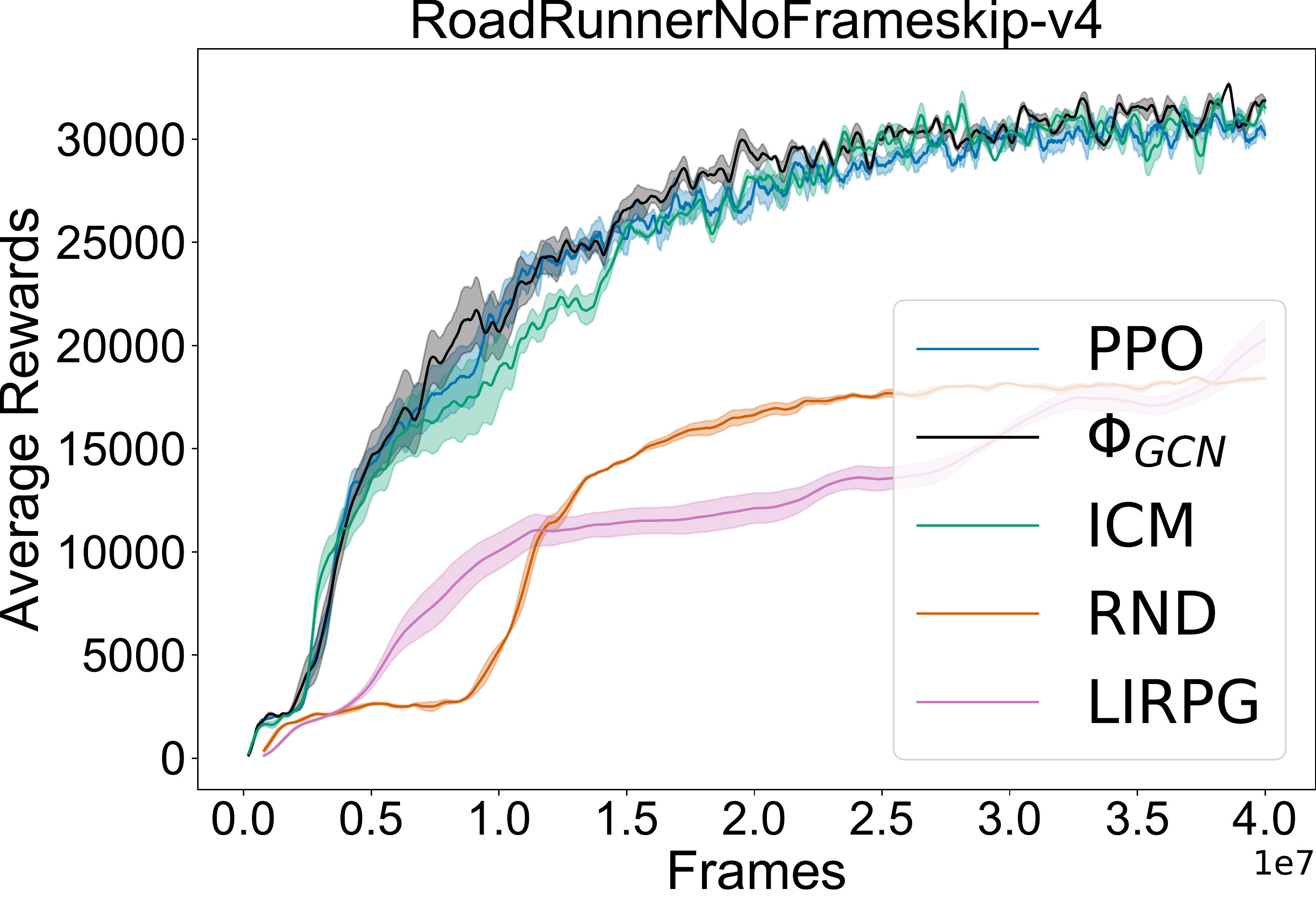}
    \end{subfigure} 
    \begin{subfigure}[c]{0.24\textwidth}
    \centering
      \includegraphics[width=1.\columnwidth]{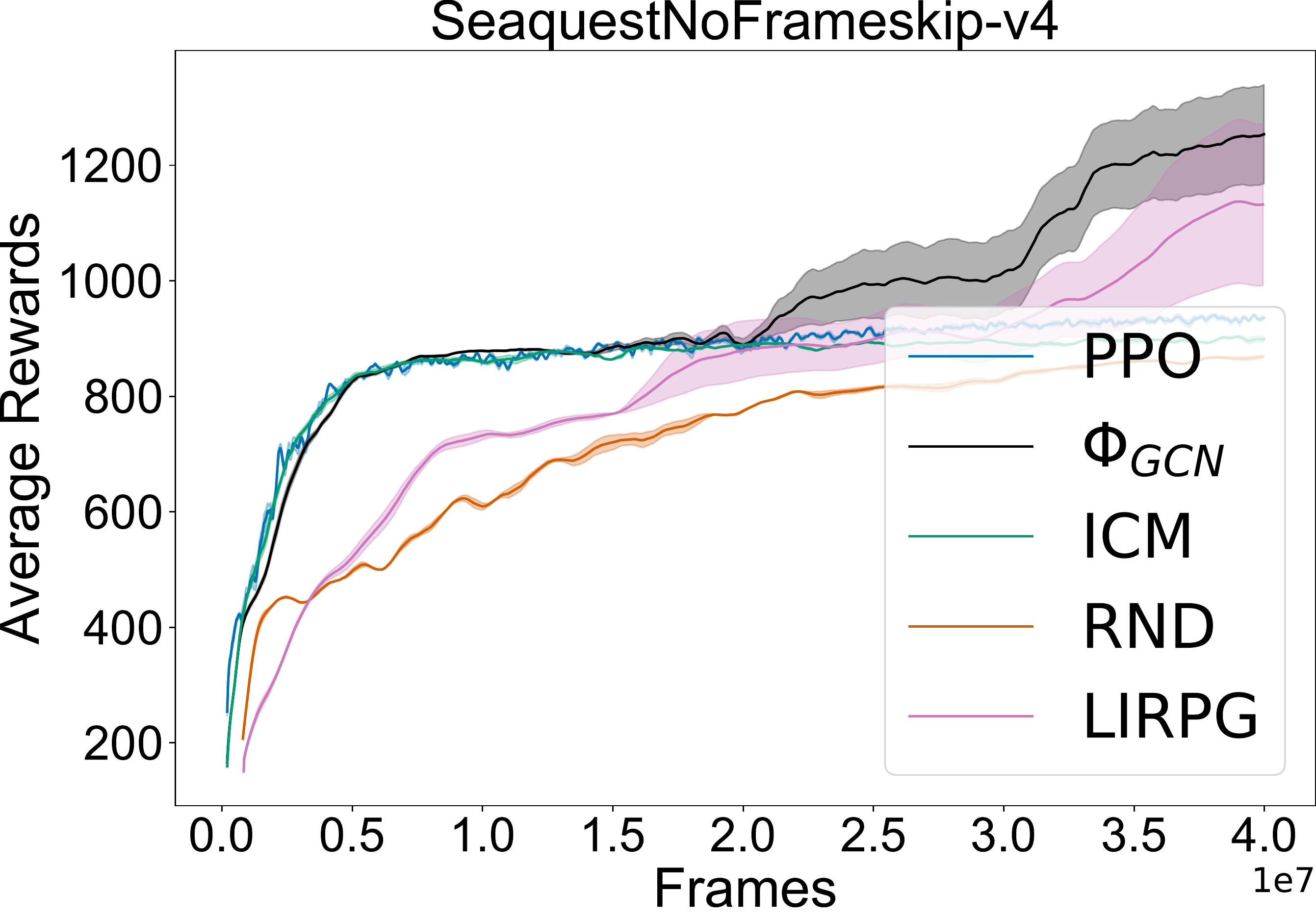}
    \end{subfigure} 
    \begin{subfigure}[c]{0.24\textwidth}
    \centering
      \includegraphics[width=1.\columnwidth]{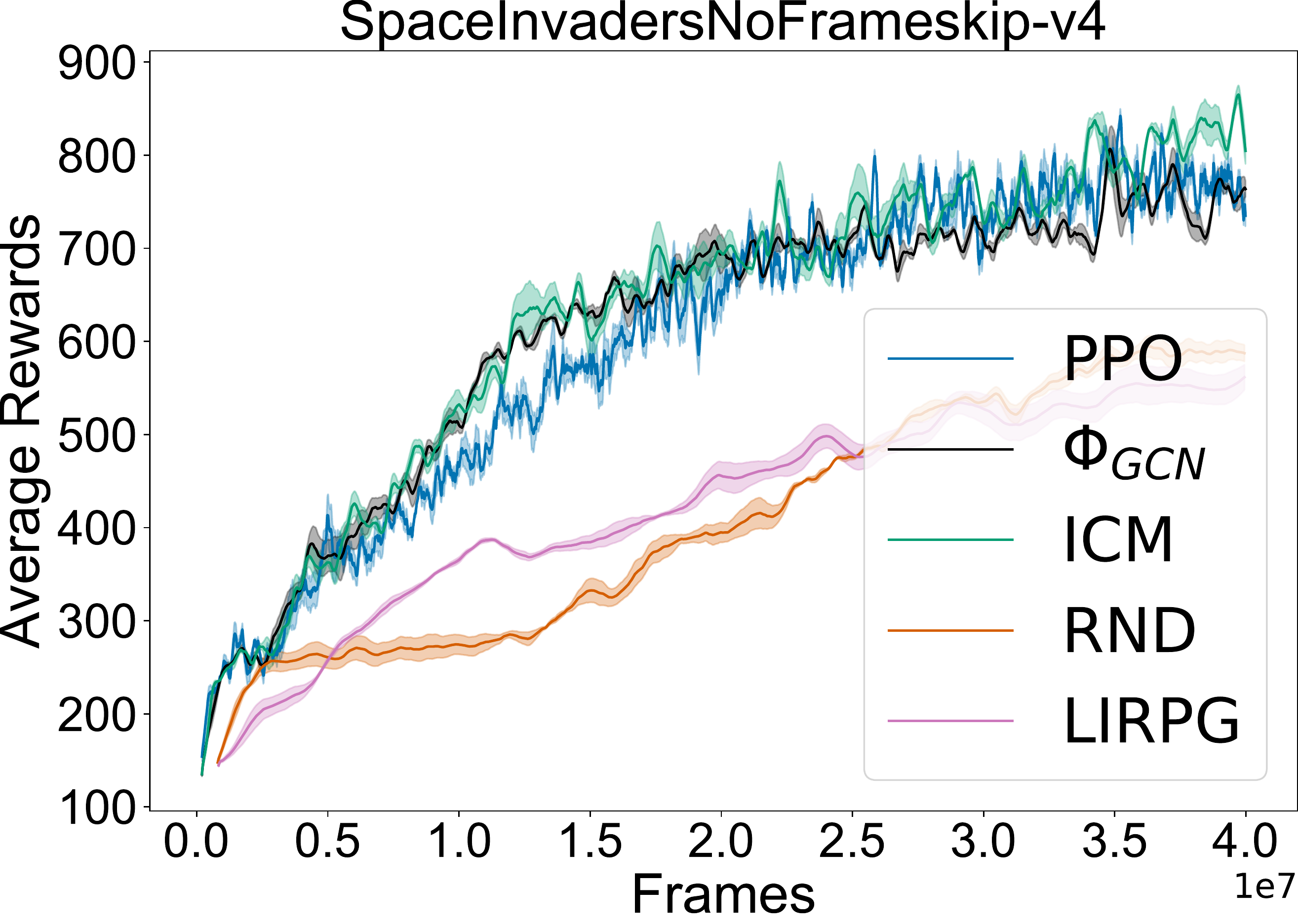}
    \end{subfigure}

    \quad
    
    \begin{subfigure}[c]{0.24\textwidth}
    \centering
      \includegraphics[width=1.\columnwidth]{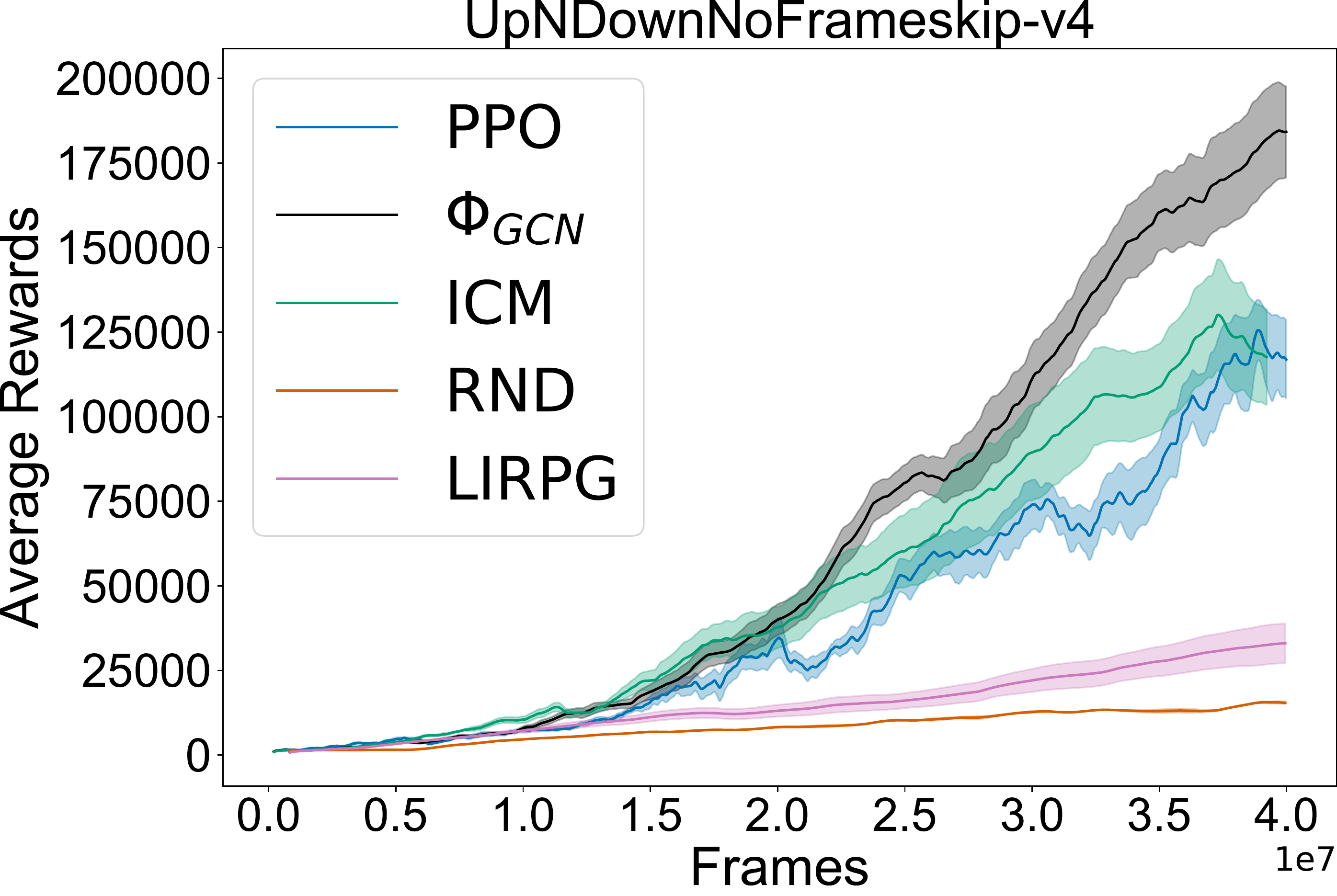}
    \end{subfigure} 
    \begin{subfigure}[c]{0.24\textwidth}
    \centering
      \includegraphics[width=1.\columnwidth]{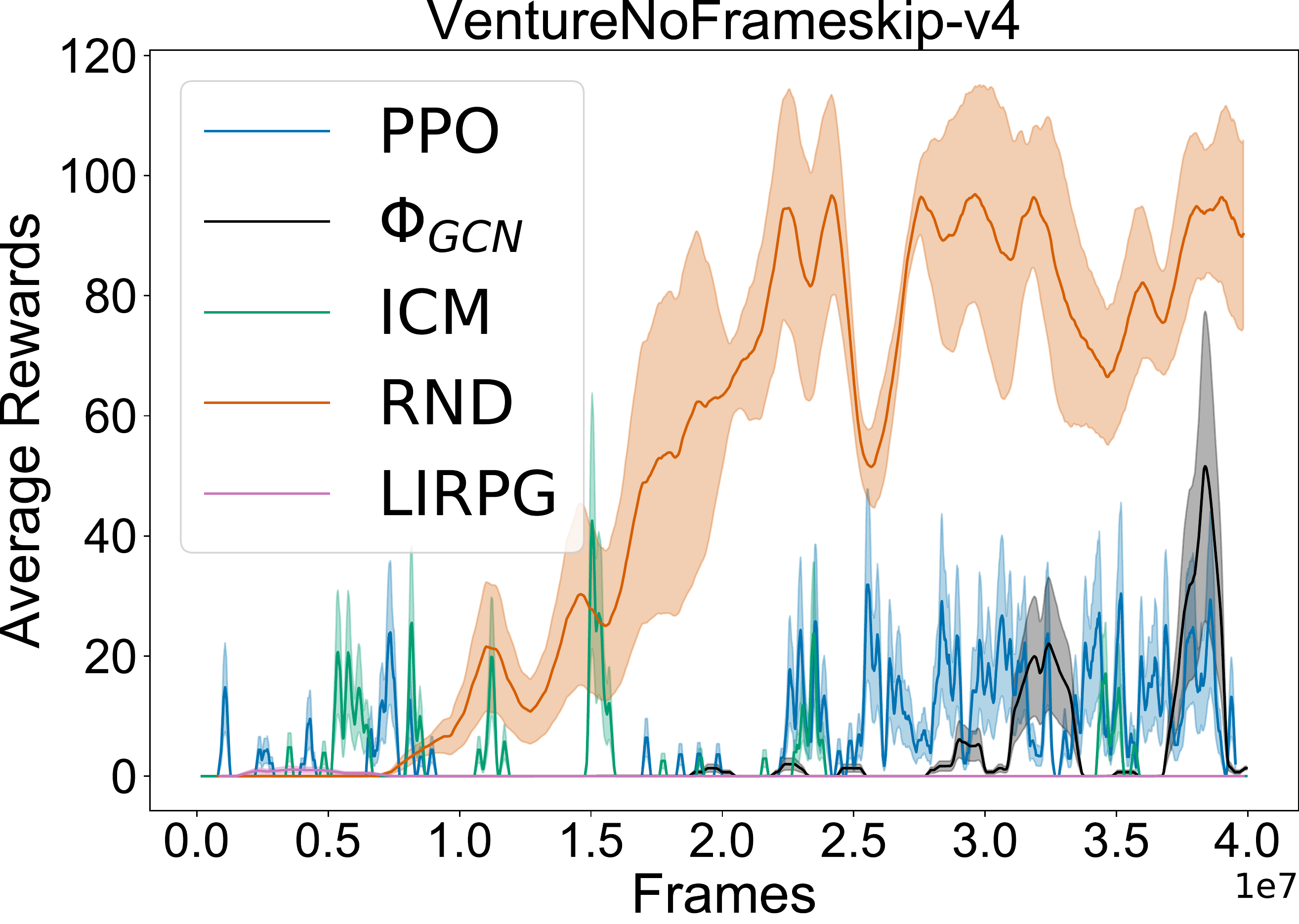}
    \end{subfigure} 
    \begin{subfigure}[c]{0.24\textwidth}
    \centering
      \includegraphics[width=1.\columnwidth]{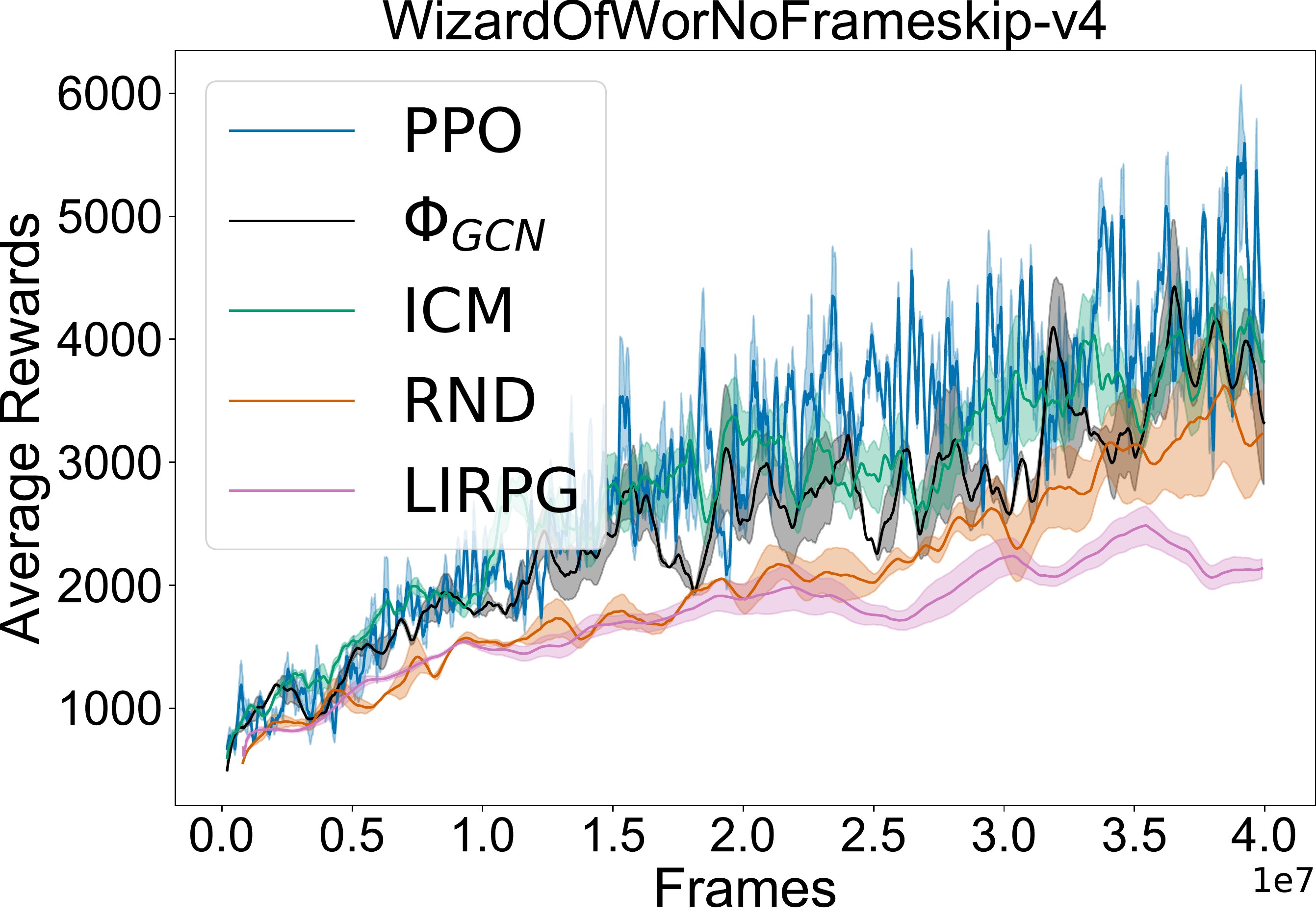}
    \end{subfigure} 
    \begin{subfigure}[c]{0.24\textwidth}
    \centering
      \includegraphics[width=1.\columnwidth]{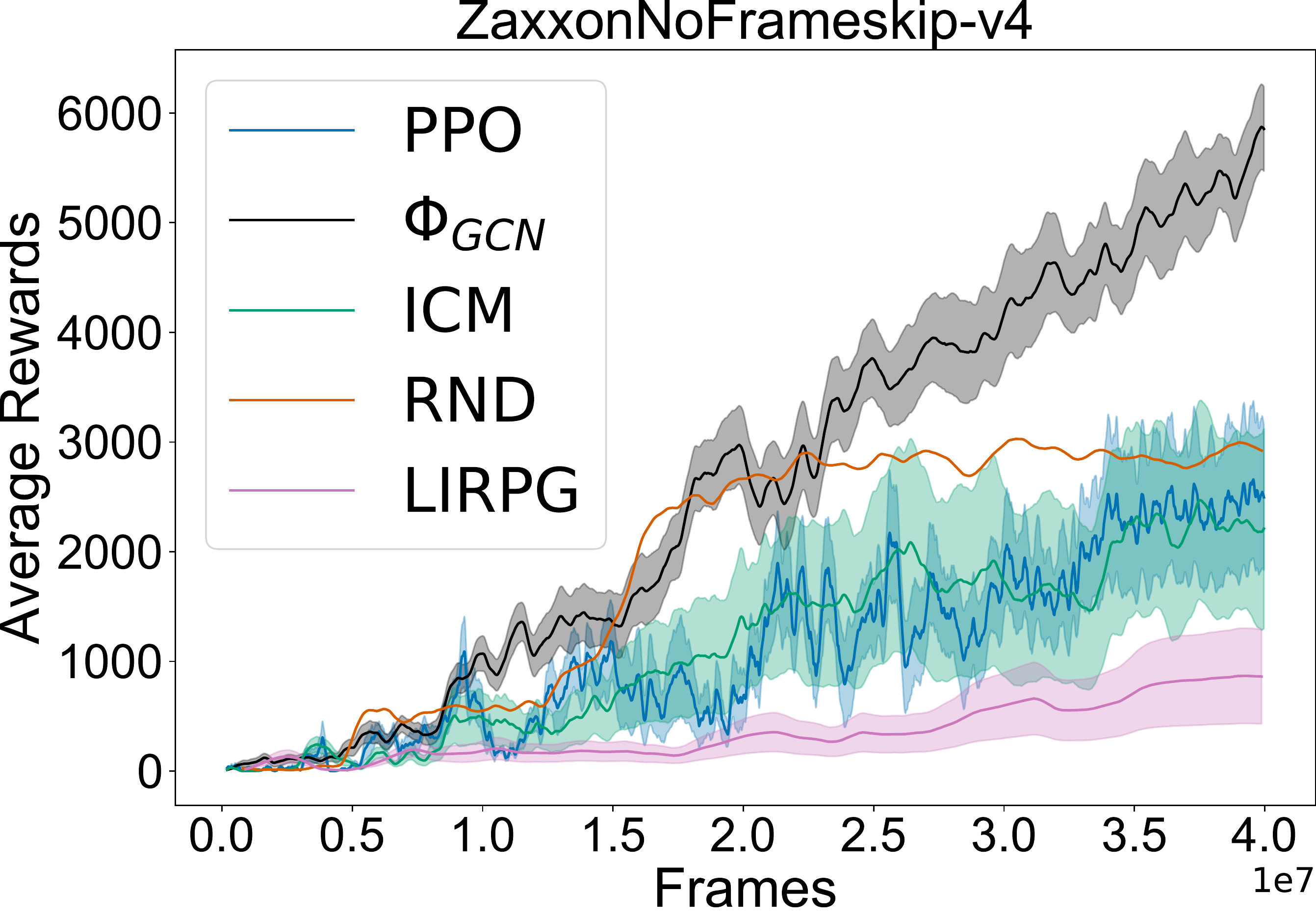}
    \end{subfigure} 
    \caption{Results on 20 Atari games using sticky actions \citep{DBLP:journals/corr/abs-1709-06009}}
    \label{fig:full_atari}
\end{figure}

\newpage
\subsection{MuJoCo}
\label{app:muj}
For the MuJoCo experiments, we based our implementation on \citep{baselines} and we ran the experiments for 3M steps over 10 random seeds. The input provided to the GCN is the last hidden layer of the actor's MLP network.  The architecture for the actor-critic was  kept to be the same with respect to the original codebase. We provide a full list of the values for the hyperparameters:

\begin{table}[hbt!]
    \centering
    \begin{tabular}{c|c}
    Hyperparameter & Value \\
    \hline
     Learing rate & 3e-4\\
     $\gamma$ & 0.99  \\
     $\lambda$ & 0.95  \\
     Entropy coefficient & 0.0  \\
     LR schedule & constant \\
     PPO steps & 2048 \\
     PPO cliping value & 0.1 \\
     \# of minibatches & 32 \\
     \# of processes & 1 \\
     GCN: $\alpha$ (Walker and Ant) & 0.6 \\
     GCN: $\alpha$ (Hopper and HalfCheetah) & 0.7 \\
     GCN: $\eta$ & 1e1 \\
     LIRPG coeff. & 0.01
     \end{tabular}
    \caption{Hyperparameters for the MuJoCo experiment }
    \label{tab:hc}
\end{table}



\clearpage
\subsection{Intepretation as a baseline}
\label{sec:app_baseline}
In the potential-based reward shaping framework \citep{Ng:1999:PIU:645528.657613}, it is shown that the optimal policy will remain invariant to the choice of potential function. This can easily be seen by considering the following relation,
\begin{align*}
    Q^{\pi}(s,a) - \Phi(s) = Q^{\pi}_{\Phi}(s,a) 
\end{align*}
where $\Phi(s)$ is the reward shaping function, $Q^{\pi}(s,a)$ is the policy's action value function and $ Q^{\pi}_{\Phi}(s,a) $ is the action value function with respect to the potential based shaped reward. When used in the context of policy-based methods, rewards shaping can be considered as introducing a baseline. In particular, if we considered the action value function obtained by combining the original reward and shaped reward through the mixing coefficient $\alpha$, that is $Q^{\pi}_{comb}= \alpha  Q^{\pi}(s,a) + (1-\alpha) Q^{\pi}_{\Phi}(s,a)$, we obtain
\begin{align*}
    \deriv[J(\theta)]{\theta} = \sum_{s} d(s;\theta) \sum_{a} \deriv[\pi\left(a \given s\right)]{\theta} \big( Q_{\pi}(s, a) - (1-\alpha) \Phi(s) \big)
\end{align*} 
To investigate the consequences of such a baseline, we consider its effect on the variance of the policy gradient. Using results from \cite{greensmith} (in particular Lemma 9), we denote $\Var(\Phi)$ the variance of the estimate when using $(1-\alpha) \Phi$ as a baseline and we obtain the following upper bound,
\begin{align*}
   \Var(\Phi) = \Var(G) +\expectation \big[ (1-\alpha)^2 \Phi(s)^2 \expectation[ (\deriv[ J(\theta)]{\theta} )^2  |s] - 2 (1-\alpha) \Phi(s)   \expectation[ (\deriv[J(\theta)]{\theta}) ^2 Q_{\pi}(s, a) |s] \big]
\end{align*}
where $\Var(G)$ is the variance obtained by not using any baselines, that is the variance of the return $G$. When compared to the baseline defined as the state value function $V^{\pi}(s)$ (and denoted $\Var(V^{\pi})$), we can show that the highest upper bound will be less or equal. \\

Let's first consider the highest upper bound for the state value function. To do this we have to consider two cases. In the case where the highest reward, $R_{max}$, is smaller in magnitude than the lowest reward, $R_{min}$, (that is $|R_{max}| \leq |R_{min}|$) we obtain the following highest upper bound for the state value function,
\begin{align}
    \Var(V^{\pi}) \leq \Var(G) +\expectation \big[ \big( (\frac{R_{max}}{1-\gamma} )^2  - 2 \frac{R_{max} R_{min}}{(1-\gamma)^2}  \big)  \expectation[ (\deriv[J(\theta)]{\theta}) ^2   |s] \big] \text{ for } |R_{max}| \leq |R_{min}|
    \label{rmax}
\end{align}
 In the case where the highest reward, $R_{max}$, is greater in magnitude than the lowest reward, $R_{min}$, (that is $|R_{max}| \geq |R_{min}|$) we obtain the following highest upper bound for the state value function,
 \begin{align}
    \Var(V^{\pi}) \leq \Var(G) +\expectation \big[ \big( (\frac{R_{min}}{1-\gamma} )^2  - 2 \frac{R_{min} R_{max}}{(1-\gamma)^2}  \big)  \expectation[ (\deriv[J(\theta)]{\theta}) ^2   |s] \big] \text{ for } |R_{min}| \leq |R_{max}|
    \label{rmin}
\end{align}
We notice that these bounds are equal when $|R_{max}| = |R_{min}|$ and are the lowest when both values are low, which suggest that normalizing the rewards (and the returns) is a good general strategy for reducing variance. When using potential based reward shaping, we obtain the following upper bound irrespective of the magnitude of $R_{max}$ and $R_{min}$,
 \begin{align}
    \Var(\Phi) \leq \Var(G) +\expectation \big[ \big( (\frac{(1- \alpha)\sigma(R_{max})}{1-\gamma} )^2  - 2 \frac{ (1-\alpha) \sigma(R_{max}) R_{min} }{(1-\gamma)}  \big)  \expectation[ (\deriv[J(\theta)]{\theta}) ^2   |s] \big] 
\end{align}
where $\sigma$ is the sigmoid function. By inspection, we verify that for values of $R_{max}/(1-\gamma)$ greater than $(1-\alpha)\sigma(R_{max})$, the highest upper bound on the variance is strictly lower in the case of potential based reward shaping as defined by our approach when compared to either bounds obtained for the value function baseline (Eq.\ref{rmax}-\ref{rmin}). This is satisfied by almost all practical applications. Moreover, we notice that $\alpha$ controls the highest upper bound. Therefore, higher values of $\alpha$ are preferred in order to limit the possibility of higher variance.

\subsection{Approximating the adjacency matrix}
\label{app_smooth}

Reconstructing the adjacency matrix is only possible if one tries to explicitly represent the underlying graph. When faced with high-dimensional MDPs, this is simply impractical. As pointed by \cite{MachadoBB17}, there is a trade-off representing accuractely the adjacency matrix and opting for a simpler model through the incidence matrix, denoted as $C$. In their work, \cite{MachadoBB17} show that by estimating the incidence matrix, they can recover the same eigenvectors as the ones obtained from the graph Laplacian. In our case, as we are not aiming to recover the eigenvectors of the graph Laplacian, we instead look at the consequence of using the incidence matrix as a transition operator for message passing.

We begin by noticing that the incidence matrix is in fact the un-normalized random walk matrix, that is $P^{rand} = D^{-1}C$ where $P^{rand}$ is the normalized random walk matrix and $D$ is the degree matrix. We show that the entropy rate of $P^{rand}$ is higher or equal when compared to the adjacency matrix or the true transition matrix $P^{\pi}$ for any policy $\pi$.

We start by providing the definition of entropy rate for the special case where the stochastic process is a Markov Chain $M$:
\begin{align*}
    H(M) = - \sum_{ss'} \mu_s P_{ss'} \log  P{ss'}
\end{align*}
where $\mu_s$ is the stationary distribution of the arbitrary policy $\pi$ taken at state $s$. As we don't have access to this distribution, we will look at each of the rows of the matrices.
We define a random walk matrix which preserve the original MDP's connectivity between states, that is, each non-zero in the original transition matrix $P$ is also a non-zero entry in the random walk matrix $P^{rand}$.
We note that the entropy of a multinomial distribution is maximized when each entry is equal to $1/N$, where $N$ is the total number of entries. As each non-zero entry of the rows of matrix $P^{rand}$ contains such value, the entropy of any row in $P^{rand}$ is higher or equal than any row in any transition matrix $P$,
\begin{align*}
    -\sum_{s'} P_{ss'} \log P_{ss'} \leq -\sum_{s'} P^{rand}_{ss'} \log  P^{rand}_{ss'} \hspace{2mm}
 \forall \hspace{2mm} i, P
 \end{align*}
We also notice that $\sum_s \gamma_s \sum_{s'} P^{rand}_{ss'} \log  P^{rand}_{ss'} = \sum_{s'} P^{rand}_{ss'} \log  P^{rand}_{ss'}$ as $\gamma$ is a probability vector that sums to 1, leading to a convex combination. We finally get that,
\begin{align*}
    -\sum_{s} \mu_{s} \sum_{s'} P_{ss'} \log P_{ss'} &\leq  -\sum_{s'} P^{rand}_{ss'} \log  P^{rand}_{ss'}\\
    -\sum_{s} \mu_{s} \sum_{s'} P_{ss'} \log P_{ss'} & \leq -\sum_{s} \gamma_s \sum_{s'} P^{rand}_{ss'} \log  P^{rand}_{ss'}
\end{align*}
 The vector $\gamma$ is any vector summing to 1 and as such can be taken as the stationary distribution of the random walk matrix $P^{rand}$. As a direct consequence, the entropy rate of any MDP is less or equal to the entropy rate of the equivalent MDP where we have replaced the transition matrix with the random walk matrix $P^{rand}$. The derivation for the adjacency matrix follows the same intuition.

As the entropy rate gets higher, the complexity of the stochastic process and the number of probable paths from one state to any other state will increase. Recall that the GCN's loss function $ \mathcal{L}_{prop}$ implements the recursive operation of message passing.

Intuitively, this means that the resulting distribution over states, $\Phi$, will be tend to be more diffuse in nature than the one obtained by true transition matrix or the adjacency matrix. This diffused signal can then be leveraged by the agent to learn more efficiently.  However, it is important to notice that this does not guarantee better exploration, but instead an assignment mechanism that will produce a smoother signal, which the agent can learn to exploit. This also highlights a natural drawback previously observed in the PVF framework \citep{Mahadevan2005}. In using diffusion operators the dependency on the policy $\pi$ is ignored. However, the potential function $\Phi(s)$ proposed in this work is not meant to replace the value function but instead to complement it in order to accelerate learning.
\end{document}